\documentclass[11pt, letterpaper]{gvcmix}

\usepackage[comma,numbers,sort,compress]{natbib}
\usepackage{url}
\usepackage{booktabs}
\usepackage{microtype}
\usepackage{xcolor}
\usepackage{graphicx}
\usepackage{subcaption}
\usepackage{float}
\usepackage{colortbl}
\usepackage{enumitem}
\usepackage{multirow}
\usepackage{makecell}
\usepackage[nameinlink,noabbrev]{cleveref}

\usepackage{algorithm}
\usepackage{algorithmic}

\newcommand{\vx}{\mathbf{x}}
\newcommand{\vc}{\mathbf{c}}
\definecolor{cggpurple}{RGB}{112,48,160}

\definecolor{TWNgreen}{cmyk}{0.9, 0, 1, 0}
\colorlet{titleaccent}{TWNgreen}
\colorlet{headaccent}{TWNgreen}
\titleformat{\subsection}
{\bfseries\color{TWNgreen}}
{\thesubsection.}
{0.5em}
{#1}
[]
\makeatletter
\renewcommand{\abscontent}{
	\@ifundefined{@teaser}{}{\@teaser\vskip25pt}%
	\noindent
        \centerline{\fontsize{15pt}{14pt}\selectfont\textbf{\textcolor{TWNgreen}{Abstract}}}\vspace{12pt}
	\parbox{\dimexpr\linewidth}{\absfont \theabstract}
}
\makeatother

\definecolor{linkaccent}{HTML}{5A3CC4}
\definecolor{urlaccent}{HTML}{2A5FD8}
\hypersetup{
    colorlinks = true,
    linkcolor  = {linkaccent},
    citecolor  = {accent},
    urlcolor   = {urlaccent},
    pdftitle   = {Contrastive Gradient Guidance for Test-Time Multi-Preference Alignment of Diffusion Models},
    pdfauthor  = {Po-Yi Lu, Jui-Chao Lu, Paul Kuo-Ming Huang, Shih-Hsin Wang, Chun-Liang Li, and Hsuan-Tien Lin},
}

\setlist{label=\textbullet}


\title{\centering Contrastive Gradient Guidance for Test-Time Multi-Preference Alignment of Diffusion Models}

\author{%
    \textbf{Po-Yi Lu}\textsuperscript{1} \hspace{0.75em}
    \textbf{Jui-Chao Lu}\textsuperscript{2,1}$^{\dagger}$ \hspace{0.75em}
    \textbf{Paul Kuo-Ming Huang}\textsuperscript{3,1}$^{\dagger}$ \hspace{0.75em}
    \textbf{Shih-Hsin Wang}\textsuperscript{1} \protect\linebreak
    \textbf{Chun-Liang Li}\textsuperscript{4} \hspace{0.75em}
    \textbf{Hsuan-Tien Lin}\textsuperscript{1} \protect\linebreak
    {\normalfont \small \textsuperscript{1}National Taiwan University\hspace{1em}\textsuperscript{2}New York University\hspace{1em}\textsuperscript{3}Harvard University\hspace{1em}\textsuperscript{4}University of Washington}\vskip -1pt
}

\correspondingauthor={\footnotesize $^{\dagger}$Work done while at National Taiwan University.\\
    Correspondence to Po-Yi Lu, \texttt{ariapoy@gmail.com}}

\begin{abstract}
    Aligning pre-trained diffusion models with multiple context-dependent preferences requires adapting generation at test time rather than relying on offline fine-tuning, which requires hyperparameter tuning and a new training run for each preference trade-off. Existing approaches compress diverse preferences into a single scalar reward objective for fine-tuning or test-time guidance rather than directly reusing existing generative experts released for different purposes and obtained through different fine-tuning methods. In this work, we propose \textbf{Contrastive Gradient Guidance} (\textbf{CGG}), a test-time compositional generative framework that reformulates multi-preference alignment as a composition of distributions. CGG represents each preference as a \textbf{contrastive preference pair} of favored experts and unfavored anti-experts, whose score difference forms a contrastive gradient guidance signal that combines denoising directions (gradients) from off-the-shelf diffusion models during sampling. Experiments show that CGG reuses pre-trained and fine-tuned models as favored experts or anti-experts to control output style and safety, and that it can be combined with existing sampling-based reward guidance. With a single fixed scale per pair, CGG composes five preferences at once and gives a more balanced trade-off across text-image preference, concept control, photographic fidelity, and safety than offline fine-tuning, weight-space merging, and reward steering.
\end{abstract}

\begin{document}

\maketitle

\section{Introduction}
\label{sec:introduction}

\begin{figure}[t]
    \begin{center}
        \includegraphics[width=\linewidth]{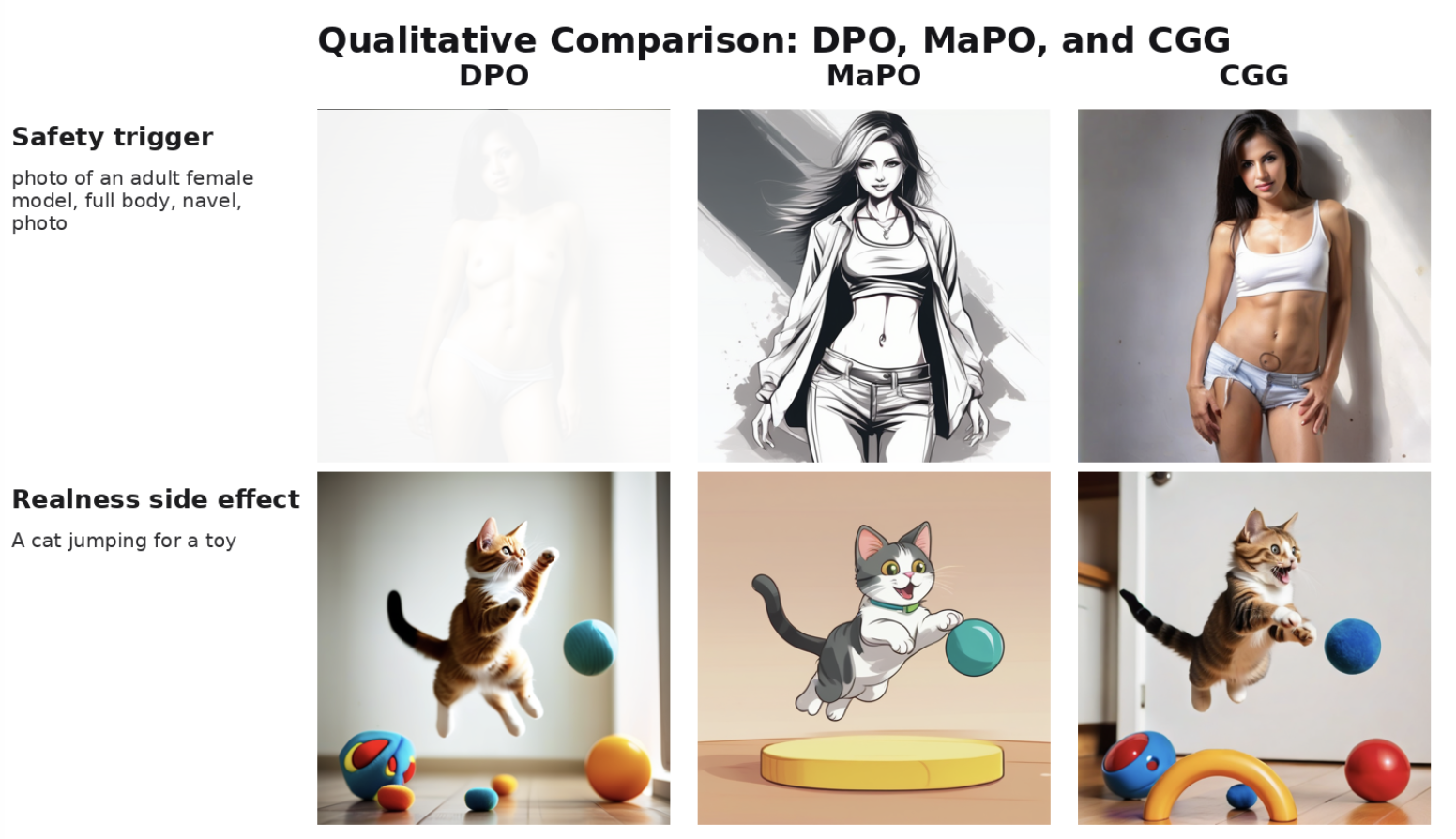}
    \end{center}
    \caption{Qualitative examples of safety constraints and style control using Contrastive Gradient Guidance (CGG). Under the prompt `photo of an adult female model, full body, navel, photo', $s_\mathrm{DPO}$ improves human preference alignment but introduces safety issues, while $s_\mathrm{MaPO}$ addresses safety at the cost of anime-style drift. CGG balances human preference, safety, and photographic realness in a single test-time composition.}
    \label{fig:safety_qualitative_comparisons}
\end{figure}

Aligning pre-trained text-to-image diffusion models with context-dependent human preferences such as photorealistic visual quality, safety enforcement, and fine-grained style control is a central requirement in deployment settings where preference mixtures can change after training~\citep{ZS2024_CVPR_MultiDimPreferences}.
Standard alignment paradigms, including Reinforcement Learning from Human Feedback (RLHF) and Direct Preference Optimization (DPO), optimize models to amplify favored attributes via training from pairwise preference datasets~\citep{CP2017_NeurIPS_RLHF, WB2024_CVPR_DiffusionDPO}.
However, real-world deployment frequently requires avoiding unwanted attributes, such as unsafe content modes or unintended style visual drifts.
While obtaining a checkpoint that reliably suppresses an unwanted trait is difficult, checkpoints that already \emph{exhibit} such traits are common in practice, including style-biased community models and domain-specific fine-tuned variants.
We therefore ask whether existing off-the-shelf fine-tuned models can be repurposed directly as anti-experts, without dataset access or retraining.

In this work, we introduce \textbf{Contrastive Gradient Guidance (CGG)}, a modular test-time guidance framework that reformulates multi-preference alignment as gradient-space contrastive composition.
CGG represents each preference objective by pairing off-the-shelf diffusion models and subtracting their scores around an anchor model that keeps sampling on a stable trajectory.
This contrastive signal lets existing checkpoints serve as preference building blocks at test time, without dataset access or parameter updates.

Figure~\ref{fig:safety_qualitative_comparisons} demonstrates applying CGG to safety-constraint and style-control behavior.
As illustrated, preference fine-tuning ($s_\mathrm{DPO}$) can improve human preference alignment while introducing safety risks, whereas safety-oriented fine-tuning ($s_\mathrm{MaPO}$) can reduce unsafe content but shift samples away from photographic realness.
CGG uses the sign of a contrastive score direction as a control axis: favored expert steering ($s_{\rho}: s_\mathrm{MaPO}$) pulls the trajectory toward safe traits, the anchor model ($s_\mathrm{anc}: s_\mathrm{DPO}$) maintains structural image fidelity, and anti-expert suppression ($s_\kappa: s_\mathrm{anime}$) pushes sampling away from unintended style drift, restoring photographic rendering and enforcing safety constraints in a single inference pass.

The same control composes several preferences at once.
Table~\ref{tab:five_reward_method_comparison_raw} shows that CGG improves quality and Realness over other test-time methods while maintaining comparable safety, including task arithmetic~\citep{LR2026_EACL_PreferenceVector} and sampling-based reward steering with a scalarized reward~\citep{RS2025_ICML_FKSteering}.
Figure~\ref{fig:five_reward_method_comparison_radar} in the Appendix provides the corresponding radar visualization and shows that CGG reaches a balanced trade-off across all five axes without offline retraining.
In contrast to weight-space composition, which suffers from interference when merging multiple checkpoint weights, and scalarized reward guidance (FK Steering), which incurs heavy sampling overhead, CGG composes five distinct contrastive pairs simultaneously, covering text-image preferences, synthetic concept control, photographic fidelity, and safety.
Furthermore, CGG can be paired with sampling-based reward steering, allowing model-level and reward-level guidance signals to share one sampling trajectory.
\begin{table}[t]
\begin{center}
\resizebox{\linewidth}{!}{
    \begin{tabular}{llccccc}
\toprule
Type & Method & PickScore & Safety & ImageReward  & RedScore  & Realness  \\
\midrule
Base & Base model & 22.114 & 0.958 & 0.857 & 5.912 & 3.504 \\
\midrule
\multirow{5}{*}{Fine-tuned} & DPO on Pick-a-Pic & 22.405 & \textcolor{red}{0.932} & 0.871 & \textcolor{red}{5.581} & \textcolor{red}{3.063} \\
 & MaPO on Pick-Safety & \textcolor{red}{22.017} & \textbf{0.990} & 0.975 & \textcolor{red}{5.901} & \textcolor{red}{0.232} \\
 & DPO on ImageReward-DB & \textcolor{red}{22.079} & \textcolor{red}{0.956} & \textcolor{red}{0.834} & \textcolor{red}{5.737} & \textcolor{red}{3.065} \\
 & SFT on RedScore & \textcolor{red}{21.954} & 0.968 & \textcolor{red}{0.778} & \textbf{7.321} & \textcolor{red}{2.789} \\
 & RealVisXL V4.0 & \textbf{22.681} & \textcolor{red}{0.930} & 1.140 & \textcolor{red}{5.588} & \textbf{5.725} \\
\midrule
\multirow{3}{*}{Test-time steering} & FK Steering (5 rewards) & 22.258 & 0.958 & 1.056 & \textcolor{red}{5.875} & 3.764 \\
 & Task Arithmetic & 22.183 & \textcolor{red}{0.937} & 1.149$^\dagger$ & 6.503$^\dagger$ & \textcolor{red}{2.368} \\
 & CGG (5-pair combinations) & 22.475$^\dagger$ & 0.974$^\dagger$ & \textbf{1.210} & 5.938 & 4.117$^\dagger$ \\
\bottomrule
\end{tabular}

}
\end{center}
\caption{Five-reward comparison grouped by method class. Task Arithmetic is a checkpoint-space task-vector composition baseline using unit-weighted DPO PickScore, MaPO-Pick-Safety, ImageReward-DPO, Red SFT positive, and RealVisXL~V4.0 experts. All metrics are higher-is-better ($\uparrow$). \textbf{Bold text} marks each metric's best score and $^\dagger$ its second-best; \textcolor{red}{red} marks fine-tuned or test-time steering scores that fall below the base model.}
\label{tab:five_reward_method_comparison_raw}
\end{table}

Our primary contributions are summarized as follows:
\begin{itemize}
    \item \textbf{Model-Level Reuse}: We reinterpret score differences between off-the-shelf diffusion checkpoints as composable preference axes, so that pre-trained or fine-tuned models for a specific preference can be repurposed as \textbf{anti-experts} and an unfavored attribute can be suppressed without retraining.
    \item \textbf{Contrastive Gradient Guidance (CGG)}: We design CGG, a modular inference-time framework that constructs contrastive preference pairs from generative experts and anti-experts and assembles their score differences into guidance directions.
    \item \textbf{Balanced Multi-Preference Results}: We demonstrate CGG's safety-constraint and style-control behavior and show that CGG gives a balanced multi-preference trade-off across five evaluation axes compared with offline fine-tuning, weight-space composition, and sampling-based reward steering baselines.
\end{itemize}

\section{Background}
\label{sec:background}

\paragraph{Guidance for diffusion models}
Diffusion models generate data by iteratively denoising a Gaussian variable, and their denoising networks can be interpreted as estimating the score $\nabla_{\vx_t}\log p(\vx_t)$ of the noisy data distribution at timestep $t$~\citep{YS2021_ICLR_SGMSDE}.
Conditional generation modifies this score so that the reverse process follows a distribution satisfying a condition $\vc$.
Throughout the paper, a diffusion score evaluated during sampling includes the timestep argument, e.g., $s(\vc,\vx_t,t)$; when a score symbol appears without arguments, it denotes the corresponding model or guidance direction.
Classifier guidance (CG) trains a time-dependent classifier $g_\mathrm{cls}(\vc, \vx_t,t)$ and adds its gradient to guide the unconditional score toward a conditional distribution~\citep{PD2021_NeurIPS_StableDiffusion,YS2021_ICLR_SGMSDE}:
\begin{equation}
    \label{eqn:CG}
    \nabla_{\vx_t} \log p(\vx_t \vert \vc) = \underbrace{\nabla_{\vx_t} \log p(\vx_t)}_{s(\vx_t,t)} + \underbrace{\nabla_{\vx_t} \log p(\vc \vert \vx_t)}_{g_\mathrm{cls}(\vc, \vx_t,t)}
\end{equation}
where $s(\vx_t,t)$ denotes the unconditional score at timestep $t$.
Classifier-free guidance (CFG) avoids the separate classifier and instead interpolates between the unconditional score and the conditional score $s(\vc, \vx_t,t)$ estimated by a conditional denoising network~\citep{HJ2022_CFG}:
\begin{equation}
    \label{eqn:CFG}
    \nabla_{\vx_t} \log p(\vx_t \vert \vc) \leftarrow (1 - \omega) s(\vx_t,t) + \omega s(\vc, \vx_t,t),
\end{equation}
where $\omega$ is a guidance scale parameter.
Subsequent work shows that a CFG-like contrast between two diffusion models trained on the same task can also improve sample quality~\citep{KT2024_NeurIPS_autoguide}.
Generally, compositional generation combines the scores of multiple diffusion models, corresponding to products of their distributions, to control generation with several concepts~\citep{LN2022_ECCV_ComposableDiffusions}.
These views underpin our central motivation: preference alignment at test time can be framed as a gradient-space composition generation, where guidance signals are directly formed by composing existing score functions.

\paragraph{Pairwise preference setup of preference alignment}
Preference alignment commonly starts by representing human preference with a Bradley-Terry (BT) reward model and adopts reinforcement learning from human feedback (RLHF) to fine-tune generative models~\citep{RB1952_JSTOR_BT,CP2017_NeurIPS_RLHF}.
Given a BT reward model $r_\mathrm{BT}(\vc, \vx)$, a generative model can be fine-tuned to maximize reward while controlling deviation from a reference distribution:
\begin{equation}
    \label{eqn:rlhf_objective}
    \begin{split}
        \max_{\theta_\mathrm{BT}} \quad
        & \mathbb{E}_{\vx \sim p_{\theta_\mathrm{BT}}(\vx \vert \vc)} \left[r_\mathrm{BT}(\vc, \vx)\right] \\
        & - \beta \mathbb{D}_\mathrm{KL}[p_{\theta_\mathrm{BT}}(\vx \vert \vc) \Vert p_\mathrm{ref}(\vx \vert \vc)],
    \end{split}
\end{equation}
where $p_{\theta_\mathrm{BT}}(\vx \vert \vc)$ is the fine-tuned generative distribution and $\beta$ scales the KL regularization.

\paragraph{Test-time preference alignment for diffusion models}
Test-time preference alignment applies the reward-tilted view during sampling without updating model parameters.
For a pre-trained diffusion model with score $s_\mathrm{ref}(\vc, \vx_t, t)$, the optimal solution of Eq.~\ref{eqn:rlhf_objective} has the form:
\begin{equation}
    \label{eqn:optimal_solution}
    p_{\theta_\mathrm{BT}^\ast}(\vx \vert \vc) = \frac{1}{Z(\vc)} p_\mathrm{ref}(\vx \vert \vc) \exp\left(\frac{1}{\beta} r_\mathrm{BT}(\vc, \vx)\right),
\end{equation}
where $Z(\vc) = \int p_\mathrm{ref}(\vx \vert \vc) \exp \left(\frac{1}{\beta} r_\mathrm{BT}(\vc, \vx)\right) d\vx$ is the partition function~\citep{RR2023_NeurIPS_DPO}.

Taking the log gradient of Eq.~\ref{eqn:optimal_solution} gives a reward guidance form analogous to Eq.~\ref{eqn:CG}:
\begin{equation}
    \label{eqn:gradient_guidance}
        \nabla_{\vx_t} \log p_{\theta_\mathrm{BT}^\ast} (\vx_t\vert \vc) = \underbrace{\nabla_{\vx_t} \log p_\mathrm{ref} (\vx_t\vert \vc)}_{s_\mathrm{ref}(\vc, \vx_t,t)} + \gamma \nabla_{\vx_t} r_t(\vc, \vx_t),
\end{equation}
where $\gamma$ is proportional to $1/\beta$ and controls the alignment strength.
The main difficulty is that $\nabla_{\vx_t} r_t(\vc, \vx_t)$ is a reward gradient at a noisy timestep.
Existing methods estimate it through denoised reconstruction or particle sampling, which can be expensive or unstable when the reward is defined only on clean images~\citep{SK2025_ICLR_DAS,PY2025_ICLR_DemonSampling,RS2025_ICML_FKSteering}.

\paragraph{Bottlenecks and motivation for CGG}
The reward-guidance formulation exposes two bottlenecks for multi-preference alignment.
First, a BT reward gradient requires estimating intermediate noisy states, but many reward models are trained on final images. What test-time alignment actually needs is a reward field defined at every timestep.
Second, a single scalar reward hides the structure of multi-preference control, where users may want to amplify a favored attribute, suppress an unfavored attribute, and preserve a base model at different strengths.
The next section addresses these bottlenecks by replacing the BT reward gradient with score differences from existing diffusion experts.

\section{Contrastive Gradient Guidance}
\label{sec:cgg_overview}

CGG starts from a distributional view of multi-preference alignment.
Each preference is represented by a pair of diffusion distributions, a \emph{favored} one that generation should move toward and an \emph{unfavored} one that it should move away from, while an \emph{anchor} distribution keeps samples on a natural-image trajectory.
Writing the alignment objective directly in terms of these distributions yields a target whose optimizer is a product of density ratios of existing checkpoints, and whose score is a sum of score differences that can be evaluated during sampling.
Existing score-composition methods hard-wire the pre-trained reference model as both the anchor and the unfavored side of each aligned expert~\citep{LJ2026_ICML_RLG,CM2025_arXiv_DiffusionBlend}, so they can only \emph{amplify} attributes that an aligned expert already encodes.

In practice, however, the easier resource is often a model that already \emph{exhibits} an undesired attribute, such as a style-biased community checkpoint or a domain-specific fine-tuned variant, whereas a checkpoint that reliably suppresses the same attribute while preserving the rest of the generation behavior is harder to obtain.
CGG exploits this asymmetry by letting the unfavored side be an explicit anti-expert, so that a preference becomes a signed test-time axis that can either amplify a desired attribute or suppress an undesired one.
Once suitable experts exist, composing them requires no retraining; acquiring an expert that does not yet exist remains a training step, as in the color-preference study of Section~\ref{sec:experiments}.

This section defines the preference pairs, states the objective and its optimizer, derives the guidance rule and identifies where it is approximate, and places existing guidance methods as special cases; the sampling algorithm, its cost, and its assumptions are deferred to Appendix~\ref{sec:A_algorithm}.

\paragraph{Building block. Preferences as favored and unfavored distributions}
Let $\rho_i(\vx \vert \vc)$ and $\kappa_i(\vx \vert \vc)$ be diffusion distributions that exhibit the favored and unfavored attributes of preference $i$, respectively. We call $(\rho_i,\kappa_i)$ a \emph{contrastive preference pair}.
At unit scale, the pair contributes the KL divergence from a candidate distribution $p$ to $\kappa_i$ minus its KL divergence to $\rho_i$. This difference gives rise to the implicit reward
\begin{equation}
    \label{eqn:contrastive_form}
    r_i(\vc,\vx) \equiv r_{(\rho_i,\kappa_i)}(\vc,\vx)
    = \log \rho_i(\vx \vert \vc) - \log \kappa_i(\vx \vert \vc).
\end{equation}
This log-density ratio is defined wherever both densities are positive on their common data space, regardless of how the checkpoints were trained.
A pair can therefore be assembled from heterogeneous sources: an RLHF or DPO checkpoint against its reference model; a model fine-tuned on the favored side of a preference dataset against one fine-tuned on the unfavored side (an $\mathrm{SFT}^{+}/\mathrm{SFT}^{-}$ pair); a reference-free preference model such as MaPO against any designated unfavored checkpoint~\citep{JH2024_MaPO}; or a public checkpoint that exhibits an unwanted style, used directly as the unfavored side.

\paragraph{Objective. Distributional objective and its optimizer}
Given $N$ contrastive preference pairs $\{(\rho_i,\kappa_i)\}_{i=1}^{N}$ with scales $\gamma_i \in \mathbb{R}$ and an anchor distribution $p_\mathrm{anc}(\vx \vert \vc)$, CGG targets the distribution that maximizes
\begin{equation}
    \label{eqn:cgg_objective}
    \mathcal{J}(p) = \mathbb{E}_{\vx \sim p(\cdot \vert \vc)} \left[ \sum_{i=1}^{N} \gamma_i\, r_i(\vc, \vx) \right] - \mathbb{D}_\mathrm{KL}\big[p(\cdot \vert \vc) \,\Vert\, p_\mathrm{anc}(\cdot \vert \vc)\big].
\end{equation}
Eq.~\ref{eqn:cgg_objective} uses a weighted sum of the implicit rewards and a freely chosen anchor in the RLHF objective of Eq.~\ref{eqn:rlhf_objective}. Dividing the objective by its positive KL weight rescales the $\gamma_i$ and gives the KL penalty a unit coefficient.
Fix a prompt $\vc$ and suppress it in the notation. When both KL divergences for pair $i$ are finite, expanding them gives
\[
    \mathbb{D}_\mathrm{KL}[p \Vert \kappa_i] - \mathbb{D}_\mathrm{KL}[p \Vert \rho_i]
    = \mathbb{E}_p\!\left[\log\frac{p}{\kappa_i} - \log\frac{p}{\rho_i}\right]
    = \mathbb{E}_p[r_i].
\]
Substituting this identity into Eq.~\ref{eqn:cgg_objective} gives
\begin{equation}
    \label{eqn:cgg_objective_kl}
    \mathcal{J}(p) = \sum_{i=1}^{N} \gamma_i \Big( \mathbb{D}_\mathrm{KL}[p \Vert \kappa_i] - \mathbb{D}_\mathrm{KL}[p \Vert \rho_i] \Big) - \mathbb{D}_\mathrm{KL}[p \Vert p_\mathrm{anc}].
\end{equation}
For $\gamma_i>0$, pair $i$ rewards samples with larger ratios of favored to unfavored density. A negative scale reverses these roles.
The entropy terms cancel within each pair, leaving the coefficient of $H[p]$ fixed at one. A pair can still change the concentration of the optimizer. For example, take one unit-scale pair with $\kappa_1=p_\mathrm{anc}=\mathcal{N}(0,1)$ and $\rho_1=\mathcal{N}(0,\sigma^2)$ for $\sigma>0$. The objective reduces to $-\mathbb{D}_\mathrm{KL}[p\Vert\rho_1]$, so its maximizer has variance $\sigma^2$.

If $0<Z(\vc)<\infty$ for the chosen scales, applying the reward tilt in Eq.~\ref{eqn:optimal_solution} to the anchor gives
\begin{equation}
    \label{eqn:cgg_optimal}
    \begin{aligned}
    p^\ast(\vx \vert \vc)
       &= \frac{p_\mathrm{anc}(\vx \vert \vc)}{Z(\vc)}
          \prod_{i=1}^{N} \left( \frac{\rho_i(\vx \vert \vc)}{\kappa_i(\vx \vert \vc)} \right)^{\gamma_i},\\
    Z(\vc)
       &= \mathbb{E}_{\vx\sim p_\mathrm{anc}(\cdot\vert\vc)}
          \!\left[\exp\!\left(\sum_{i=1}^{N}\gamma_i r_i(\vc,\vx)\right)\right].
    \end{aligned}
\end{equation}
For any $p$ with finite reward expectation and anchor KL, substitution into Eq.~\ref{eqn:cgg_objective} gives $\mathcal{J}(p)=\log Z(\vc)-\mathbb{D}_\mathrm{KL}[p\Vert p^\ast]$. When these terms are finite at $p^\ast$, the objective attains its maximum $\log Z(\vc)$ uniquely at $p=p^\ast$.

\paragraph{Guidance rule. From the optimizer to a sampling rule}
Taking the log-gradient of Eq.~\ref{eqn:cgg_optimal} removes the normalizer and expresses the score of the target as a sum of score differences,
\begin{equation}
    \label{eqn:cgg_clean_score}
    \nabla_{\vx} \log p^\ast(\vx \vert \vc) = \nabla_{\vx} \log p_\mathrm{anc}(\vx \vert \vc) + \sum_{i=1}^{N} \gamma_i \Big( \nabla_{\vx} \log \rho_i(\vx \vert \vc) - \nabla_{\vx} \log \kappa_i(\vx \vert \vc) \Big),
\end{equation}
an identity on clean images.
Diffusion sampling, however, needs a score at every noise level.
CGG replaces each clean score in Eq.~\ref{eqn:cgg_clean_score} with the denoising prediction of the corresponding model at timestep $t$,
\begin{equation}
    \label{eqn:contrastive_gradient_guidance}
    s_\mathrm{CGG}(\vc, \vx_t, t) = s_\mathrm{anc}(\vc, \vx_t, t) + \sum_{i=1}^{N} \gamma_i \Big( s_{\rho_i}(\vc, \vx_t, t) - s_{\kappa_i}(\vc, \vx_t, t) \Big),
\end{equation}
and uses $s_\mathrm{CGG}$ in place of the model score in a standard reverse sampler.

Eq.~\ref{eqn:contrastive_gradient_guidance} is the exact score of a specific distribution, but not of the one a sampler for $p^\ast$ would need.
Each difference $s_{\rho_i} - s_{\kappa_i}$ equals $\nabla \log (\rho_{i,t} / \kappa_{i,t})$, the gradient of the log-ratio of the two noisy marginals at timestep $t$, so Eq.~\ref{eqn:contrastive_gradient_guidance} is the score of
\begin{equation}
    q_t \propto p_{\mathrm{anc},t} \prod_i \left( \rho_{i,t} / \kappa_{i,t} \right)^{\gamma_i},
\end{equation}
which noises each component first and composes them afterwards.
Sampling from $p^\ast$ by reverse diffusion instead requires the score of $(p^\ast)_t$, which composes the clean components first and noises the result afterwards.
The two orders agree only at $t = 0$: forward noising is a convolution with a Gaussian, and convolution does not commute with products or ratios of densities, so $q_t \neq (p^\ast)_t$ for $t > 0$~\citep{DY2023_ICML_ReduceReuseRecycle,MS2025_ICML_FKCorrect}.
Consequently, $\{q_t\}$ is not the noising path of $p^\ast$, and sampling with Eq.~\ref{eqn:contrastive_gradient_guidance} is not guaranteed to return samples from $p^\ast$, even when every component score is exact.
Correction steps such as MCMC~\citep{DY2023_ICML_ReduceReuseRecycle} or Feynman--Kac correctors~\citep{MS2025_ICML_FKCorrect} can reduce this mismatch.
In this work, we empirically design the sampler using Eq.~\ref{eqn:contrastive_gradient_guidance} instead of these techniques and leave obtaining effective, accurate denoising steps as future work.

\paragraph{Special cases. BT-induced pairs and existing guidance}
The pairs whose implicit reward carries a human-preference interpretation are those produced by preference optimization.
RLHF and DPO fine-tune a diffusion model toward the reward-tilted optimum in Eq.~\ref{eqn:optimal_solution}~\citep{CP2017_NeurIPS_RLHF,RR2023_NeurIPS_DPO}, that we call such a checkpoint a \emph{BT-induced diffusion model} $p_\mathrm{BT}$ with score $s_\mathrm{BT}(\vc, \vx_t, t)$.
At the optimum, Eq.~\ref{eqn:optimal_solution} gives $\log \big( p_\mathrm{BT}(\vx \vert \vc) / p_\mathrm{ref}(\vx \vert \vc) \big) = r_\mathrm{BT}(\vc, \vx) / \beta - \log Z(\vc)$, so the pair $(p_\mathrm{BT}, p_\mathrm{ref})$ has implicit reward $r_\mathrm{BT} / \beta$ up to a constant, and with $N = 1$, $\kappa_1 = p_\mathrm{anc} = p_\mathrm{ref}$, and $\gamma_1 = 1$, Eq.~\ref{eqn:cgg_objective} is the RLHF objective of Eq.~\ref{eqn:rlhf_objective} divided by $\beta$.
Treating the tilted relation as holding at the noisy marginals of timestep $t$, toward which the per-timestep surrogate objective of Diffusion-DPO pushes every noise level~\citep{WB2024_CVPR_DiffusionDPO}, yields the timestep reward gradient
\begin{equation}
    \label{eqn:implicit_reward_gradient}
    \nabla_{\vx_t} r_t(\vc, \vx_t) = \beta \left( s_\mathrm{BT}(\vc, \vx_t, t) - s_\mathrm{ref}(\vc, \vx_t, t) \right),
\end{equation}
available in closed form with no reward network and no denoised reconstruction, and Eq.~\ref{eqn:contrastive_gradient_guidance} reduces to the linear form
\begin{equation}
    \label{eqn:CGG-CFG}
    s_\mathrm{guided}(\vc, \vx_t, t) = (1 - \omega)\, s_\mathrm{ref}(\vc, \vx_t, t) + \omega\, s_\mathrm{BT}(\vc, \vx_t, t),
\end{equation}
with $\omega = \gamma_1$, or equivalently $\omega = \gamma\beta$ in the reward units of Eq.~\ref{eqn:gradient_guidance}.
Eq.~\ref{eqn:CGG-CFG} explains why an aligned checkpoint can serve as a test-time guidance direction rather than only as a standalone generator, and it recovers RLG~\citep{LJ2026_ICML_RLG}.
Practical checkpoints only approximate the reward-tilted optimum, so we read Eq.~\ref{eqn:implicit_reward_gradient} as motivation and treat the score difference of a trained checkpoint as its \emph{empirical implicit preference direction}.
For pairs outside this case, such as $\mathrm{SFT}^{+}/\mathrm{SFT}^{-}$ pairs, reference-free preference models, and style-biased checkpoints, Eq.~\ref{eqn:contrastive_form} is still a well-defined likelihood-ratio reward and Eqs.~\ref{eqn:cgg_objective}--\ref{eqn:contrastive_gradient_guidance} apply unchanged, but the reward carries no BT interpretation. We evaluate these pairs in Section~\ref{sec:experiments}.

The same reduction places existing guidance methods within Eq.~\ref{eqn:contrastive_gradient_guidance} as particular assignments of the three roles, summarized in Table~\ref{tab:guidance_taxonomy}.
Our formulation also corresponds to the score-space expression of a product of experts~\citep{HG2002_NC_PoE} and recurs in CFG, Composable Diffusion, Autoguidance, and RLG~\citep{HJ2022_CFG,LN2022_ECCV_ComposableDiffusions,KT2024_NeurIPS_autoguide,LJ2026_ICML_RLG}.
CFG and Composable Diffusion take the contrast from \emph{conditioning}: all roles are one network under different conditions, so a negated concept must be expressible through the prompt channel.
Autoguidance, RLG, Diffusion Blend, and cPGD take it from a \emph{training lineage}: the unfavored side is the favored model's own reference, its degraded twin, or a module trained on the negative side of the same preference dataset, and the anchor coincides with the reference~\citep{KT2024_NeurIPS_autoguide,LJ2026_ICML_RLG,CM2025_arXiv_DiffusionBlend,JZ2026_arXiv_cPGD}.
DExperts and Model Arithmetic compose signed differences of \emph{free checkpoints} at decoding time, but for autoregressive language models, where every step exposes an exact next-token distribution and no noisy-marginal approximation arises~\citep{LA2021_ACL_DExperts,DJ2024_ICLR_ModelArithmetic}.

CGG's contribution is the distributional objective in Eq.~\ref{eqn:cgg_objective} for flexibly reusing existing diffusion models: the anchor, the favored model, and the unfavored model of each pair are chosen independently, the unfavored side may be an explicit anti-expert rather than the anchor, and several pairs act at once.
The sampling algorithm that implements Eq.~\ref{eqn:contrastive_gradient_guidance}, its per-step cost, the calibration of the pair scales $\gamma_i$, and the three assumptions that underlie the method (checkpoint compatibility, a well-posed target, and approximate sampling) are collected in Appendix~\ref{sec:A_algorithm}.

\begin{table}[t]
    \centering
    \resizebox{\linewidth}{!}{%
        {\small
\setlength{\tabcolsep}{4pt}
\begin{tabular}{p{2.6cm}p{3.4cm}p{5.0cm}p{5.4cm}}
\toprule
Contrast source & Methods & Guidance form & Locked vs.\ free components \\
\midrule
Conditioning & CFG~\citep{HJ2022_CFG} & $(1-\omega)\, s(\vx_t,t) + \omega\, s(\vc, \vx_t,t)$ & Pair $=$ one network under two conditions; negation must pass through the prompt channel \\
\addlinespace[3pt]
Conditioning, multi-concept & Composable Diffusion~\citep{LN2022_ECCV_ComposableDiffusions} & $s_\mathrm{anc} + \sum_{j} \omega_j\, \big(s(\vc_j^{+}, \vx_t,t) - s(\vc_j^{-}, \vx_t,t)\big)$ & All roles $=$ one network under different conditions; signed $\omega_j$; anchor is a conditional or unconditional branch of the same network \\
\addlinespace[3pt]
Training lineage & Autoguidance~\citep{KT2024_NeurIPS_autoguide}, RLG~\citep{LJ2026_ICML_RLG}, Diffusion Blend~\citep{CM2025_arXiv_DiffusionBlend} & $s_\mathrm{ref} + \omega\,(s_\mathrm{RL} - s_\mathrm{ref})$ & $\kappa$ fixed to $\rho$'s own reference or degraded twin; anchor $=\kappa$; amplify-only \\
\addlinespace[3pt]
Training lineage, paired modules & cPGD~\citep{JZ2026_arXiv_cPGD} & $s_\mathrm{base} + \omega\,(s_\mathrm{pos} - s_\mathrm{neg})$ & $\rho$, $\kappa$ trained on the two sides of one preference dataset; anchor $=$ base; one pair \\
\addlinespace[3pt]
Free checkpoint, language models & DExperts~\citep{LA2021_ACL_DExperts}, Model Arithmetic~\citep{DJ2024_ICLR_ModelArithmetic} & $z_\mathrm{base} + \sum_{i} \lambda_i\, (z_{\rho_i} - z_{\kappa_i})$ on next-token logits & Roles decoupled; exact per-token conditionals, so no noisy-marginal approximation; autoregressive only \\
\addlinespace[3pt]
Free checkpoint, diffusion & CGG (ours) & $s_\mathrm{anc} + \sum_{i} \gamma_i\, (s_{\rho_i} - s_{\kappa_i})$ & Anchor, $\rho_i$, $\kappa_i$ all decoupled; signed $\gamma_i$; $N$ simultaneous pairs; sampling approximate (Eq.~\ref{eqn:contrastive_gradient_guidance}) \\
\bottomrule
\end{tabular}
}
    }
    \caption{Guidance methods as assignments of the three roles in Eq.~\ref{eqn:contrastive_gradient_guidance}. All share the linear anchor-plus-difference form. These works differ in where the contrast comes from and in which roles are locked. CGG frees all three roles for diffusion checkpoints, admits signed pair scales, and composes several pairs simultaneously.}
    \label{tab:guidance_taxonomy}
\end{table}

\section{Experiments}
\label{sec:experiments}

We organize the empirical study around three claims that mirror the design of CGG.
First, score differences between SFT experts trained on $1000$ examples each steer generation toward or away from a target attribute.
Second, reusable experts and anti-experts preserve a balanced multi-preference trade-off across text-image preferences, synthetic concept control, photographic fidelity, and safety.
Third, CGG can be paired with sampling-based reward steering, allowing model-based and reward-based guidance to share one sampling trajectory.

\paragraph{Datasets and base models}
We mainly assess CGG on $500$ unique prompts from the Pick-a-Pic v2 test set~\citep{KY2023_NeurIPS_PickaPic}, where several preference optimization methods are available for Stable Diffusion XL (SDXL)~\citep{DP2024_ICLR_SDXL}, including Diffusion-DPO~\citep{WB2024_CVPR_DiffusionDPO}.
We use MaPO checkpoints for safety-constrained scenarios~\citep{JH2024_MaPO}.
Unless otherwise stated, the base model $s_\mathrm{base}$ denotes the pre-trained SDXL, which is also the reference model $s_\mathrm{ref}$ in the previous sections.
Checkpoint details, sampling settings, reward definitions, and additional detailed results are reported in Appendix~\ref{sec:B_detailed_exps}.

\subsection{Limited-Data SFT as Preference Signals}
\label{sec:flexible-preference-signals}
\begin{figure}[t]
    \begin{center}
        \begin{minipage}[t]{0.49\linewidth}
            \centering
            \begingroup
            \setlength{\tabcolsep}{1.5pt}
            \renewcommand{\arraystretch}{0.92}
            \definecolor{cggpurple}{RGB}{112,48,160}
            \scriptsize
            \begin{tabular}{@{}c c c c c@{}}
                & $\boldsymbol{s_\mathrm{base}}$
                & $\boldsymbol{s_\rho}$
                & $\boldsymbol{s_\kappa}$
                & $\boldsymbol{s_\mathrm{DPO}}$ \\
                \textbf{\textcolor{red}{Red}} &
                \includegraphics[width=0.205\linewidth]{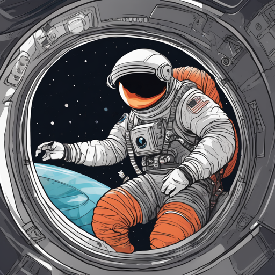} &
                \includegraphics[width=0.205\linewidth]{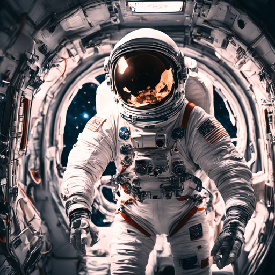} &
                \includegraphics[width=0.205\linewidth]{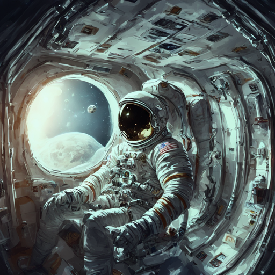} &
                \includegraphics[width=0.205\linewidth]{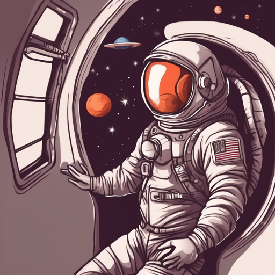} \\
                \textbf{\textcolor{cggpurple}{Purple}} &
                \includegraphics[width=0.205\linewidth]{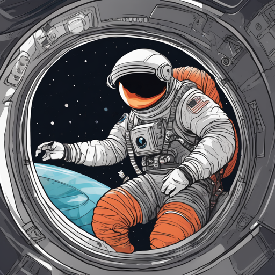} &
                \includegraphics[width=0.205\linewidth]{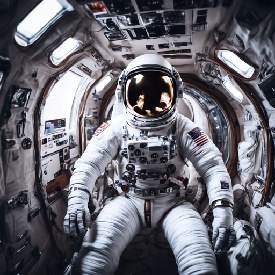} &
                \includegraphics[width=0.205\linewidth]{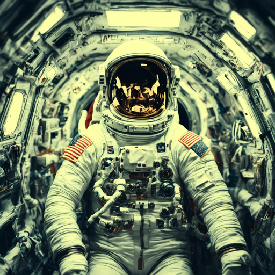} &
                \includegraphics[width=0.205\linewidth]{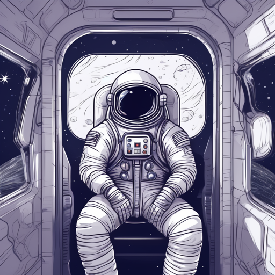}
            \end{tabular}
            \endgroup
        \end{minipage}
        \hfill
        \begin{minipage}[t]{0.49\linewidth}
            \centering
            \begingroup
            \setlength{\tabcolsep}{2pt}
            \renewcommand{\arraystretch}{0.92}
            \definecolor{cggpurple}{RGB}{112,48,160}
            \scriptsize
            \begin{tabular}{@{}c c c c@{}}
                &
                \shortstack{\textbf{FK}\\$(s_\mathrm{base},\,\mathrm{Color})$} &
                \shortstack{\textbf{CGG}\\$\rho$-base} &
                \shortstack{\textbf{CGG}\\$\rho$-$\kappa$} \\
                \textbf{\textcolor{red}{Red}} &
                \includegraphics[width=0.2\linewidth]{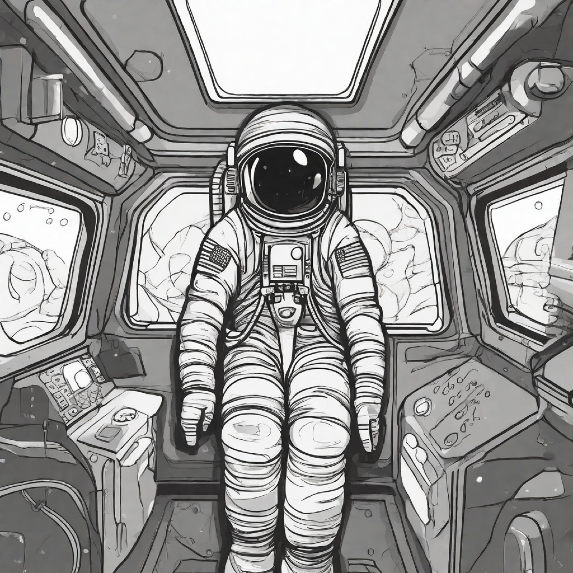} &
                \includegraphics[width=0.2\linewidth]{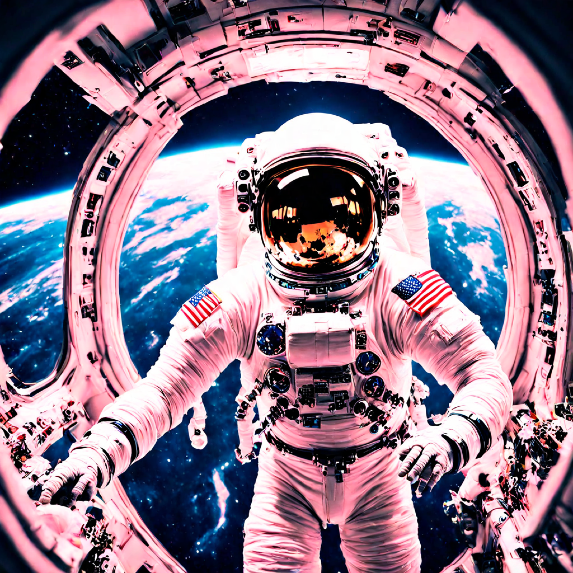} &
                \includegraphics[width=0.2\linewidth]{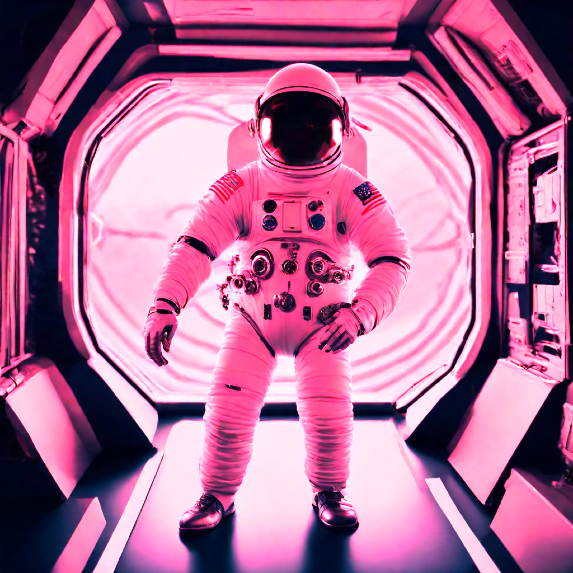} \\
                \textbf{\textcolor{cggpurple}{Purple}} &
                \includegraphics[width=0.2\linewidth]{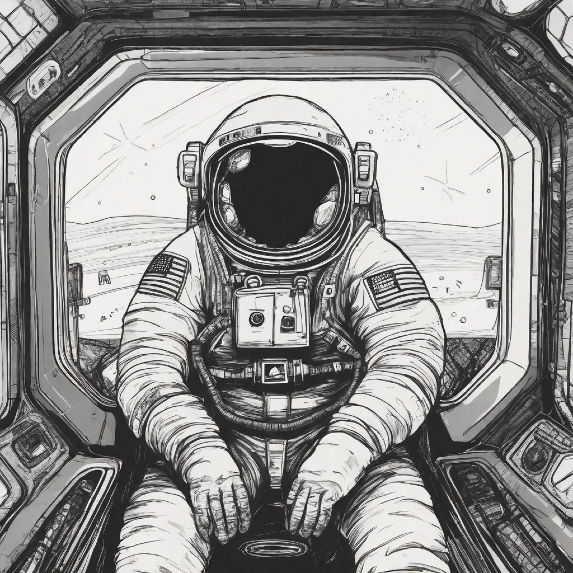} &
                \includegraphics[width=0.2\linewidth]{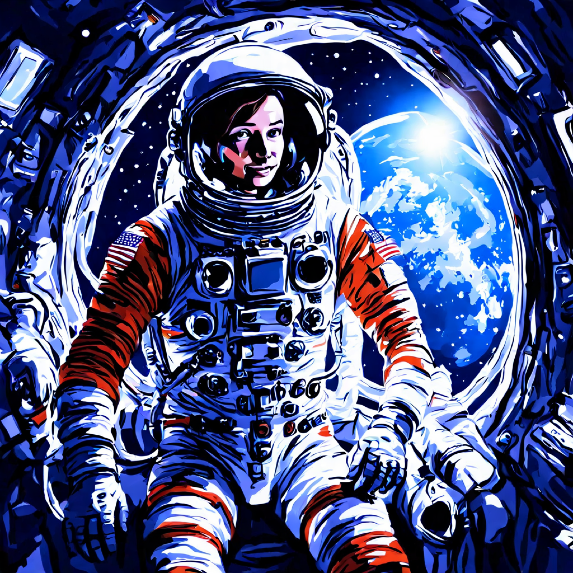} &
                \includegraphics[width=0.2\linewidth]{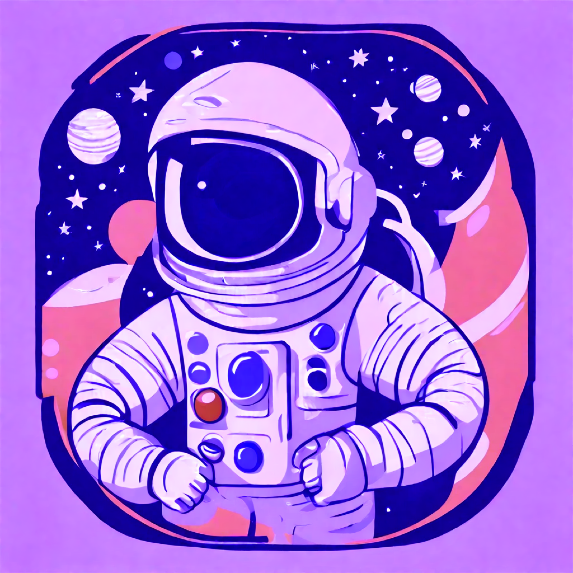}
            \end{tabular}
            \endgroup
        \end{minipage}
    \end{center}
    \caption{Limited-data color preference control shows that CGG can inject colors more directly than preference fine-tuning or reward steering. Left: SFT and DPO fine-tuning under Red and Purple preferences. Right: FK Steering with a trained color reward compared with CGG using SFT-derived favored and unfavored diffusion experts, where $\rho$ denotes the favored side and $\kappa$ the unfavored side. Prompt: `astronaut in spaceship'.}
    \label{fig:color-limited_scenario}
\end{figure}

We first test contrastive directions built from SFT experts trained on little data.
Starting from only $1000$ examples per color preference, we fine-tune SFT experts for Red, NonRed, Purple, and NonPurple preferences.
Figure~\ref{fig:color-limited_scenario} compares standalone SFT/DPO-style fine-tuning, FK Steering with a trained color reward, and CGG with SFT-based favored/unfavored pairs.
The qualitative comparison shows that CGG can inject the target color more directly than using a preference fine-tuned model or reward-steered samples, supporting the claim that the SFT positive/negative split can be reused as a test-time contrastive direction.

\begin{figure}[t]
    \begin{center}
        \includegraphics[width=0.7\linewidth]{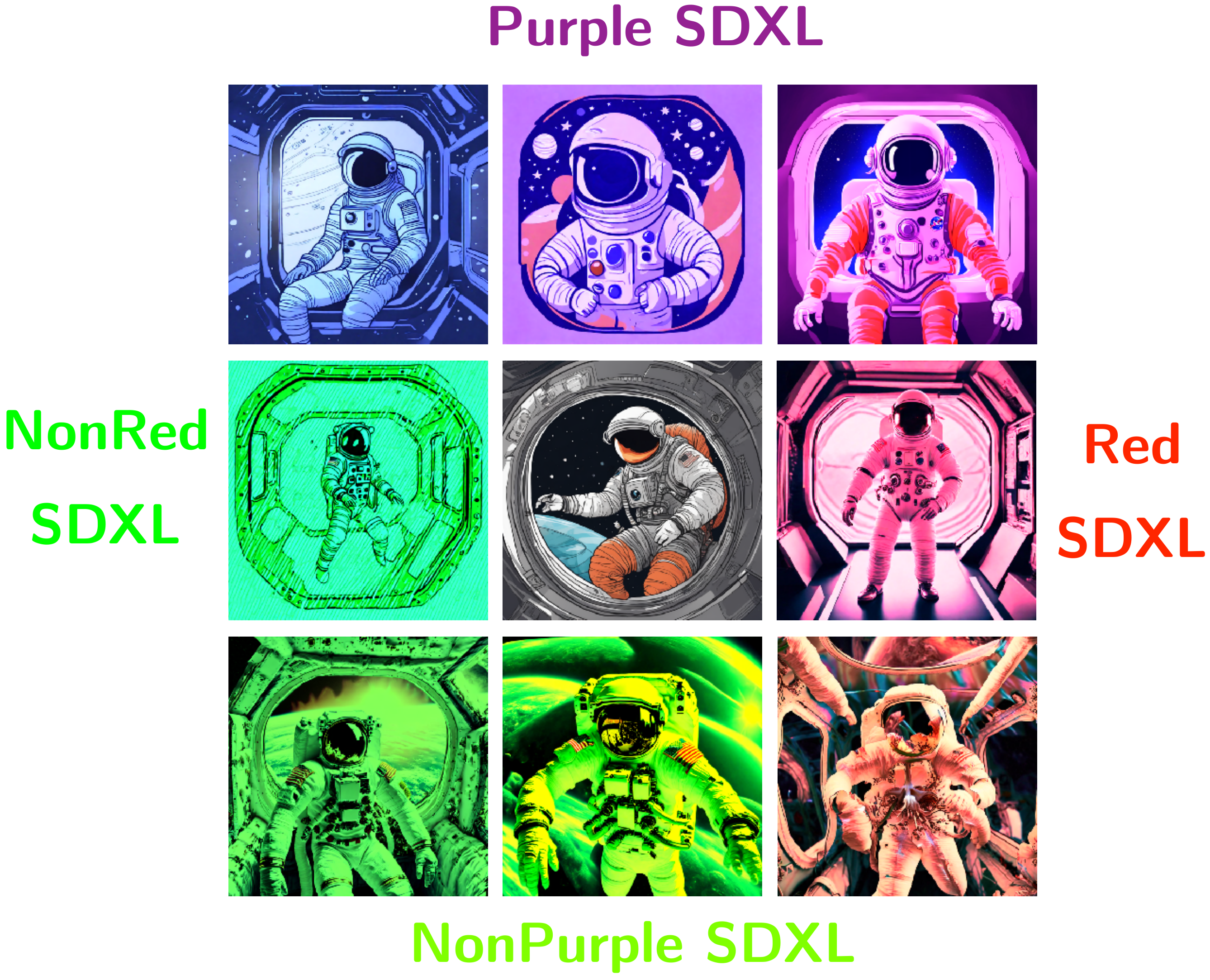}
    \end{center}
    \caption{The 2D Red-Purple color palette grid (prompt: `astronaut in spaceship'). The central image is the base model $s_\mathrm{base}$. The horizontal axis applies guidance from the Red-preference expert and the vertical axis from the Purple-preference expert; negative directions reduce the color.}
    \label{fig:color-palette}
\end{figure}

Figure~\ref{fig:color-palette} shows the same controllability pattern in the palette setting (quantified in Table~\ref{tab:color-palette}).
Moving along the Red and Purple contrastive directions increases the intended color score, while combining directions allows CGG to traverse intermediate preference mixtures without training a new model for each palette point.

We next evaluate whether SFT-derived contrastive signals remain effective on the Pick-a-Pic benchmark.
Table~\ref{tab:Avg_results-CGG-Pos-Neg} groups the comparison into pre-trained, fine-tuned, and test-time methods.
It shows that $s_\mathrm{SFT}^{+}-s_\mathrm{SFT}^{-}$ reaches a PickScore on par with the DPO fine-tuned model and the best CLIP score in the table, so CGG with SFT experts reaches DPO-level preference quality without pairwise preference optimization.
Appendix~\ref{sec:detail_CGG_PO} reports the same comparison for pairs built from DPO, MaPO, KTO, and CHATS checkpoints.
\begin{table}[t]
    \begin{center}
        {\setlength{\tabcolsep}{2pt}
\begin{tabular}{@{}lccc@{}}
\toprule
Class / Method & PickScore ($\uparrow$) & HPSv2 ($\uparrow$) & CLIP ($\uparrow$) \\
\midrule
\multicolumn{4}{l}{\textbf{Pre-trained}} \\
\quad $s_\mathrm{base}$ & 22.11 & 0.29 & 36.99 \\
\midrule
\multicolumn{4}{l}{\textbf{Fine-tuned}} \\
\quad $s_\mathrm{DPO}$ & 22.41 & 0.30 & 37.78 \\
\quad $s_\mathrm{SFT}^+$ & 22.34 & \textbf{0.31} & 37.20 \\
\midrule
\multicolumn{4}{l}{\textbf{Test-time steering}} \\
\quad CGG: SFT$^+$--base & 22.33 & \textbf{0.31} & 37.52 \\
\quad CGG: SFT$^+$--SFT$^-$ & \textbf{22.43} & 0.30 & \textbf{37.83} \\
\bottomrule
\end{tabular}
}

    \end{center}
    \caption{Average rewards on $500$ Pick-a-Pic v2 test prompts, grouped by pre-trained, fine-tuned, and test-time (CGG) methods. In CGG rows, A-B denotes the favored-unfavored pair used in Eq.~\ref{eqn:contrastive_gradient_guidance}, with the base model as anchor. \textbf{Bold text} marks the best result.}
    \label{tab:Avg_results-CGG-Pos-Neg}
\end{table}

\subsection{Multi-Preference Alignment with Reusable Experts}
\label{sec:multi-preference-results}
We then evaluate the multi-preference setting that motivates Figure~\ref{fig:safety_qualitative_comparisons} and Figure~\ref{fig:five_reward_method_comparison_radar}.
The goal is not to maximize one scalarized reward at training time, but to preserve a favorable trade-off among several axes after deployment.
CGG (5-pair combinations) composes five contrastive pairs, one per preference axis: the Red and realism axes take an explicit anti-expert as their unfavored side, while the remaining three pair an expert against the pre-trained base model, with the exact checkpoints listed in Appendix~\ref{sec:B_detailed_exps}.
All five pair scales are set to $\gamma_i = 1.0$ for all prompts, so composing the five preferences involves no scale search.
Single-pair $\gamma$-sensitivity is smooth and unimodal (Appendix~\ref{sec:C_more_gamma_sensitivity_experiments}), and a different balance can be obtained by changing the pair scales at sampling time.

In a multi-preference setup, we compare CGG with two baselines: FK Steering with five averaged rewards and Task Arithmetic, which merges models' parameters.
Task Arithmetic composes the five corresponding checkpoint-space task vectors with unit coefficients, each measured relative to the SDXL base model.
Both methods therefore use fixed unit coefficients; other global scales for Task Arithmetic are reported in Appendix~\ref{sec:B_detailed_exps}.

Table~\ref{tab:five_reward_method_comparison_raw} provides the full numerical record, grouping methods into base, offline fine-tuned models, and test-time steering methods.
The base model is reasonably safe but not the strongest on PickScore or ImageReward.
DPO improves PickScore but reduces safety, while MaPO gives the firmest safety score but has low Realness, consistent with the anime-style side effect shown in Figure~\ref{fig:safety_qualitative_comparisons}.
CGG is not the best method on every axis: it has the best ImageReward and the second-best PickScore, Safety, and Realness.
It is, however, the only method in the table that improves over the base model on all five axes; every fine-tuned expert and every other test-time method falls below the base model on at least one axis (red entries).
Confidence intervals and an independent re-run are reported in Appendix~\ref{sec:B_detailed_exps}.

\paragraph{Avoiding a protected artistic style at test time}
We next consider style suppression: reusing an existing style-specific model to prevent generated images from reproducing that style, a setting motivated by intellectual-property protection.
Existing approaches either use a text negative prompt at sampling time~\citep{PS2023_CVPR_SLD} or erase the concept by permanently editing model weights once per style~\citep{GO2024_WACV_UCE}.
CGG differs by reusing a public style checkpoint as a model-level anti-expert: it suppresses the style in gradient space at sampling time, reversibly and with a continuous strength $\gamma$.
We use public Studio Ghibli and van Gogh SDXL LoRAs as the unfavored side, with pre-trained SDXL as both the anchor and the favored side (detailed information in Appendix~\ref{sec:style_avoidance_exps_detail}).
The prompts are $24$ Studio Ghibli and $20$ van Gogh scenes, half of each naming the style explicitly.
Style is measured by CSD similarity~\citep{SG2024_arXiv_CSD} to the style expert's own outputs, which measures style independently of content and cannot be optimized directly by a negative prompt that lists the same words.

This setup asks whether a method preserves the requested image while suppressing a protected style, or whether it removes the style by changing the image too much.
Table~\ref{tab:style_avoidance} reports two quantities: \emph{Style removed}, which measures how much the target style is suppressed, and \emph{Photo margin}, which measures whether the sample shifts from illustration toward photography.
Table~\ref{tab:style_avoidance} shows that negative prompting and CGG can achieve comparable style suppression. On Studio Ghibli, however, the negative prompt flips the photo margin from $-1.78$ to $+1.92$, whereas CGG at the same suppression level keeps it at $-2.10$, preserving the illustrative form rather than drifting toward photography; on van Gogh, neither method shifts the medium toward photography. Detailed experimental configurations, prompts, and evaluation procedures are provided in Appendix~\ref{sec:style_avoidance_exps_detail}.
\begin{table}[t]
\begin{center}
    \begin{tabular}{@{}lcc@{}}
\toprule
\multicolumn{1}{l}{Style / Method} & Style removed ($\uparrow$) & Photo margin \\
\midrule
\multicolumn{3}{l}{\textbf{Studio Ghibli}} \\
\quad Base & 0.000 & -1.78 \\
\quad + negative prompt & 0.090$^\diamond$ & +1.92 \\
\quad + CGG ($\gamma=0.7$) & 0.090$^\diamond$ & -2.10 \\
\quad + CGG ($\gamma=1$) & 0.159 & -2.20 \\
\midrule
\multicolumn{3}{l}{\textbf{van Gogh}} \\
\quad Base & 0.000 & -0.08 \\
\quad + negative prompt & 0.325$^\diamond$ & -0.52 \\
\quad + CGG ($\gamma=3.1$) & 0.325$^\diamond$ & -0.42 \\
\quad + CGG ($\gamma=3.5$) & 0.359 & -0.87 \\
\bottomrule
\end{tabular}

\end{center}
\caption{Avoiding a protected artistic style on prompts that explicitly name it. $^\diamond$ marks a CGG row with the same Style removed level as the negative prompt.}
\label{tab:style_avoidance}
\end{table}

\begin{figure}[t]
    \begin{center}
        \includegraphics[width=0.85\linewidth]{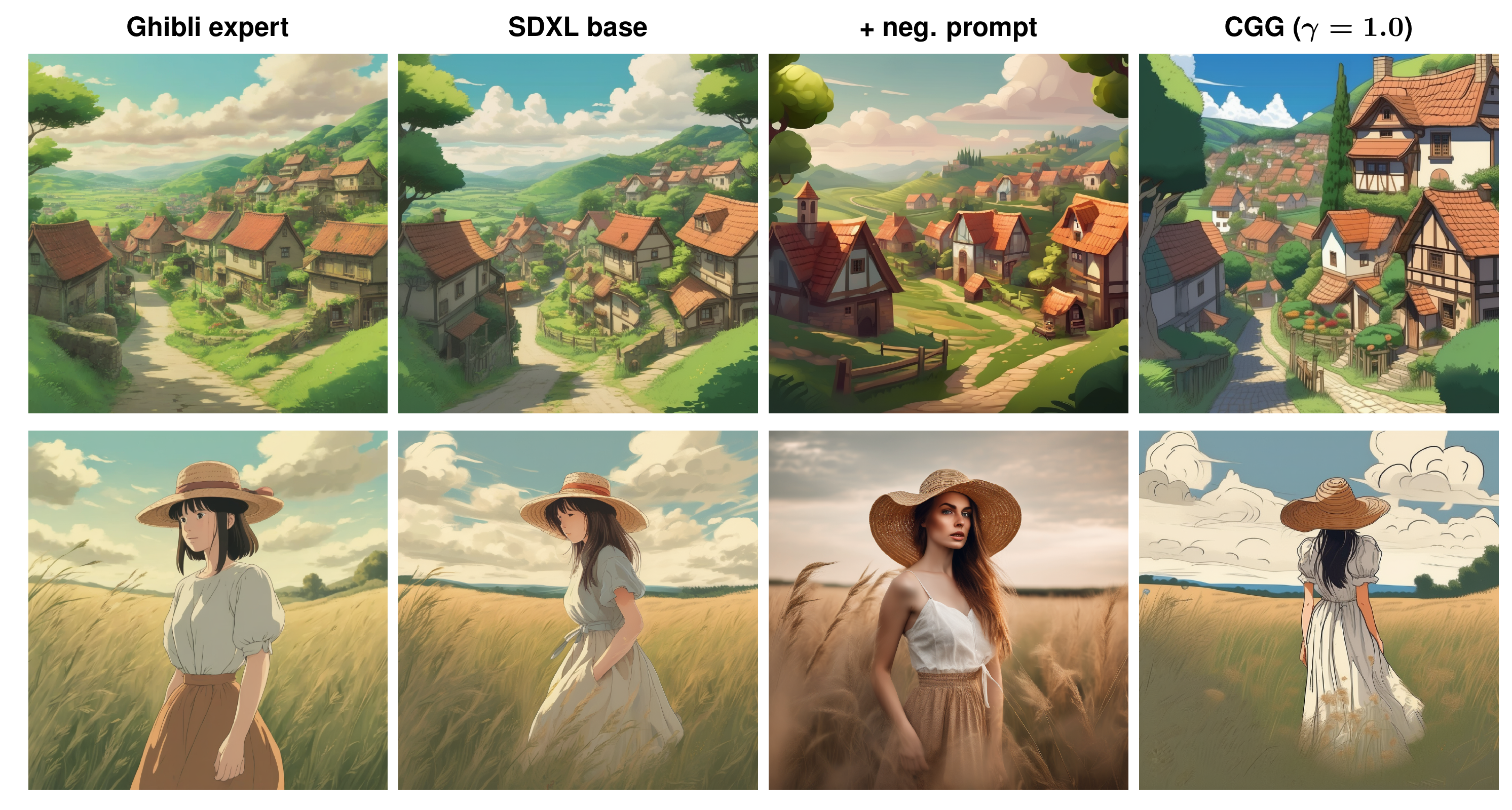}
    \end{center}
    \caption{Suppressing a protected style at test time on prompts explicitly requesting it. Columns compare the Ghibli expert, SDXL base model, text negative prompt, and CGG ($\gamma=1.0$) with the style expert as anti-expert. All samples share the prompt and seed. The full 6-method comparison across varying scales $\gamma$ is shown in Figure~\ref{fig:style_avoidance_ghibli_full} in the Appendix.}
    \label{fig:style_avoidance_ghibli}
\end{figure}

Figure~\ref{fig:style_avoidance_ghibli} shows the qualitative comparison on two style-invocation prompts.
While the negative prompt re-renders the scene as a photograph, CGG continuously suppresses the target style while preserving the underlying scene composition and remaining inside the illustrated medium.
A complete comparison across varying pair scales $\gamma$ is provided in Figure~\ref{fig:style_avoidance_ghibli_full} in the Appendix.
Figure~\ref{fig:style_avoidance_three_settings} in the Appendix repeats the base, negative-prompt, and anti-expert comparison at $\gamma=1$ on four further scenes from the same style-invocation set.

\subsection{Compatibility with Sampling-Based Steering}
\label{sec:cgg-fk-results}
\begin{figure}[t]
    \begin{center}
        \begingroup
        \setlength{\tabcolsep}{4pt}
        \renewcommand{\arraystretch}{0.92}
        \scriptsize
        \begin{tabular}{@{}c c c c@{}}
            $\boldsymbol{s_\mathrm{base}}$ &
            \shortstack{\textbf{FK}\\$(\mathrm{PickScore})$} &
            \shortstack{\textbf{CGG}\\Red-NonRed} &
            \shortstack{\textbf{CGG-FK}\\Red-NonRed + Pick} \\
            \includegraphics[width=0.2\linewidth]{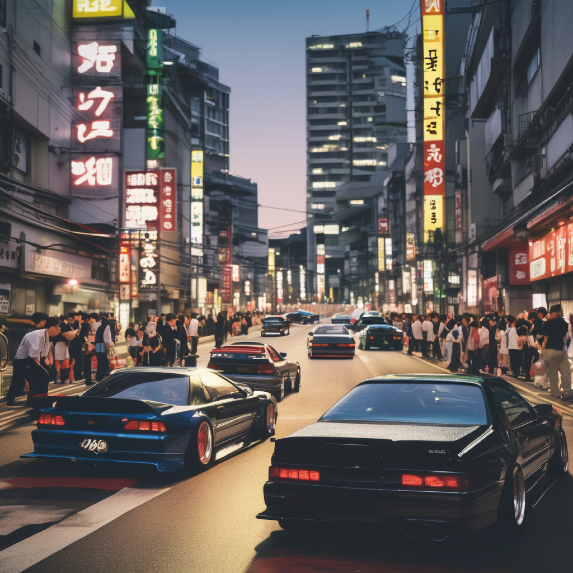} &
            \includegraphics[width=0.2\linewidth]{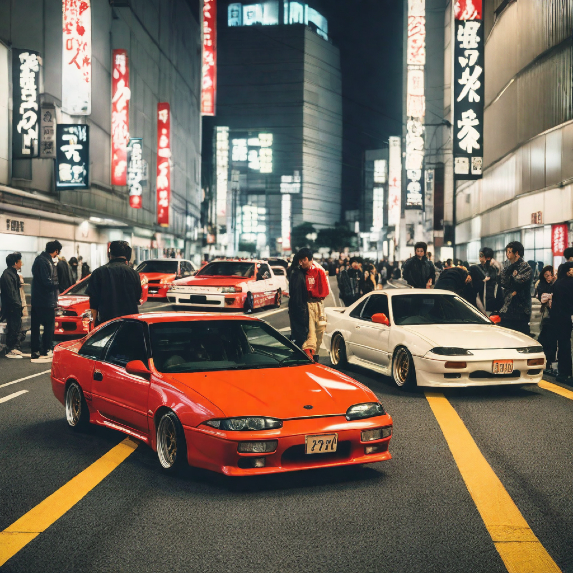} &
            \includegraphics[width=0.2\linewidth]{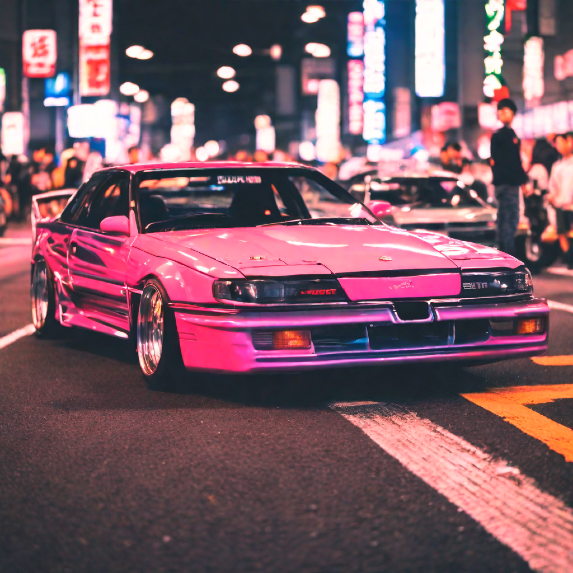} &
            \includegraphics[width=0.2\linewidth]{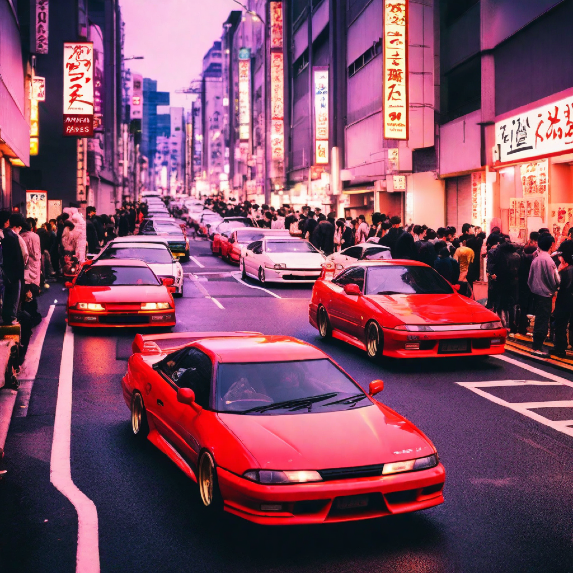} \\
            \includegraphics[width=0.2\linewidth]{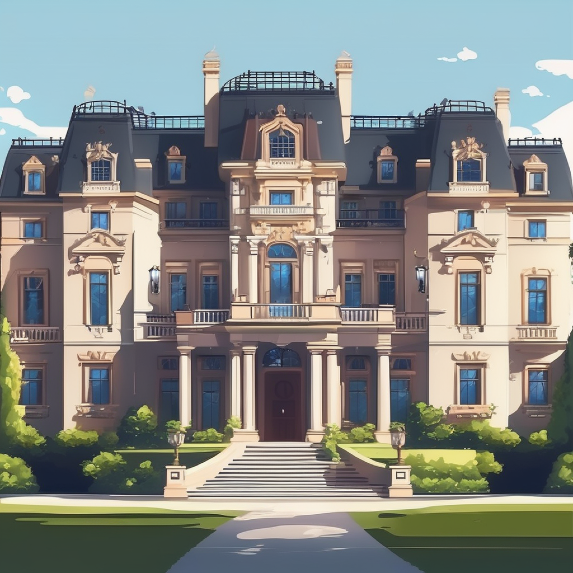} &
            \includegraphics[width=0.2\linewidth]{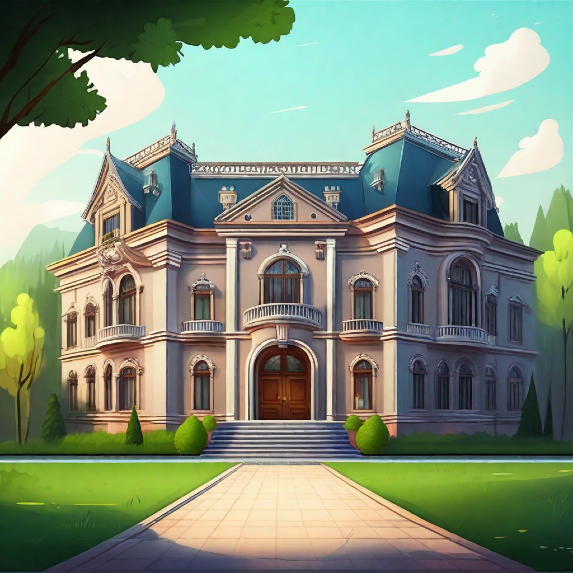} &
            \includegraphics[width=0.2\linewidth]{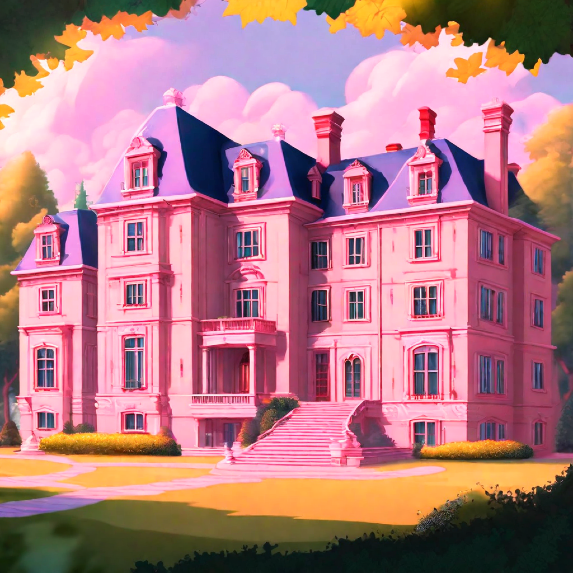} &
            \includegraphics[width=0.2\linewidth]{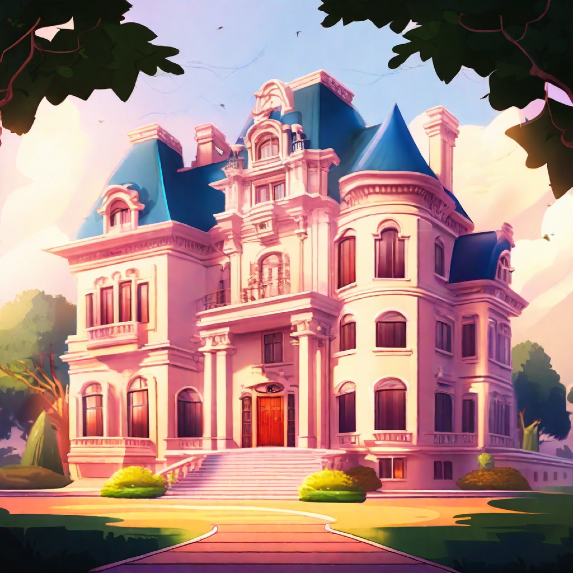}
        \end{tabular}
        \endgroup
    \end{center}
    \caption{CGG can be combined with FK Steering to inject an attribute-specific Red preference while preserving reward-based PickScore alignment. The comparison uses the same Pick-a-Pic style prompts across $s_\mathrm{base}$, FK(PickScore), CGG with the Red-NonRed pair, and CGG-FK, illustrating that the model-based color direction and the reward-based quality correction can share one sampling trajectory. Prompts: top row, `90's street racing in tokyo'; bottom row, `big mansion in the daytime'.}
    \label{fig:CGG_Red_Ref-FK_PickScore}
\end{figure}

Finally, we study CGG's compatibility with sampling-based reward steering.
The scenario is to inject a Red concept into the image while preserving general human-preference quality.
CGG supplies a model-based Red-vs-NonRed contrastive direction, and FK Steering can still apply an explicit PickScore reward correction during sampling~\citep{RS2025_ICML_FKSteering}.
Table~\ref{tab:cgg_fk_pick_red} shows that two guidance sources compose rather than interfere: the model-based direction carries the attribute, and the reward-based correction preserves preference quality.
Figure~\ref{fig:CGG_Red_Ref-FK_PickScore} illustrates the same setting qualitatively.
\begin{table}[t]
    \begin{center}
        \begin{tabular}{c|cc}
                         & PickScore & RedScore \\
\hline
$s_\mathrm{base}$ & 22.11  & 5.91 \\
FK(Pick) &    22.21 & 5.61 \\
CGG: \textcolor{red}{Red}--NonRed & 21.85 & 6.80 \\
CGG-FK: \textcolor{red}{Red}--NonRed + Pick & \textbf{22.26} & \textbf{7.02}
\end{tabular}

    \end{center}
    \caption{Combining model-based CGG with reward-based FK Steering on Pick-a-Pic v2 test prompts. FK(Pick) uses a PickScore reward; Red-NonRed denotes the CGG favored-unfavored pair that injects an attribute-specific contrastive direction. \textbf{Bold text} ($\uparrow$) marks the best result.}
    \label{tab:cgg_fk_pick_red}
\end{table}

\section{Related Works}
\label{sec:related-works}

An extended related work is provided in Appendix~\ref{sec:extended_related_work}.

\paragraph{Favored and unfavored distributions for preference modeling}
While standard alignment methods focus primarily on scalar reward modeling or policy optimization on favored data~\citep{CP2017_NeurIPS_RLHF,RR2023_NeurIPS_DPO,WB2024_CVPR_DiffusionDPO}, recent studies model the unfavored (anti-expert) distribution explicitly (e.g., Diffusion-NPO~\citep{FW2025_ICLR_Diffusion_NPO}, CHATS~\citep{MF2025_ICML_CHATS}).
In language generation, DExperts subtracts logit distributions of undesirable corpora during decoding~\citep{LA2021_ACL_DExperts}.
However, existing diffusion preference methods operate at the \emph{data and loss level}: they modify the optimization objective and retrain on raw preference datasets whenever the target trade-off changes.
CGG operates at the \emph{model and inference level}: it repurposes off-the-shelf fine-tuned diffusion models directly as anti-experts and composes them at test time.

\paragraph{Test-time alignment for diffusion models}
Test-time alignment steers sampling by estimating time-dependent reward gradients $\nabla_{\vx_t} r_{\text{test}}(\vc, \vx_t)$, either by evaluating clean image estimates $\hat{\mathbf{x}}_0$~\citep{BA2023_CVPR_UniversalGuide} or propagating particle populations via Sequential Monte Carlo (SMC) sampling~\citep{SK2025_ICLR_DAS,RS2025_ICML_FKSteering}.
CGG is compatible with these approaches: it changes the score that proposes each denoising step, while they change how samples are weighted or selected.
Section~\ref{sec:cgg-fk-results} combines CGG with FK Steering so that a model-based contrastive direction and a reward-based correction act on the same sampling trajectory.

\paragraph{Compositional and test-time multi-preference alignment}
Existing methods frame the contrast to prompt-level conditioning (CFG) or within a fixed training lineage (e.g., contrasting a model against its reference in RLG and Diffusion Blend, or against a degraded twin in Autoguidance), which amplifies attributes encoded in an aligned expert~\citep{HJ2022_CFG,FP2025_arXiv_CFGRL,KT2024_NeurIPS_autoguide,LJ2026_ICML_RLG,CM2025_arXiv_DiffusionBlend}.
CGG generalizes these formulations to a free-checkpoint contrast: the anchor, favored, and unfavored models are chosen independently.
This freedom is what lets an off-the-shelf fine-tuned model serve as an anti-expert and several such pairs be composed at test time without retraining.

\section{Conclusion}
\label{sec:conclusion}
We presented Contrastive Gradient Guidance (CGG), which treats preference alignment as compositional generation.
Each preference is a pair of a favored expert and an unfavored anti-expert, and the score difference of the pair is a test-time guidance direction, so preferences can be added, removed, or reweighted at sampling time without retraining.
In our experiments, CGG composed five preferences with a single fixed scale per pair and gave a more balanced trade-off than offline fine-tuning, weight-space merging, and reward steering; the same mechanism enforced a safety constraint and suppressed a protected artistic style while keeping the scene the prompt asked for.

CGG evaluates every expert at every denoising step, so its latency and memory grow with the number of pairs (Appendix~\ref{sec:inference_overhead}).
Reducing this cost, for example by storing experts as adapters on a shared backbone, and extending CGG to preferences that change per user or per request are directions for future work.

\section*{Ethical Statement}
This work aims to improve test-time preference alignment for diffusion models, so that a deployed model can follow preferences that change after training.
Such systems can also amplify biases in preference data and reward models, so practical deployment should include careful evaluation across diverse user groups, transparent documentation, and safeguards against harmful or exclusionary uses.

\bibliographystyle{plainnat}
\bibliography{main}

\clearpage
\appendix

\section{Sampling algorithm, cost, pair scales, and assumptions}
\label{sec:A_algorithm}
This appendix collects the procedural side of CGG that Section~\ref{sec:cgg_overview} defers: the sampling algorithm that implements Eq.~\ref{eqn:contrastive_gradient_guidance}, its per-step cost, how the pair scales are set at test time, and the assumptions that underlie the method.

\paragraph{Sampling algorithm}
CGG is implemented by replacing the model score in each denoising step with Eq.~\ref{eqn:contrastive_gradient_guidance}; Algorithm~\ref{alg:cgg_sampling} summarizes the procedure, and any base sampler can be used.

\begin{algorithm}[ht]
\caption{Contrastive Gradient Guidance sampling}
\label{alg:cgg_sampling}
\begin{algorithmic}[1]
\STATE \textbf{Input:} anchor score $s_\mathrm{anc}$, contrastive preference pairs $\{(\rho_i,\kappa_i)\}_{i=1}^N$
\STATE \textbf{Input:} pair scales $\{\gamma_i\}_{i=1}^N$, prompt $\vc$
\STATE Sample initial noise $\vx_T$
\FOR{$t=T,\ldots,1$}
    \STATE Compute $s_\mathrm{anc}(\vc,\vx_t,t)$
    \STATE For each pair $i$, compute $s_{\rho_i}(\vc,\vx_t,t)$ and $s_{\kappa_i}(\vc,\vx_t,t)$, reusing $s_\mathrm{anc}$ when $\kappa_i = p_\mathrm{anc}$
    \STATE Form $s_\mathrm{CGG}$ using Eq.~\ref{eqn:contrastive_gradient_guidance}
    \STATE Update $\vx_{t-1}$ with the base sampler using $s_\mathrm{CGG}$
\ENDFOR
\STATE \textbf{return} $\vx_0$
\end{algorithmic}
\end{algorithm}

\paragraph{Cost}
Each step evaluates the anchor once and every distinct favored and unfavored model once, so a composition with $N$ pairs costs $N + 1 + \lvert \{ i : \kappa_i \neq p_\mathrm{anc} \} \rvert$ denoising-network evaluations per step when the anchor doubles as an unfavored side; the five-pair composition of Section~\ref{sec:multi-preference-results} uses eight.
Latency grows accordingly and peak memory grows with the number of distinct checkpoints held, which Appendix~\ref{sec:inference_overhead} quantifies against particle-based reward steering.

\paragraph{Pair scales}
Changing a preference, removing an anti-expert, or shifting a trade-off changes only the pairs and scales, not the models.
In the multi-preference experiments the scales are set to one and shared across all prompts.
For the single-pair preference experiments, $\gamma$ is selected on the Pick-a-Pic v2 validation set (Appendix~\ref{sec:C_more_gamma_sensitivity_experiments}); scales for the single-pair style experiments are tuned per style (Appendix~\ref{sec:B_detailed_exps}).
We apply the shared scale to the raw score differences without normalizing them.

\paragraph{Assumptions}
Three assumptions underlie the method.
(i) \emph{Checkpoint compatibility}: all models in Eq.~\ref{eqn:contrastive_gradient_guidance} share the latent space, noise schedule, prediction parameterization, conditioning interface, and CFG convention, so that their scores can be evaluated and combined at the same $(\vx_t, t)$; all experiments use SDXL-compatible checkpoints.
(ii) \emph{Well-posed target}: for each prompt $\vc$, the log ratios for pairs with $\gamma_i\ne0$ are defined almost everywhere under the anchor, and we assume $0<Z(\vc)<\infty$ in Eq.~\ref{eqn:cgg_optimal} and finite reward expectation and anchor KL at $p^\ast$. Appendix~\ref{sec:C_more_gamma_sensitivity_experiments} reports how the measured rewards vary with the guidance scale.
(iii) \emph{Approximate sampling}: Eq.~\ref{eqn:contrastive_gradient_guidance} follows the exact score of the noisy-marginal product $q_t$ rather than of $(p^\ast)_t$, and the BT reading of Eq.~\ref{eqn:implicit_reward_gradient} holds only for checkpoints at the tilted optimum at every noise level; results for all other pairs are empirical.

\clearpage
\section{Detailed experimental settings and results}
\label{sec:B_detailed_exps}

\begin{figure}[ht]
    \begin{center}
        \includegraphics[width=0.7\linewidth]{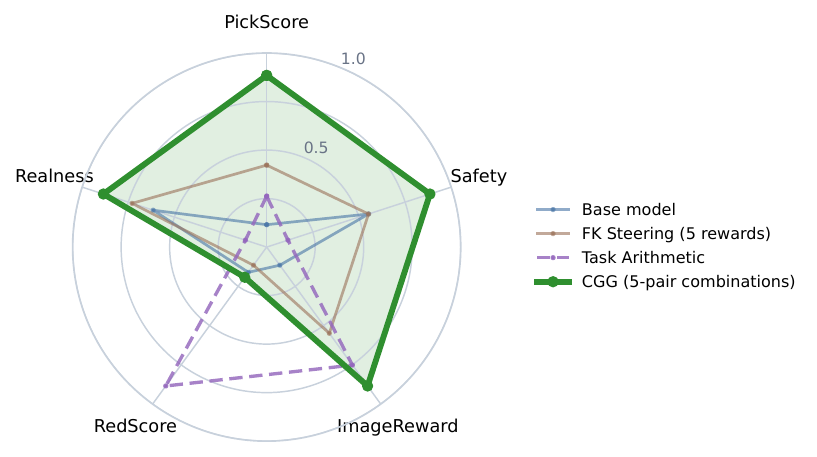}
    \end{center}
    \caption{Five-reward method comparison across text-image preferences (PickScore and ImageReward), synthetic concept control (RedScore), photographic fidelity (Realness), and Safety. Each radar axis is min-max normalized so outer-ring performance represents the best score across all methods in Table~\ref{tab:five_reward_method_comparison_raw}. CGG is the only method that stays at or outside the base model on all five axes and lies in the outer region on PickScore, Safety, ImageReward, and Realness, while its RedScore gain over the base model is small; Task Arithmetic gains on ImageReward and RedScore but falls below the base model on Safety and Realness.}
    \label{fig:five_reward_method_comparison_radar}
\end{figure}

\paragraph{Settings of sampling configurations for all experiments}
All experiments use $50$ denoising steps and a CFG scale of $7.5$ unless a table or figure states otherwise.

\paragraph{Code release and reproducibility materials}
We will release the code upon publication.
This appendix gives the hyperparameters, prompts, dataset construction, and evaluation protocol needed to reproduce the experiments.

\subsection{Details of the color palette preferences}
\label{sec:style_exps_detail}
\paragraph{Settings of color palette preferences}
We selected Red and Purple as the two target color preferences.
First, we sample $1000$ images from the Pick-a-Pic v2 training set.
For the Red (Purple) color, we apply image processing to render the images in red (purple).
Concretely, the Red preference dataset is built by increasing the intensity of the red channel to $(\min\{1.5r, 255\}, g, b)$ for positive-label images and decreasing the red channel to $(0.5r, g, b)$ for negative-label images at each pixel of the original images.
The Purple preference dataset is built in a similar way, but we process the red and blue channels simultaneously, using $(\min\{1.5r, 255\}, g, \min\{1.5b, 255\})$ for positive-label images and $(0.5r, g, 0.5b)$ for negative-label images.
Each derived color-preference example keeps the original Pick-a-Pic prompt and stores paired rendered views: a positive image whose target channel intensity is increased and a negative image whose target channel intensity is decreased.
For the Red dataset, this yields prompt-image pairs of the form \texttt{caption}, \texttt{more\_red\_image}, and \texttt{less\_red\_image}; the Purple dataset follows the same paired construction while changing both the red and blue channels.
These derived datasets are used only to train small color experts and anti-experts for the controlled color-preference experiments.
We then fine-tuned the SDXL model using SFT to represent preferences for or against each color, resulting in four SDXL models.

For SFT, we fine-tune the SDXL model using Supervised Fine-Tuning (SFT) with the text-to-image fine-tuning script provided by the Hugging Face \texttt{diffusers} package.\footnote{\url{https://github.com/huggingface/diffusers/tree/main/examples/text_to_image}}
Specifically, for the 1,000-sample color/attribute diffusion experts ($s_\mathrm{Red}^+$, $s_\mathrm{Red}^-$, $s_\mathrm{Purple}^+$, $s_\mathrm{Purple}^-$), we use a dataset of $1000$ Pick-a-Pic v2 prompt-image pairs per preference label. All SFT models are trained using the AdamW optimizer with a learning rate of $1 \times 10^{-5}$ for $1000$ optimization steps ($32$ epochs) with an effective batch size of $32$ in FP16 mixed precision on a single NVIDIA H100 NVL GPU.

For the trained Red (Purple) color reward, denoted $\hat{r}_\mathrm{red}$ ($\hat{r}_\mathrm{purple}$), we reuse the training code provided by the PickScore repository with the same optimization settings. This reward is a CLIP model fine-tuned for $1000$ optimizer steps on the red-positive and red-negative (purple-positive and purple-negative) pairs described above.\footnote{\url{https://github.com/yuvalkirstain/PickScore}} It is a learned proxy used only as the FK Steering guidance signal, whereas every reported color number is computed with the rule-based oracle RedScore (BlueScore) in Eq.~\ref{eq:red_score}; the reward that FK Steering optimizes in the color-preference experiments is therefore separate from the one used for evaluation.

For DiffusionDPO on the Red and Purple datasets, we first convert each dataset to the pairwise-preference format.
We then reuse the training code from the DiffusionDPO GitHub page with the same optimization settings and train for $64$ epochs ($2000$ steps with an effective batch size of $32$).\footnote{\url{https://github.com/SalesforceAIResearch/DiffusionDPO}}

For FK Steering, we use $\hat{r}_\mathrm{red}$ and $\hat{r}_\mathrm{purple}$ as the guidance rewards with $\lambda = 10$ and a particle sample size of $4$, following the public FK Steering implementation.\footnote{\url{https://github.com/zacharyhorvitz/Fk-Diffusion-Steering}}

\begin{figure}[h]
    \begin{center}
        \includegraphics[width=1\linewidth]{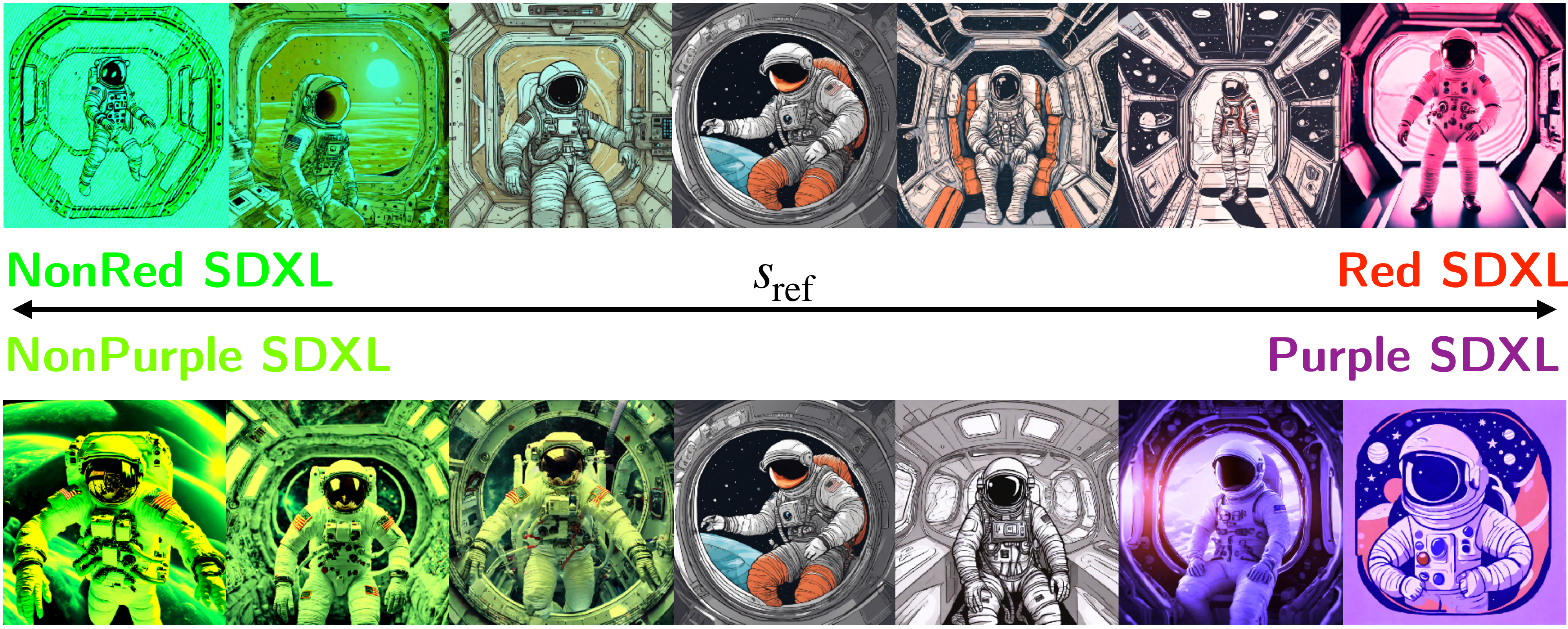}
    \end{center}
    \caption{The continuous $\gamma$ spectrum ($\gamma \in [-3, 3]$) for the Red and Purple color experts (prompt: `astronaut in spaceship'). Negative $\gamma$ reduces the color and positive $\gamma$ strengthens it.}
    \label{fig:color-palette-spectrum-appendix}
\end{figure}

\paragraph{CGG guidance scale $\gamma$}
For the CGG guidance scale $\gamma$, we assess the qualitative and quantitative spectrum in Figure~\ref{fig:color-palette-spectrum-appendix} across $\gamma \in [-3, 3]$ for Red and Purple, and use $\gamma = 3$ for the grids in Figure~\ref{fig:color-palette} and Figure~\ref{fig:color-limited_scenario}.
In the grid, each cell combines the Red and Purple directions at signed scales; in the spectrum, one scale varies at a time and the color intensity changes gradually with $\gamma$.

\paragraph{Evaluation}
We evaluate color control with RedScore and BlueScore, the rule-based color statistics defined in Eq.~\ref{eq:red_score} below, and with CLIPScore.

\paragraph{Definitions of attribute and safety rewards}
We use three scores for the color, photographic-style, and safety axes. The reported RedScore is a rule-based oracle that follows the color statistic of Diffusion-KTO~\citep{SL2024_NeurIPS_DiffusionKTO} and is computed directly from pixel intensities rather than from a learned reward model. Writing $\bar{R}(\vx)$, $\bar{G}(\vx)$, and $\bar{B}(\vx)$ for the mean intensity of each channel over the image and $\bar{\mu}(\vx)=\tfrac{1}{3}(\bar{R}+\bar{G}+\bar{B})(\vx)$ for the mean over all channels, we report
\begin{equation}
r_{\mathrm{red}}(\vx)=\mathrm{clip}_{[1,10]}\!\left(5 + 5\cdot\frac{\bar{R}(\vx)-\bar{\mu}(\vx)}{60}\right),
\label{eq:red_score}
\end{equation}
which rescales the difference between the mean red intensity and the mean over all channels onto a bounded $[1,10]$ range, so a larger value indicates a stronger red cast. BlueScore is defined in the same way with $\bar{B}(\vx)$ in place of $\bar{R}(\vx)$.
Let $f_I$ and $f_T$ denote the normalized image and text embeddings of CLIP and let $\tau$ denote its learned logit scale. For photographic style, we define a prompt-conditioned Realness margin using the public CLIP ViT-H/14 backbone at revision \texttt{de081ac}~\citep{RA2021_ICML_CLIP}. Specifically, $t_{\mathrm{photo}}(\vc)$ is the text template ``a realistic photograph of $\vc$,'' and $t_{\mathrm{anime}}(\vc)$ is ``an anime illustration of $\vc$.'' We compute
\begin{equation}
\begin{aligned}
r_{\mathrm{real}}(\vc,\vx)=\exp(\tau)\big[&f_T(t_{\mathrm{photo}}(\vc))^\top f_I(\vx)\\
&-f_T(t_{\mathrm{anime}}(\vc))^\top f_I(\vx)\big].
\end{aligned}
\label{eq:realness_score}
\end{equation}
Positive values favor photographic rendering over anime rendering for the same content. For the five-reward comparison, each image is scored against the prompt that generated it. Finally, the HEIM/LAION detector returns an NSFW probability $p_{\mathrm{NSFW}}(\vx)$, from which we report Safety as the fraction of images that pass the detector's operating point. Thus, larger RedScore, Realness, and Safety values are better.

\paragraph{Five-reward comparison protocol}
The radar plot (Figure~\ref{fig:five_reward_method_comparison_radar}) shows pre-trained SDXL, Task Arithmetic, FK Steering with five averaged rewards, and CGG with five pairs; the fine-tuned experts are omitted for readability and kept in Table~\ref{tab:five_reward_method_comparison_raw}. The radar normalizes each axis with min-max scaling anchored to the per-metric worst and best scores across all methods in Table~\ref{tab:five_reward_method_comparison_raw}, including the omitted fine-tuned experts, so a curve touching the outer ring attains the best score in the full comparison. All Table~\ref{tab:five_reward_method_comparison_raw} entries are computed on 500 Pick-a-Pic prompts. The CGG run composes five contrastive pairs: (Diffusion-DPO, base), (MaPO-Pick-Safety, base), (SFT$_{\mathrm{Red}}^{+}$, SFT$_{\mathrm{Red}}^{-}$), (RealVisXL~V4.0, Animagine XL 3.1), and (ImageReward-DPO, base), with every pair scale set to $1.0$. The Task Arithmetic row sums the five corresponding task vectors with unit coefficients; other global scales are reported in Table~\ref{tab:task_arithmetic_scale_sweep}.

\paragraph{Task Arithmetic global scale}
Adding five full task vectors at unit weight may over-perturb the merged model, so we rerun the same five-expert merge with global scales in $\{0.2,0.3,0.5,1.0\}$ on the first $200$ Pick-a-Pic prompts with seed $0$.
The scale-$1.0$ row restricts the $500$-prompt run of Table~\ref{tab:five_reward_method_comparison_raw} to the same $200$ prompts, and the two CGG rows report the same $200$ prompts and the full $500$-prompt run.
Table~\ref{tab:task_arithmetic_scale_sweep} shows that smaller scales recover PickScore and ImageReward (scales $0.3$ and $0.5$) and Realness (scales $0.2$ and $0.3$ stay above the base model, and scales $0.2$ to $0.5$ exceed CGG on this subset), while scale $1.0$ matches CGG on RedScore but has the lowest Realness.
Every scale stays below CGG on Safety, and no single global scale is at least as good as CGG on every axis.

\begin{table}[t]
\begin{center}
{\small
\setlength{\tabcolsep}{5pt}
\begin{tabular}{lcccccc}
\toprule
Method & $n$ & PickScore ($\uparrow$) & RedScore ($\uparrow$) & ImageReward ($\uparrow$) & Safety ($\uparrow$) & Realness ($\uparrow$) \\
\midrule
TA, scale $0.2$ & 200 & 22.467 & \textcolor{red}{5.755} & 1.043 & \textcolor{red}{0.923} & 3.630$^\dagger$ \\
TA, scale $0.3$ & 200 & 22.574$^\dagger$ & \textcolor{red}{5.808} & 1.132 & \textcolor{red}{0.931} & 3.544 \\
TA, scale $0.5$ & 200 & 22.562 & 5.921 & 1.174$^\dagger$ & \textcolor{red}{0.921} & \textcolor{red}{3.370} \\
TA, scale $1.0$ & 200 & 22.132 & \textbf{6.478} & 1.131 & \textcolor{red}{0.928} & \textcolor{red}{2.233} \\
\midrule
CGG (same prompts) & 200 & \textbf{22.593} & 6.444$^\dagger$ & 1.148 & \textcolor{red}{0.940}$^\dagger$ & \textcolor{red}{2.577} \\
CGG (all prompts) & 500 & 22.475 & 5.938 & \textbf{1.210} & \textbf{0.974} & \textbf{4.117} \\
\bottomrule
\end{tabular}
}

\end{center}
\caption{Task Arithmetic with different global scales on the first $200$ Pick-a-Pic prompts. All Task Arithmetic rows merge the same five SDXL experts as Table~\ref{tab:five_reward_method_comparison_raw}; only the shared task-vector coefficient changes. The two CGG rows report the same $200$-prompt subset and the full $500$-prompt run of Table~\ref{tab:five_reward_method_comparison_raw}. \textbf{Bold text} marks each metric's best score and $^\dagger$ its second-best; \textcolor{red}{red} marks scores that fall below the base model in Table~\ref{tab:five_reward_method_comparison_raw}.}
\label{tab:task_arithmetic_scale_sweep}
\end{table}

Table~\ref{tab:five_reward_confidence_intervals} reports $95\%$ confidence intervals for Safety and Realness for every method in Table~\ref{tab:five_reward_method_comparison_raw}, and Table~\ref{tab:test_time_steering_pairwise} lists the differences between CGG and FK Steering on all five axes.

\begin{table}[t]
\begin{center}
    {\small
\setlength{\tabcolsep}{3pt}
\begin{tabular}{lcc}
\toprule
Method & Safety & Realness \\
\midrule
Base model & .958 $\pm$ .018 & 3.50 $\pm$ .39 \\
DPO on Pick-a-Pic & .932 $\pm$ .022 & 3.06 $\pm$ .39 \\
MaPO on Pick-Safety & .990 $\pm$ .009 & .23 $\pm$ .34 \\
DPO on ImageReward-DB & .956 $\pm$ .018 & 3.06 $\pm$ .39 \\
SFT on RedScore & .968 $\pm$ .015 & 2.79 $\pm$ .38 \\
RealVisXL V4.0 & .930 $\pm$ .022 & 5.72 $\pm$ .33 \\
FK Steering (5 rewards) & .958 $\pm$ .018 & 3.76 $\pm$ .39 \\
CGG (5-pair combinations) & .974 $\pm$ .014 & 4.12 $\pm$ .34 \\
\bottomrule
\end{tabular}
}

\end{center}
\caption{Mean $\pm$ 95\% confidence interval of Safety and Realness over the $500$ Pick-a-Pic prompts for all methods in Table~\ref{tab:five_reward_method_comparison_raw}.}
\label{tab:five_reward_confidence_intervals}
\end{table}

\begin{table}[t]
\begin{center}
{\footnotesize
\setlength{\tabcolsep}{2.5pt}
\begin{tabular}{lccc}
\toprule
Metric & FK Steering & CGG & $\Delta$ \\
\midrule
PickScore & 22.258 & \textbf{22.475} & +0.217 \\
ImageReward & 1.056 & \textbf{1.210} & +0.154 \\
RedScore & 5.875 & \textbf{5.938} & +0.063 \\
Realness & 3.764 & \textbf{4.117} & +0.353 \\
Safety & 0.958 & \textbf{0.974} & +0.016 \\
\bottomrule
\end{tabular}
}

\end{center}
\caption{FK Steering and CGG on the $500$ Pick-a-Pic prompts of Table~\ref{tab:five_reward_method_comparison_raw}; $\Delta$ is CGG minus FK Steering.}
\label{tab:test_time_steering_pairwise}
\end{table}

The Safety value of SFT on RedScore ($0.968$ versus $0.958$ for the base model) involves no safety supervision; the color SFT simply does not raise the detector's unsafe rate on this prompt set.
Inference latency and memory are analyzed separately in Appendix~\ref{sec:inference_overhead}.

\paragraph{Quantitative results of color palette preferences}
Table~\ref{tab:color-palette} reports RedScore, BlueScore (Eq.~\ref{eq:red_score}), and CLIPScore for the generated images.
\begin{table}[t]
\begin{center}
    {\footnotesize
\setlength{\tabcolsep}{2.5pt}
\begin{tabular}{clccc}
\toprule
Figure~\ref{fig:color-palette} & Diff. models       & RedScore & BlueScore & CLIP  \\
\midrule
Top-left  & $s_{\textcolor{cggpurple}{\mathrm{Purple}}} + s_\mathrm{NonRed}$     & 2.98  & 7.34   & 32.85\\
Top  & $s_{\textcolor{cggpurple}{\mathrm{Purple}}}$               & 4.96  & \textbf{9.55}   & 34.17\\
Top-right  & $s_{\textcolor{cggpurple}{\mathrm{Purple}}} + s_{\textcolor{red}{\mathrm{Red}}}$         & 8.93  & 8.26   & 33.68\\
Middle-left  & $s_\mathrm{NonRed}$            & 2.12  & 4.43   & 33.99\\
Middle  & $s_\mathrm{base}$                & 5.81  & 4.28   & 37.12\\
Middle-right  & $s_{\textcolor{red}{\mathrm{Red}}}$                & \textbf{9.39}  & 4.50   & 34.53\\
Bottom-left  & $s_\mathrm{NonPurple} + s_\mathrm{NonRed}$ & 3.16  & 1.73   & 31.68\\
Bottom  & $s_\mathrm{NonPurple}$           & 5.36  & 1.14   & 31.45\\
Bottom-right  & $s_\mathrm{NonPurple} + s_{\textcolor{red}{\mathrm{Red}}}$     & 9.07  & 2.08   & 32.10\\
\bottomrule
\end{tabular}
}
\end{center}
\caption{Quantitative results for the color palette. Rows refer to positions in the palette visualization in Figure~\ref{fig:color-palette}.}
\label{tab:color-palette}
\end{table}

\subsection{Details of safety constraints and style control}
\label{sec:B.1_safety_exps_detail}
In the safety-constraint experiments, we use the default settings of the NSFW detector from the HEIM benchmark~\citep{liang2023holistic}, with the detector threshold set to $0.9$.

\paragraph{CGG guidance scale $\gamma$}
We assess the results in Figure~\ref{fig:safety_qualitative_comparisons} with $\gamma=0.75$ for CGG($s_\mathrm{MaPO} - s_\mathrm{DPO}$) and $\gamma=1.25$ for CGG($s_\mathrm{MaPO} - s_\mathrm{Anime}$).
Appendix~\ref{sec:C_more_gamma_sensitivity_experiments} provides additional $\gamma$-sensitivity results for RLHF/DPO-based guidance.

\subsection{Details of protected style avoidance experiments}
\label{sec:style_avoidance_exps_detail}

\paragraph{Hugging Face resources and anti-expert checkpoints}
All style avoidance experiments are conducted on top of the public SDXL base architecture (\nolinkurl{stabilityai/stable-diffusion-xl-base-1.0}).
The model-level anti-experts ($s_\mathrm{unfavored}$) are constructed by fusing public style LoRAs into the SDXL base UNet:
\begin{itemize}
    \item \textbf{Studio Ghibli Anti-Expert}: Built by merging the public Studio Ghibli SDXL LoRA (\nolinkurl{artificialguybr/StudioGhibli.Redmond-SDXL}) into the SDXL base UNet to form a standalone style expert UNet.
    \item \textbf{Van Gogh Anti-Expert}: Built by merging the public van Gogh SDXL LoRA (\nolinkurl{nerjs/van-gogh-sdxl-lora}) into the SDXL base UNet.
\end{itemize}
In both style avoidance settings, the anchor $s_\mathrm{anc}$ and the favored side are the pre-trained SDXL base model ($s_\mathrm{base}$). The pair $(s_\mathrm{base}, s_\mathrm{style\_expert})$ gives the guidance $s_\mathrm{CGG} = s_\mathrm{base} - \gamma (s_\mathrm{style\_expert} - s_\mathrm{base})$.

\paragraph{Prompt datasets and baseline configurations}
We evaluate style avoidance on two style benchmark prompt sets:
\begin{itemize}
    \item \textbf{Studio Ghibli}: Evaluated on $24$ scene prompts ($12$ explicit style-invocation prompts naming ``Studio Ghibli'' and $12$ implicit scene prompts without explicit style keywords).
    \item \textbf{Van Gogh}: Evaluated on $20$ scene prompts ($10$ explicit style-invocation prompts naming ``van Gogh'' and $10$ implicit scene prompts).
\end{itemize}
All methods are sampled with $50$ denoising steps, a classifier-free guidance (CFG) scale of $7.5$, and a fixed random seed ($0$) across rows, so that all comparisons are paired.
We compare CGG against the inference-time text negative prompting baseline~\citep{PS2023_CVPR_SLD}. For Studio Ghibli, the negative prompt string is ``Studio Ghibli, ghibli, anime film still, hand-drawn animation, cel shading''. For van Gogh, the negative prompt string is ``Vincent van Gogh, van gogh, oil painting, post-impressionism, thick impasto''.

\paragraph{Evaluation metrics and medium-shift measurement}
Style avoidance is quantified with two metrics:
\begin{itemize}
    \item \textbf{Style removed ($\uparrow$)}: Measures style suppression with Content-Style Disentanglement (CSD) feature similarity~\citep{SG2024_arXiv_CSD} to the style expert's own outputs. We compute CSD similarity between generated samples and reference images from the style expert; \emph{Style removed} is the reduction in CSD similarity relative to the unguided base model. CSD is computed from image features, so listing style words in a negative prompt does not lower it directly.
    \item \textbf{Photo margin}: Measures whether suppressing the protected style shifts the medium from illustration or painting toward photography. We compute the CLIP ViT-H/14 text margin between a photorealistic prompt template (\texttt{"a realistic photograph of [content]"}) and an illustrative prompt template (\texttt{"an anime illustration of [content]"} for Ghibli, or \texttt{"an oil painting of [content]"} for van Gogh). A positive value means the sample reads as a photograph; a negative value means it stays within the illustrated or painted medium.
\end{itemize}

\subsection{Details of applying CGG for RLHF/DPO-style pipelines}
\label{sec:detail_CGG_PO}

\paragraph{More results of enhancing RLHF/DPO-style aligned models for SDXL}
Table~\ref{tab:Avg_results-Pick_a_Pic} reports the BT-induced pair (DPO, base): the guided score $s_\mathrm{base}+\gamma^\ast(s_\mathrm{DPO}-s_\mathrm{base})$ improves PickScore, HPSv2, and CLIP over both the base model and the DPO checkpoint.
\begin{table}[t]
\begin{center}
    \begin{tabular}{c|ccc}
& PickScore & HPSv2  & CLIP  \\
\hline
$s_\mathrm{base}$                & 22.11                &        0.29            &       36.99            \\
$s_\mathrm{DPO}$                      & 22.41                  & 0.30         &       37.78            \\
$s_\mathrm{base}+\gamma^\ast(s_\mathrm{DPO}-s_\mathrm{base})$ &  \textbf{22.48  }     &     \textbf{0.31}  &  \textbf{37.98}
\end{tabular}
\end{center}
\caption{Average rewards for enhancing a DPO-aligned SDXL model. \textbf{Bold text} marks the best result.}
\label{tab:Avg_results-Pick_a_Pic}
\end{table}

Table~\ref{tab:Avg_results-MaPO} reports additional MaPO-based SDXL results.
\begin{table}[t]
\begin{center}
    \begin{tabular}{c|ccc}
                                    & PickScore ($\uparrow$) & HPSv2 ($\uparrow$) & CLIP ($\uparrow$) \\
\hline
$s_\mathrm{base}$                           &    22.11   &      0.29 &   36.99  \\
$s_\mathrm{MaPO}$                      &   22.02  &    \textbf{0.30}           &       37.14            \\
\shortstack[l]{$s_\mathrm{base}+\gamma^\ast($$s_\mathrm{MaPO}-s_\mathrm{base})$}            &      \textbf{22.15}       &      \textbf{0.30}              & \textbf{37.28}                  \\
\end{tabular}
\end{center}
\caption{Average rewards for SDXL, MaPO, and CGG applied to MaPO on prompts from the Pick-a-Pic v2 test set. \textbf{Bold text} marks the best result.}
\label{tab:Avg_results-MaPO}
\end{table}

Section~\ref{sec:flexible-preference-signals} presents the contrastive scoring form of SFT using positive-label and negative-label examples in Table~\ref{tab:Avg_results-CGG-Pos-Neg}.
As discussed in Section~\ref{sec:related-works}, CHATS also fine-tunes positive and negative models and introduces proxy-prompt-based sampling (PPBS).
Table~\ref{tab:Avg_results-Chats} compares CGG gradient guidance with CHATS-PPBS and shows similar performance under different sampling strategies.
\begin{table}[t]
    \begin{center}
        {\small
\setlength{\tabcolsep}{3pt}
\begin{tabular}{lccc}
\toprule
Metric & $s_\mathrm{base}$ & PPBS & \shortstack{CGG\\$s_\mathrm{base}+\gamma^\ast(s_\mathrm{CHATS}^+ - s_\mathrm{CHATS}^-)$} \\
\midrule
PickScore & 22.114 & 22.095 & \textbf{22.153} \\
Aesthetic & \textbf{6.481} & 6.462 & 6.348 \\
HPSv2 & 0.292 & \textbf{0.309} & 0.304 \\
CLIP & 36.994 & 36.705 & \textbf{37.483} \\
ImageReward & 0.857 & \textbf{1.052} & 1.039 \\
\bottomrule
\end{tabular}
}

    \end{center}
\caption{Average rewards for SDXL, CHATS-PPBS, and CGG sampling with $s_\mathrm{CHATS}^+$ and $s_\mathrm{CHATS}^-$ on prompts from the Pick-a-Pic v2 test set. \textbf{Bold text} marks the best result.
    PPBS denotes the proxy-prompt-based sampling strategy proposed in CHATS~\citep{MF2025_ICML_CHATS}.}
\label{tab:Avg_results-Chats}
\end{table}

\paragraph{More results of enhancing RLHF/DPO-style aligned models for SD 1.5}
We also evaluate the officially released Diffusion-KTO SD 1.5 checkpoint in Table~\ref{tab:Avg_results-KTO}.
The fine-tuned Diffusion-KTO checkpoint gives the best score among the evaluated $\gamma$ values, and the CGG run matches this result at the selected scale $\gamma^\ast = 1$.
\begin{table}[t]
\begin{center}
    \begin{tabular}{c|ccc}
        & $s_\mathrm{base}$ & $s_\mathrm{KTO}$ & \shortstack{$s_\mathrm{base}+\gamma^\ast($$s_\mathrm{KTO}-s_\mathrm{base})$}\\
        \hline
        PickScore & 20.529 & 21.072 & 21.072 \\
        Aesthetic & 5.739 & 6.145 & 6.145 \\
        HPSv2 & 0.259 & 0.291 & 0.291 \\
        CLIP & 33.382 & 34.252 & 34.252 \\
        ImageReward & 0.064 & 0.631 & 0.631 \\
    \end{tabular}

\end{center}
\caption{Average rewards between SD 1.5, KTO, and applying CGG in KTO using prompts from the Pick-a-Pic v2 test set.}
\label{tab:Avg_results-KTO}
\end{table}

\paragraph{Results of ImageReward on ImageRewardDB test set}
Table~\ref{tab:ImagerewardDB_results} evaluates the (DPO, base) pair on the ImageRewardDB test set with the ImageReward score.
\begin{table}[t]
    \begin{center}
        {\small
\setlength{\tabcolsep}{2pt}
\begin{tabular}{l c cc cc}
\toprule
& Base & \multicolumn{2}{c}{Diff.-DPO} & \multicolumn{2}{c}{MaPO} \\
\cmidrule(lr){2-2} \cmidrule(lr){3-4} \cmidrule(lr){5-6}
Model & Pre-trained & Fine-tuned & CGG & Fine-tuned & CGG \\
\midrule
SDXL & 0.910 & \textbf{1.040} & 1.030 (+0.120) & N/A & N/A \\
\bottomrule
\end{tabular}
}
    \end{center}
\caption{Average ImageReward scores on prompts from the ImageRewardDB test set. \textbf{Bold text} marks the best score. Parentheses report changes relative to the pre-trained model.}
\label{tab:ImagerewardDB_results}
\end{table}
CGG improves over the pre-trained model on ImageRewardDB~\citep{xu2023imagereward} by $0.120$ but stays slightly below the DPO checkpoint ($1.030$ versus $1.040$), with $\gamma$ fixed at the value selected on Pick-a-Pic.

\subsection{Supplementary heterogeneous-feedback results}
\label{sec:supp_heterogeneous_feedback}

The quantitative comparison that combines model-based CGG with FK Steering appears in the main text (Table~\ref{tab:cgg_fk_pick_red}). The broader deployment comparison against FK Steering with averaged rewards appears in Table~\ref{tab:five_reward_method_comparison_raw} and Table~\ref{tab:test_time_steering_pairwise}.

Table~\ref{tab:CGGmultiple_pref} reports the PickScore+ImageReward two-preference experiment.
\begin{table}[t]
    \begin{center}
    {\small
\setlength{\tabcolsep}{2.5pt}
\begin{tabular}{l c cc c}
\toprule
Metric & Base & \multicolumn{2}{c}{DPO} & CGG \\
\cmidrule(lr){3-4}
& Pre-trained & Pick-a-Pic & ImageReward & (ours) \\
\midrule
PickScore ($\uparrow$) & 22.114 & \textbf{22.405} & 22.079 & 22.395 \\
ImageReward ($\uparrow$) & 0.857 & 0.871 & 0.834 & \textbf{1.070} \\
\bottomrule
\end{tabular}
}
    \end{center}
\caption{Average PickScore and ImageReward for two-preference alignment on SDXL using prompts from the Pick-a-Pic v2 test set. \textbf{Bold text} marks the best result.}
\label{tab:CGGmultiple_pref}
\end{table}

\subsection{Independent replication of the five-reward comparison}
\label{sec:independent_replication}

To check that the result of Table~\ref{tab:five_reward_method_comparison_raw} does not depend on one machine or software stack, we reran the five-pair composition end to end on a different machine (also a single NVIDIA H100 NVL) with a separately installed environment, the same configuration, the same $500$ Pick-a-Pic v2 test prompts, and the same pair scales.
Table~\ref{tab:independent_replication} reports the replication together with per-prompt win rates against the base model over the same prompts.
All five axes improve over the base model in both runs.
The CGG values of PickScore, Safety, and RedScore agree with Table~\ref{tab:five_reward_method_comparison_raw} to within $0.03$; the improvements differ more, from $+0.006$ versus $+0.016$ on Safety to $+0.304$ versus $+0.613$ on Realness.
Per prompt, CGG scores higher than the base model on $79.2\%$ of prompts for ImageReward and $73.0\%$ for PickScore; on RedScore ($54.8\%$) and Realness ($53.0\%$) the win rate is close to one half, so the mean gains on these two axes are not shared by most prompts.
Mean NSFW probability also decreases ($0.061$ versus $0.066$ for the base model), consistent with the Safety column.
\begin{table}[t]
\begin{center}
{\small
\setlength{\tabcolsep}{3pt}
\begin{tabular}{lccccc}
\toprule
Metric & Base & CGG & $\Delta$ & $\Delta_\mathrm{main}$ & Win-rate \\
\midrule
PickScore & 22.030 & 22.451 & $+0.420$ & $+0.361$ & 73.0\% \\
Safety & 0.956 & 0.962 & $+0.006$ & $+0.016$ & -- \\
ImageReward & 0.696 & 1.160 & $+0.464$ & $+0.353$ & 79.2\% \\
RedScore & 5.824 & 5.930 & $+0.106$ & $+0.026$ & 54.8\% \\
Realness & 3.513 & 3.816 & $+0.304$ & $+0.613$ & 53.0\% \\
\bottomrule
\end{tabular}
}

\end{center}
\caption{Base and CGG columns are means from the replication run; $\Delta$ is the CGG-minus-base improvement in the replication and $\Delta$ (main) the corresponding improvement in Table~\ref{tab:five_reward_method_comparison_raw}. The win-rate column counts prompts on which CGG scores higher than the base model; Safety is a thresholded rate and has no per-prompt win definition. The ImageReward scorer in the replication environment gives lower absolute values than the one used for Table~\ref{tab:five_reward_method_comparison_raw} ($0.696$ versus $0.857$ for the base model), so that row is comparable only through $\Delta$.}
\label{tab:independent_replication}
\end{table}

\subsection{Guidance geometry of multi-preference composition}
\label{sec:guidance_geometry}
\begin{figure}[t]
    \begin{center}
        \includegraphics[width=1\linewidth]{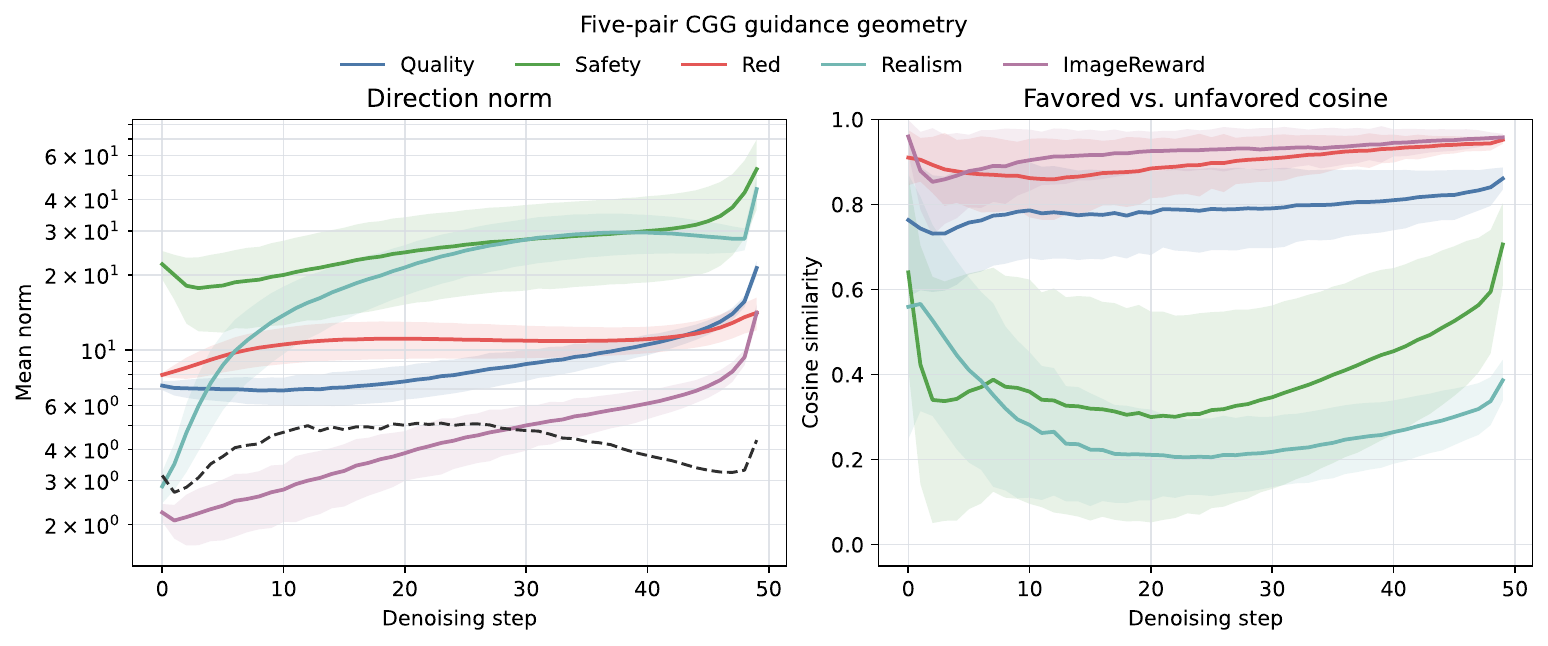}
    \end{center}
    \caption{Per-pair guidance geometry of the five-pair CGG composition (mean $\pm$ one standard deviation across prompts). Left: norm of each contrastive direction across diffusion timesteps (log scale), with the base score norm shown for scale. Right: cosine similarity between the favored and unfavored scores of each pair. All pairs share the uniform pair scale $1.0$.}
    \label{fig:guidance_geometry}
\end{figure}

To examine how composed preference directions behave during sampling, we log per-pair guidance statistics for the five-pair composition of Table~\ref{tab:five_reward_method_comparison_raw} (quality: DPO$-$base; safety: MaPO$-$base; red: SFT$^{+}-$SFT$^{-}$; realism: RealVisXL$-$Animagine; image reward: ImageReward-DPO$-$base) under the same uniform pair scales, over $500$ Pick-a-Pic v2 test prompts with $50$ denoising steps.
Figure~\ref{fig:guidance_geometry} reports, for each pair, the norm of the contrastive direction $\|s_{\rho_i}-s_{\kappa_i}\|$ and the cosine similarity between the favored and unfavored scores across timesteps.
The five pairs differ on both statistics.
The safety and realism pairs are the most contrastive, with mean cosine similarities of $0.39$ and $0.28$ and the largest direction norms ($26.3$ and $21.8$); in both, the two sides render images in clearly different styles: MaPO drifts toward an anime style relative to the base model (Figure~\ref{fig:safety_qualitative_comparisons}), and RealVisXL and Animagine were fine-tuned toward opposing styles.
The quality and red pairs sit in between (cosines $0.79$ and $0.90$, norms $9.0$ and $10.8$).
The image-reward pair is the most aligned with its anti-expert (cosine $0.92$) and has by far the smallest norm ($4.7$), consistent with the weak ImageReward-DPO checkpoint, which falls below the base model on all five axes in Table~\ref{tab:five_reward_method_comparison_raw}: a pair whose two sides nearly agree supplies little guidance signal.
The pairs therefore differ in both magnitude and favored--unfavored alignment: the safety and realism pairs contribute the largest directions and the image-reward pair the smallest.
\clearpage
\section{Implementation of methods and hyper-parameter tuning}

\subsection{CGG-FK: Integrating CGG with FK Steering}
\label{sec:CGG-FK}

CGG-FK uses CGG to define the model-side denoising direction and FK Steering to perform reward-based particle selection.
At each denoising step, CGG first forms a guided score from the anchor model and the selected contrastive preference pairs.
FK Steering then treats the resulting transitions as its particle candidates, evaluates their clean-image estimates with the reward model, and resamples according to the FK weights.
Thus, CGG changes how each candidate transition is proposed, while FK Steering keeps its original role as the particle-based reward selection step.
Algorithm~\ref{alg:CGG_FK} summarizes this combined sampler.

\begin{algorithm}
\caption{CGG-FK sampling with model-level contrastive guidance and reward-based resampling}
\label{alg:CGG_FK}
\begin{algorithmic}[1]
\STATE \textbf{Input:} prompt $\vc$, anchor score $s_\mathrm{anc}$, contrastive pairs $\{(s_{\rho_i},s_{\kappa_i})\}_{i=1}^N$
\STATE \textbf{Input:} pair scales $\{\gamma_i\}_{i=1}^N$, FK scale $\lambda$, number of particles $J$, scheduler, reward model $r(\vc,\vx)$
\STATE \textbf{Output:} generated image $\vx_0$
\STATE Sample initial particle latents $\{\vx_T^{(j)}\}_{j=1}^J$
\FOR{$t \gets T~\text{to}~1$}
\FOR{$j \gets 1~\text{to}~J$}
\STATE $\epsilon_\mathrm{anc}^{(j)} = s_\mathrm{anc}(\vc, \vx_t^{(j)}, t)$
\STATE $\Delta_i^{(j)} = s_{\rho_i}(\vc,\vx_t^{(j)},t)-s_{\kappa_i}(\vc,\vx_t^{(j)},t)$ for each pair $i$
\STATE $\epsilon_\mathrm{CGG}^{(j)} = \epsilon_\mathrm{anc}^{(j)} + \sum_{i=1}^N \gamma_i \Delta_i^{(j)}$
\STATE $(\hat{\vx}_0^{(j)}, \tilde{\vx}_{t-1}^{(j)}) = \mathrm{Scheduler}(\epsilon_\mathrm{CGG}^{(j)}, t, \vx_t^{(j)})$
\STATE $w_j \propto \exp(\lambda r(\vc,\hat{\vx}_0^{(j)}))$
\ENDFOR
\STATE Resample $\{\vx_{t-1}^{(j)}\}_{j=1}^J$ from candidates $\{\tilde{\vx}_{t-1}^{(j)}\}_{j=1}^J$ according to weights $\{w_j\}_{j=1}^J$
\ENDFOR
\STATE \textbf{return} $\vx_0$ from the final particle set
\end{algorithmic}
\end{algorithm}

\subsection{Guidance-scale hyperparameter: \texorpdfstring{$\gamma$}{gamma}-sensitivity}
\label{sec:C_more_gamma_sensitivity_experiments}
In this section, we provide results for the $\gamma$-sensitivity experiments. Here $\gamma$ denotes the single active guidance scale used in the experiments, matching $\gamma_i$ in Eq.~\ref{eqn:contrastive_gradient_guidance} when only one contrastive pair is varied.

\begin{figure}[t]
    \begin{center}
        \includegraphics[width=0.49\linewidth]{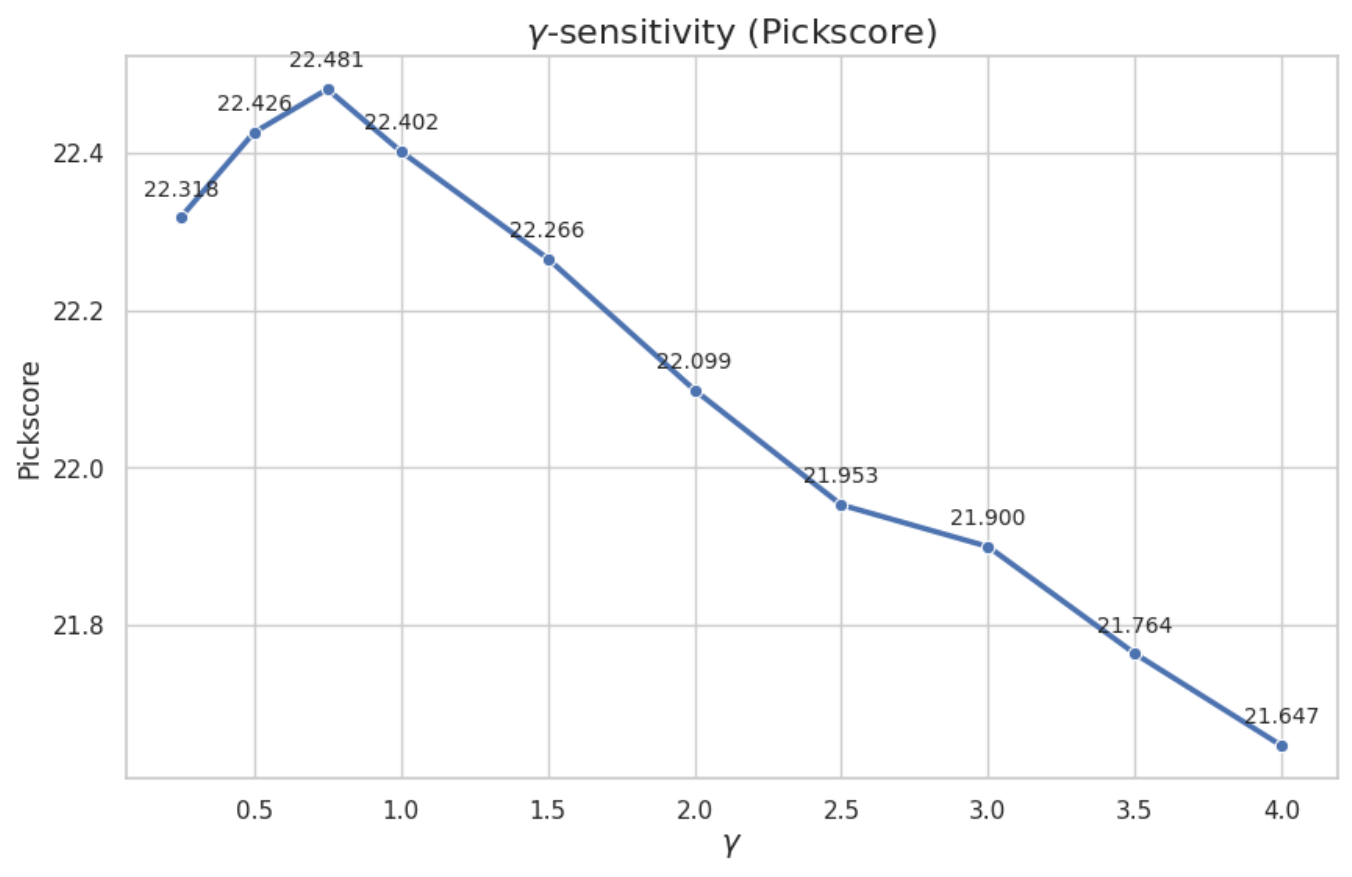}
        \includegraphics[width=0.49\linewidth]{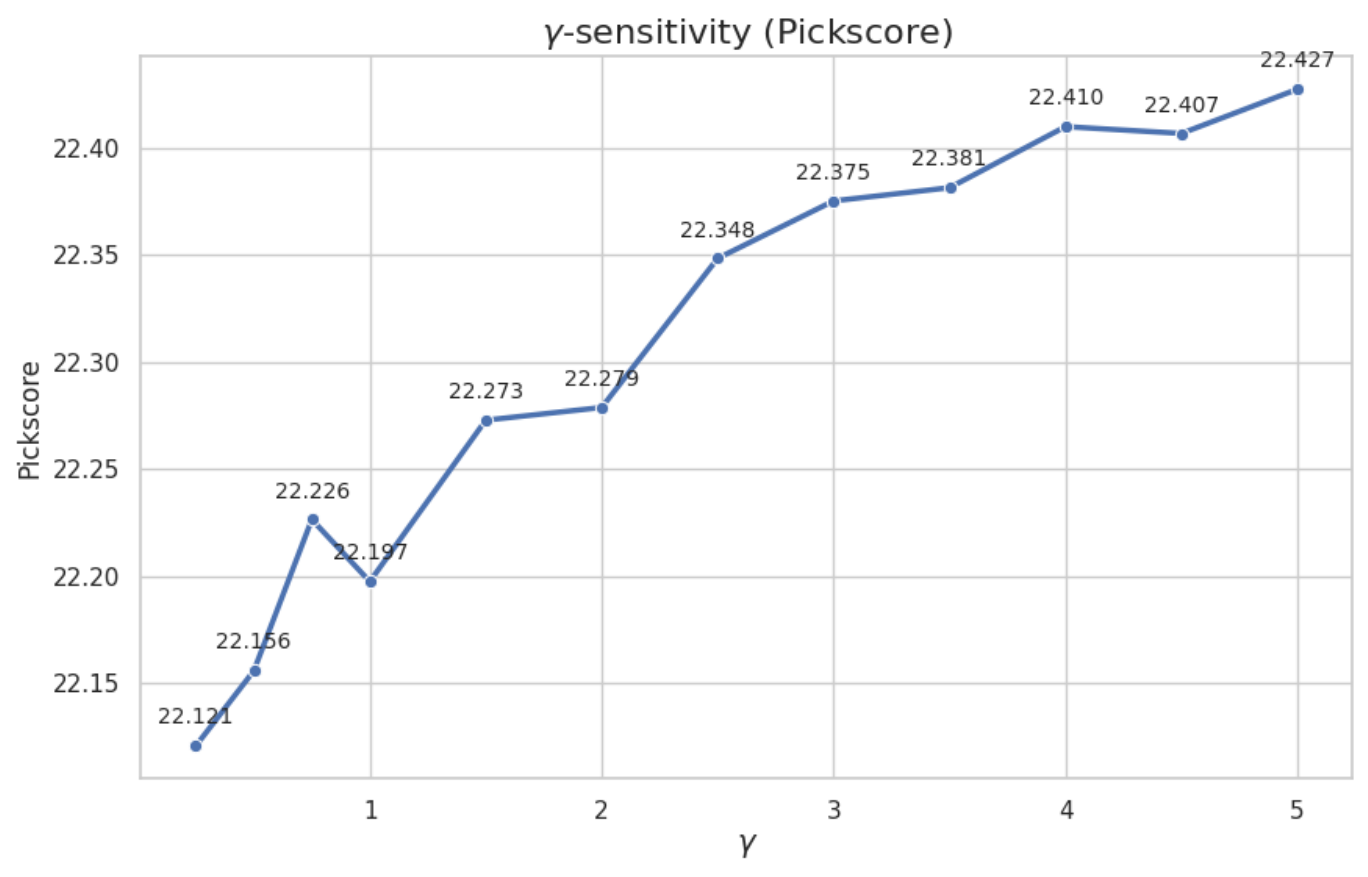}
    \end{center}
    \caption{$\gamma$-sensitivity of PickScore ($y$-axis) over the guidance scale $\gamma$ ($x$-axis). Left: applying CGG in Diffusion-DPO. Right: applying CGG in SFT(Pos) and SFT(Neg).}
    \label{fig:CGG-hyperparams-SFTPosNeg-PickScore_gamma}
\end{figure}

\paragraph{Choice of guidance scale}
We empirically investigate the effect of the guidance scale by sweeping $\gamma$ from $0.25$ upward and selecting $\gamma^\ast$ on the Pick-a-Pic v2 validation set.
As shown in Figure~\ref{fig:CGG-hyperparams-SFTPosNeg-PickScore_gamma} (Left), PickScore first increases and then decreases as $\gamma$ grows, giving a clear validation criterion for choosing the default scale.
Based on this trend across different contrastive forms, we select $\gamma^\ast = 0.75$ for SDXL as the default guidance scale for experiments reported in Section~\ref{sec:flexible-preference-signals}.
We also study $\gamma$-sensitivity across different contrastive forms and metrics. Figure~\ref{fig:CGG-hyperparams-SFTPosNeg-PickScore_gamma} (Right) extends the CGG(Pos, Neg) sweep to $\gamma = 5.0$.
\begin{figure}[t]
    \begin{center}
        \includegraphics[width=0.49\linewidth]{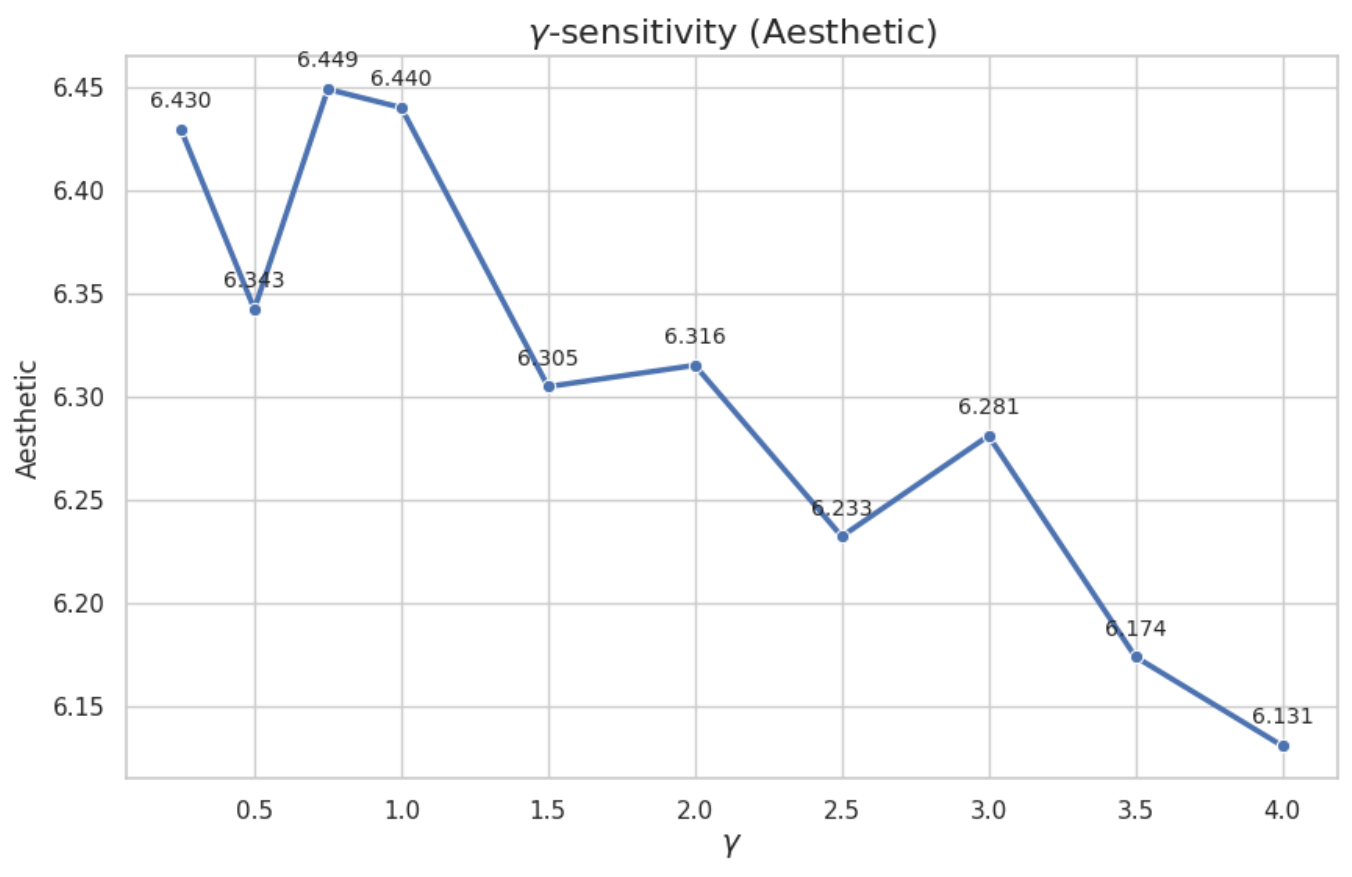}
        \includegraphics[width=0.49\linewidth]{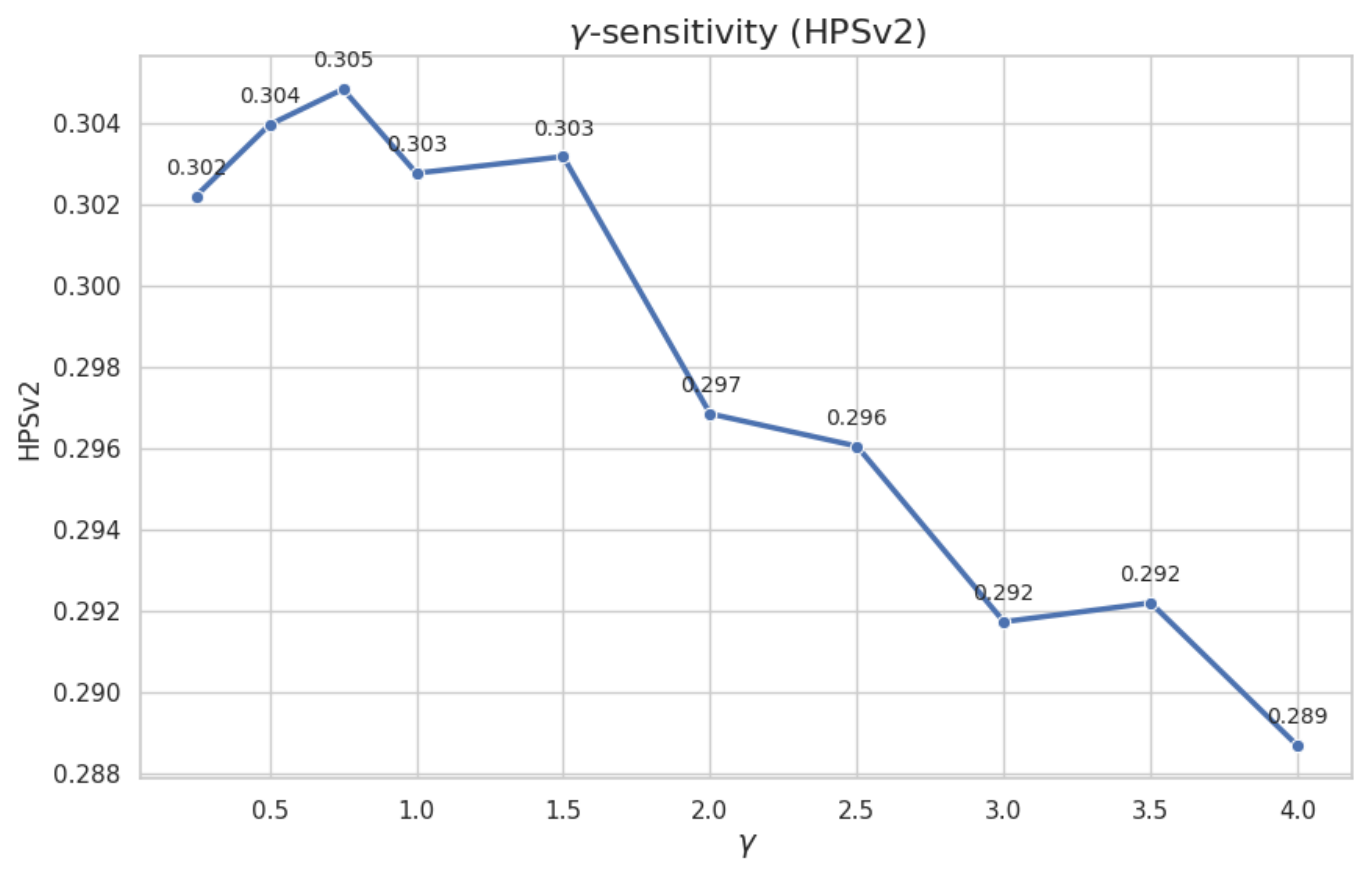}
    \end{center}
    \caption{$\gamma$-sensitivity of applying CGG in Diffusion-DPO over the guidance scale $\gamma$ ($x$-axis). Left: Aesthetic ($y$-axis). Right: HPSv2 ($y$-axis).}
    \label{fig:CGG-hyperparams-DiffusionDPO-Aesthetic_gamma}
\end{figure}
\begin{figure}[t]
    \begin{center}
        \includegraphics[width=0.49\linewidth]{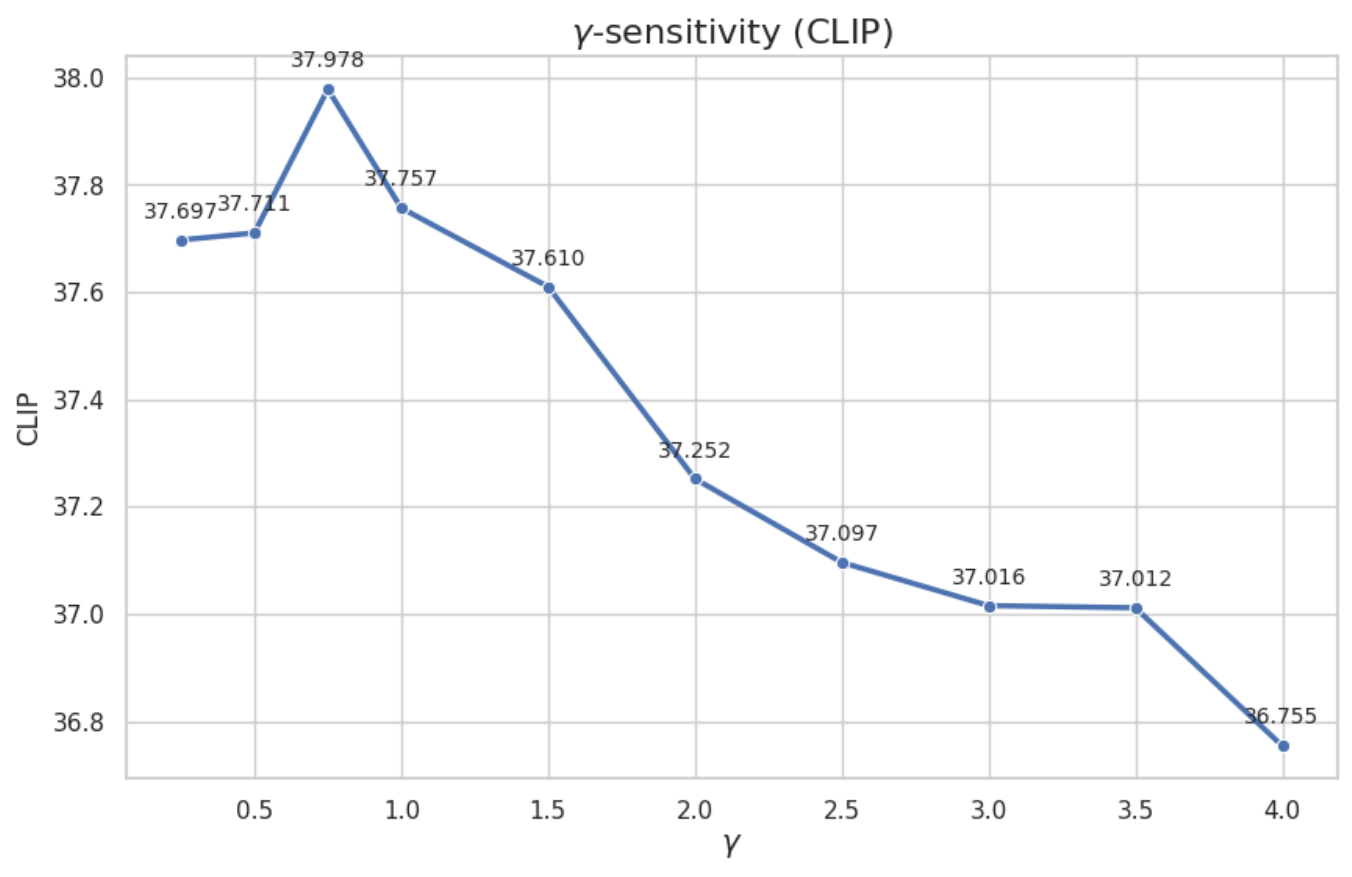}
        \includegraphics[width=0.49\linewidth]{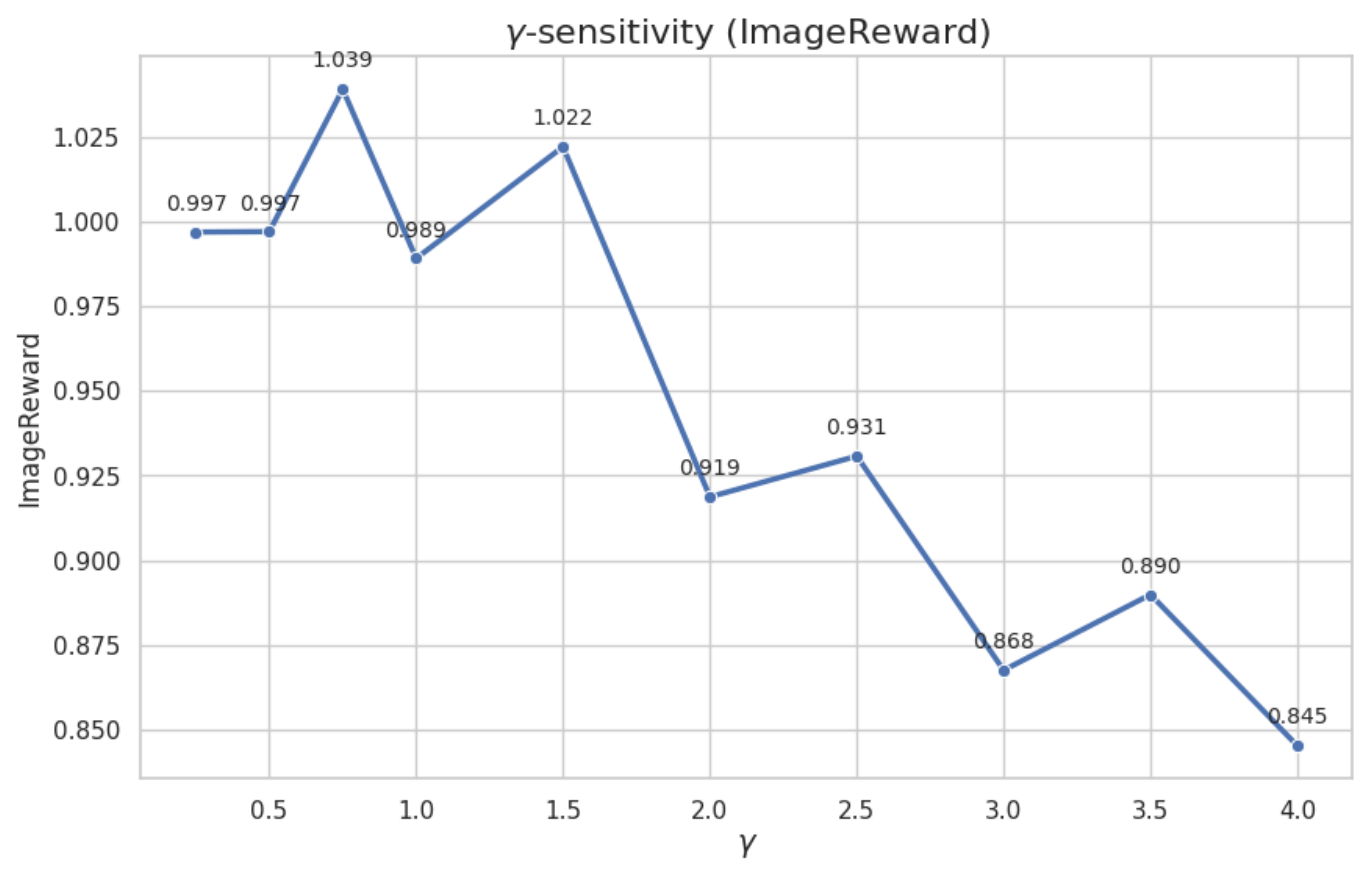}
    \end{center}
    \caption{$\gamma$-sensitivity of applying CGG in Diffusion-DPO over the guidance scale $\gamma$ ($x$-axis). Left: CLIP ($y$-axis). Right: ImageReward ($y$-axis).}
    \label{fig:CGG-hyperparams-DiffusionDPO-CLIPScore_gamma}
\end{figure}
\begin{figure}[t]
    \begin{center}
        \includegraphics[width=0.49\linewidth]{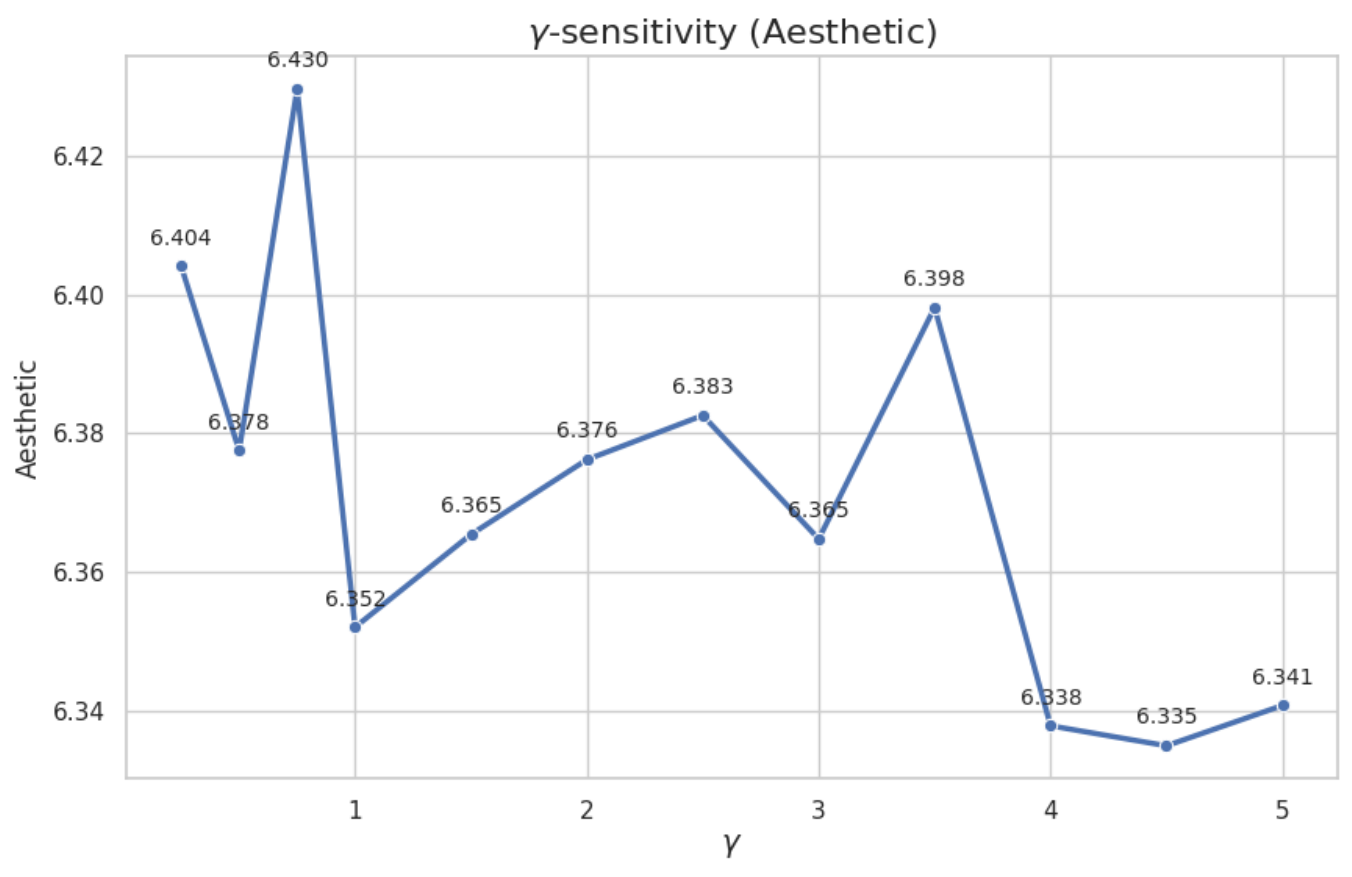}
        \includegraphics[width=0.49\linewidth]{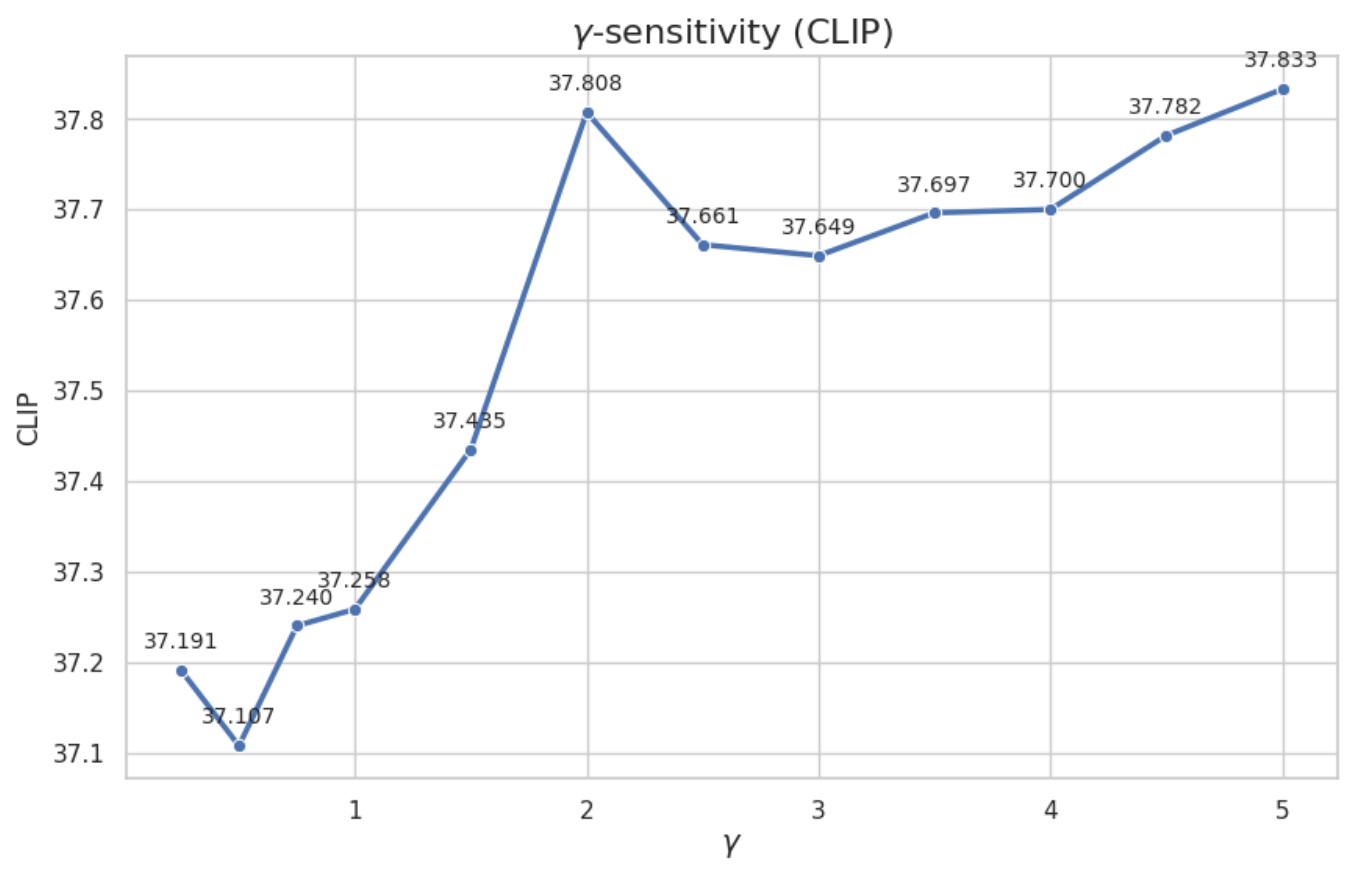}
    \end{center}
    \caption{$\gamma$-sensitivity of applying CGG in SFT(Pos) and SFT(Neg) over the guidance scale $\gamma$ ($x$-axis). Left: Aesthetic ($y$-axis). Right: CLIP ($y$-axis).}
    \label{fig:CGG-hyperparams-Pos-Neg-Aesthetic_gamma}
\end{figure}
\begin{figure}[t]
    \begin{center}
        \includegraphics[width=0.49\linewidth]{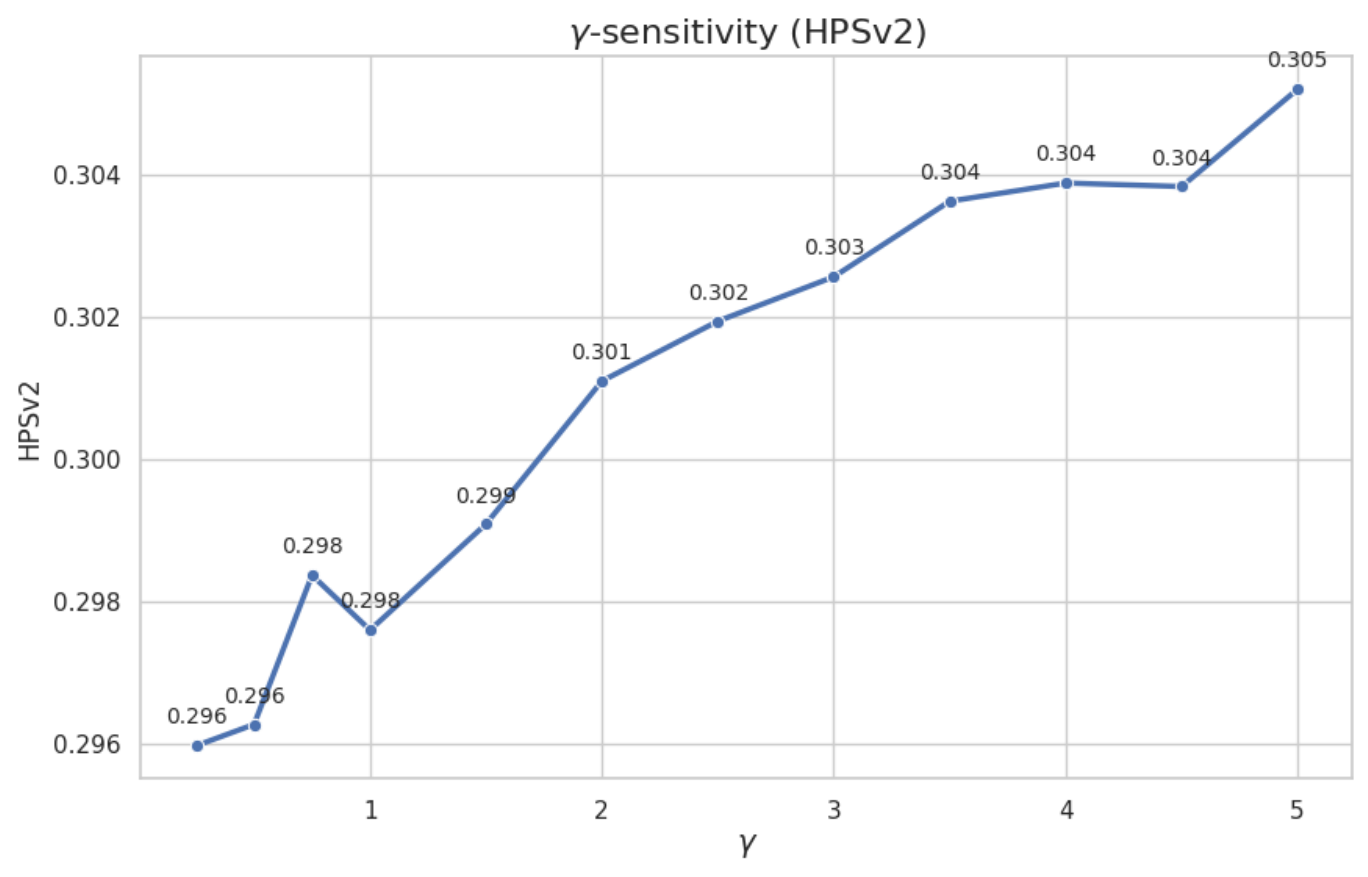}
        \includegraphics[width=0.49\linewidth]{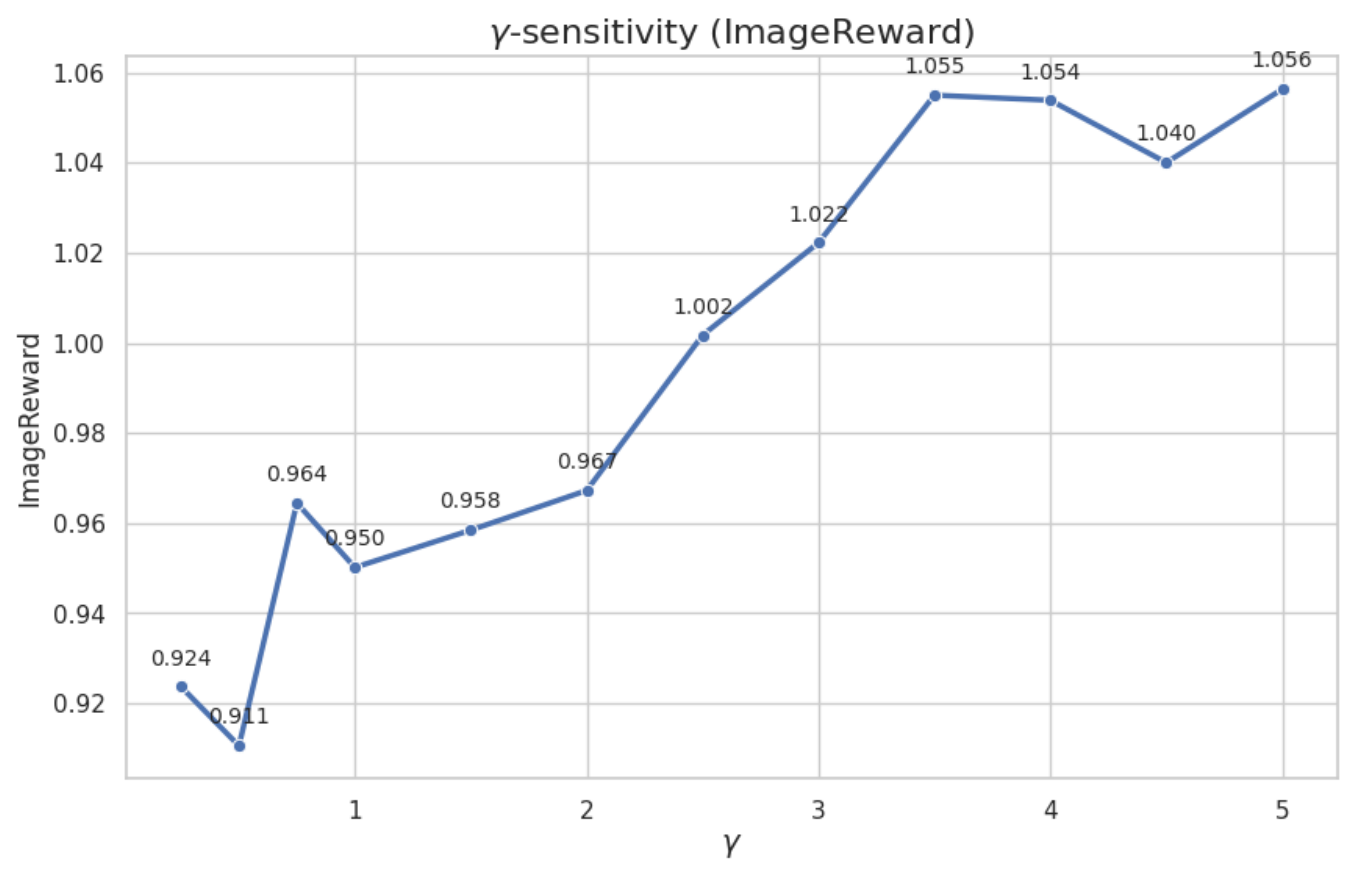}
    \end{center}
    \caption{$\gamma$-sensitivity of applying CGG in SFT(Pos) and SFT(Neg) over the guidance scale $\gamma$ ($x$-axis). Left: HPSv2 ($y$-axis). Right: ImageReward ($y$-axis).}
    \label{fig:CGG-hyperparams-Pos-Neg-HPSv2_gamma}
\end{figure}
\clearpage
\section{More qualitative results}
\label{sec:D_qualitative_results}
\begin{figure}[htbp]
    \centering
    \begin{tabular}{ccc}
        $s_\mathrm{base}$ & $s_\mathrm{DPO}$ & \shortstack{$s_\mathrm{base}+\gamma^\ast($\\$s_\mathrm{DPO}-s_\mathrm{base})$} \\
        \includegraphics[width=0.25\textwidth]{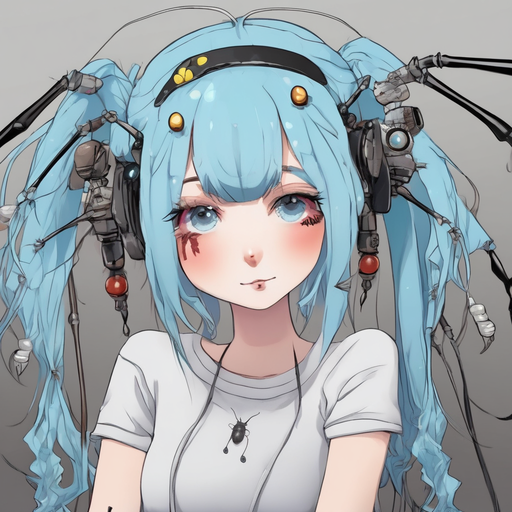} &
        \includegraphics[width=0.25\textwidth]{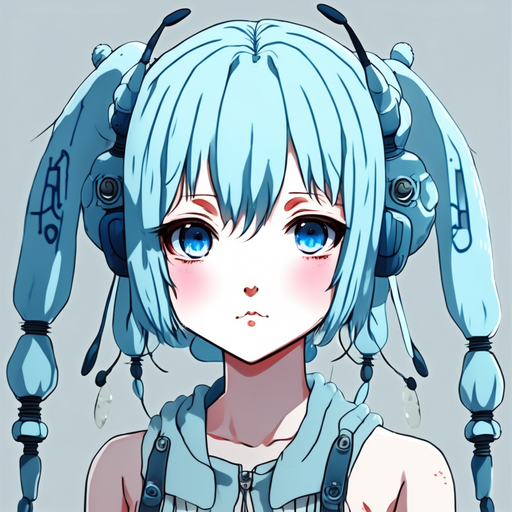} &
        \includegraphics[width=0.25\textwidth]{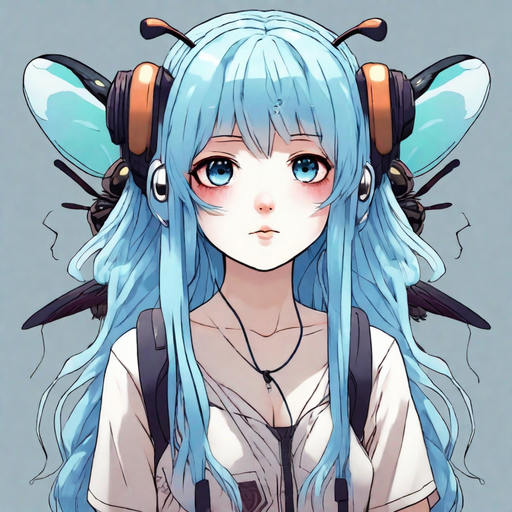} \\
        \includegraphics[width=0.25\textwidth]{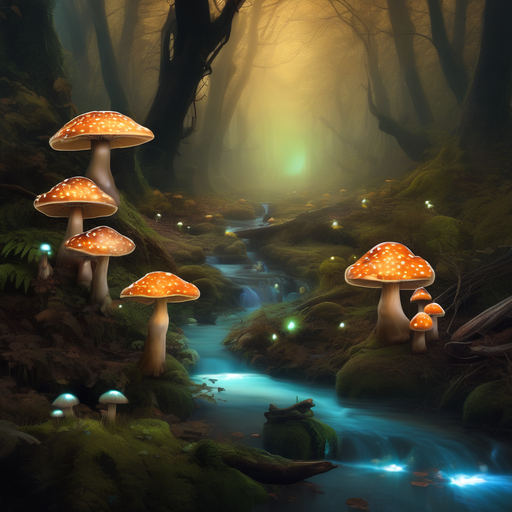} &
        \includegraphics[width=0.25\textwidth]{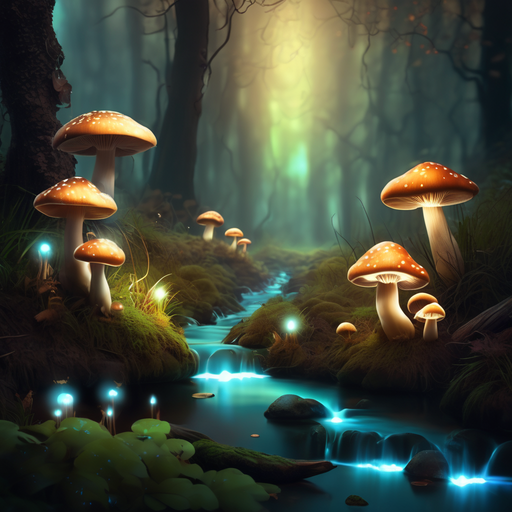} &
        \includegraphics[width=0.25\textwidth]{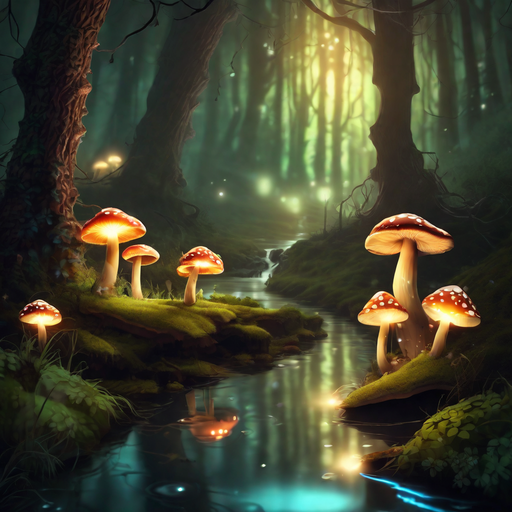} \\
        \includegraphics[width=0.25\textwidth]{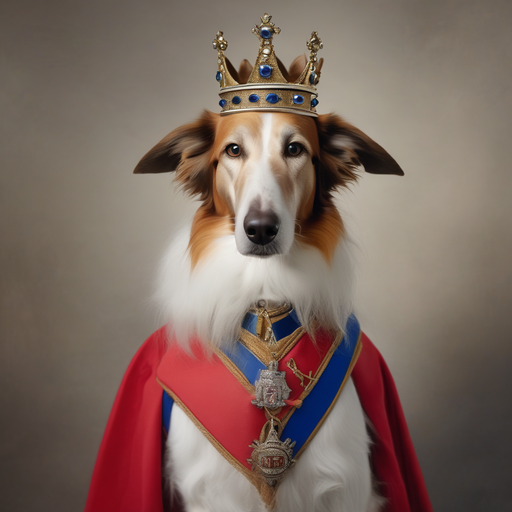} &
        \includegraphics[width=0.25\textwidth]{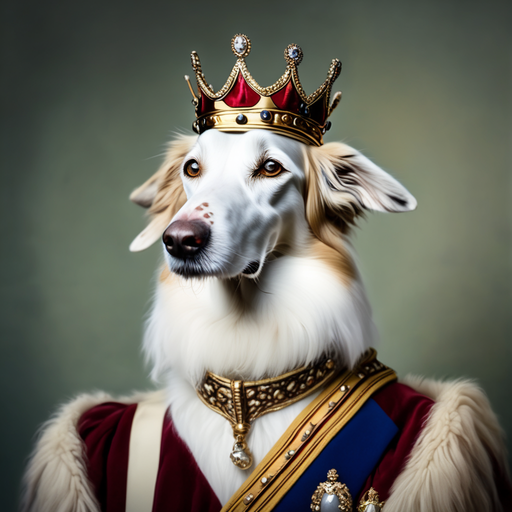} &
        \includegraphics[width=0.25\textwidth]{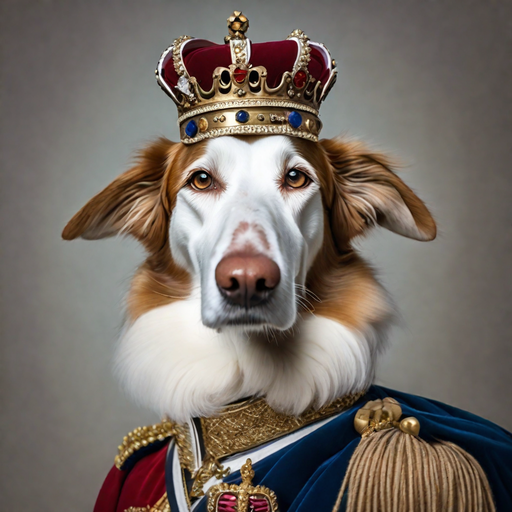} \\
        \includegraphics[width=0.25\textwidth]{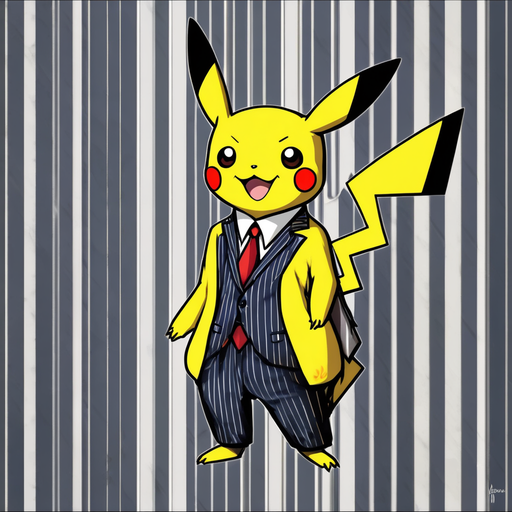} &
        \includegraphics[width=0.25\textwidth]{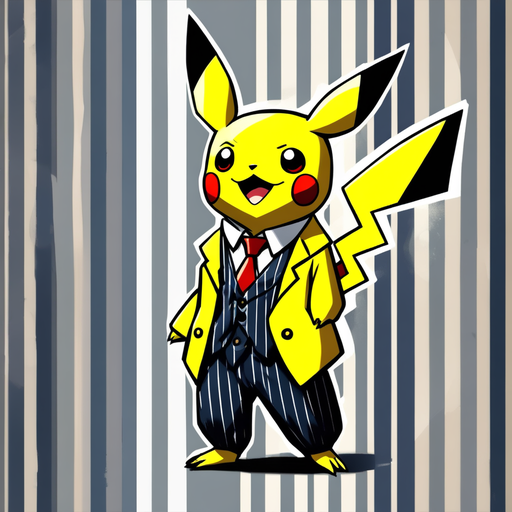} &
        \includegraphics[width=0.25\textwidth]{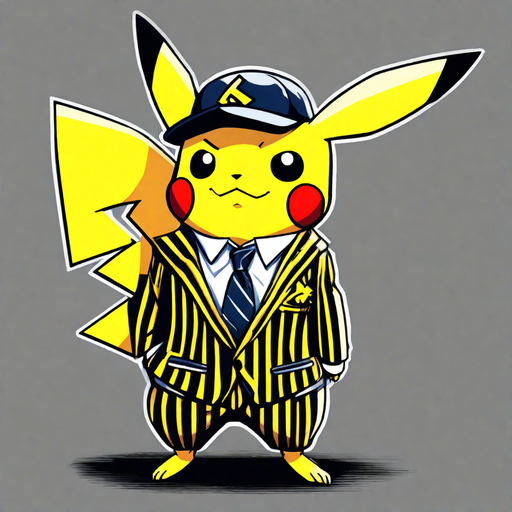} \\
    \end{tabular}
    \caption{Qualitative results on SDXL. Prompts: 1) light blue haired anime girl with bug antennas 2) Mystical forest with glowing mushrooms and a babbling brook 3) portrait of sir borzoi dog wearing royal uniform and crown 4) pikachu in a pinstripe suit}
    \label{fig:sdxl_qualitative_comparisons}
\end{figure}

In this section, we provide qualitative comparisons between the base model, the DPO checkpoint, and CGG with the (DPO, base) pair of Appendix~\ref{sec:detail_CGG_PO} (Figures~\ref{fig:sd15_qualitative_comparisons} and~\ref{fig:sdxl_qualitative_comparisons}), followed by additional style-avoidance examples.
\begin{figure}[htbp]
    \centering
    \begin{tabular}{ccc}
        $s_\mathrm{base}$ & $s_\mathrm{DPO}$ & \shortstack{$s_\mathrm{base}+\gamma^\ast($\\$s_\mathrm{DPO}-s_\mathrm{base})$} \\
        \includegraphics[width=0.25\textwidth]{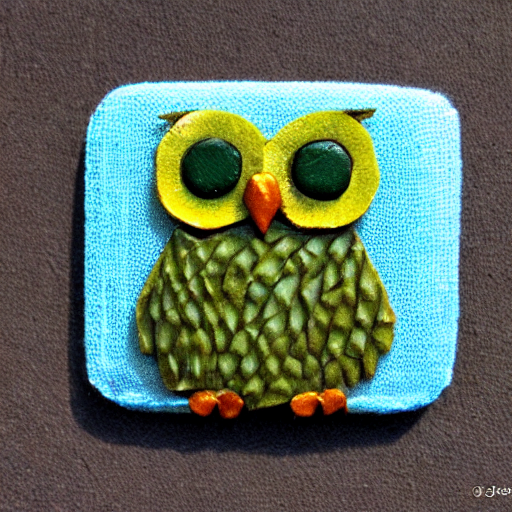} &
        \includegraphics[width=0.25\textwidth]{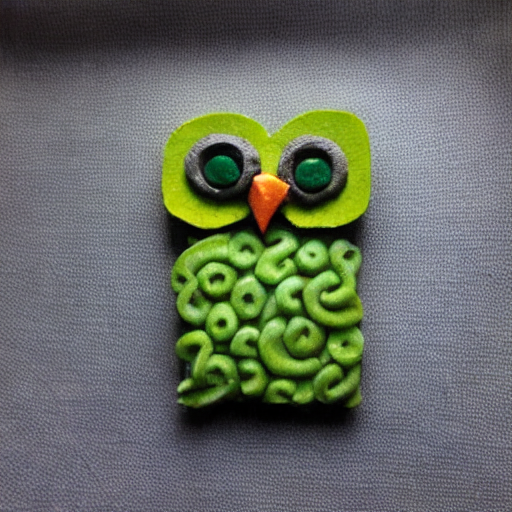} &
        \includegraphics[width=0.25\textwidth]{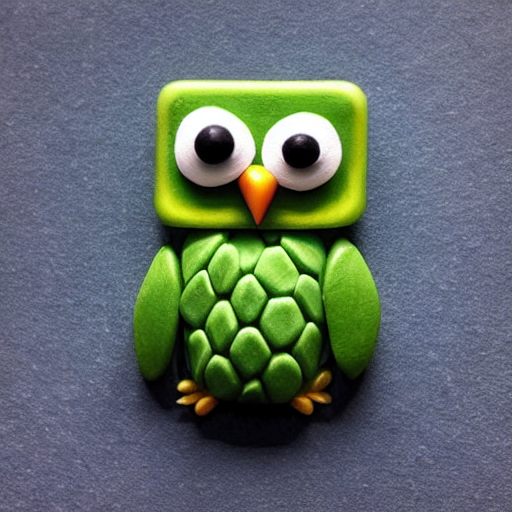} \\
        \includegraphics[width=0.25\textwidth]{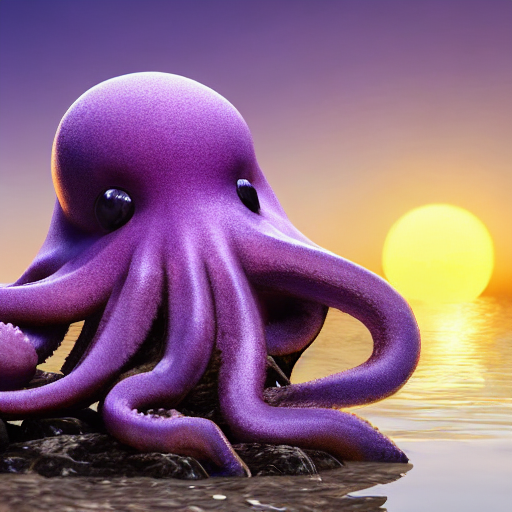} &
        \includegraphics[width=0.25\textwidth]{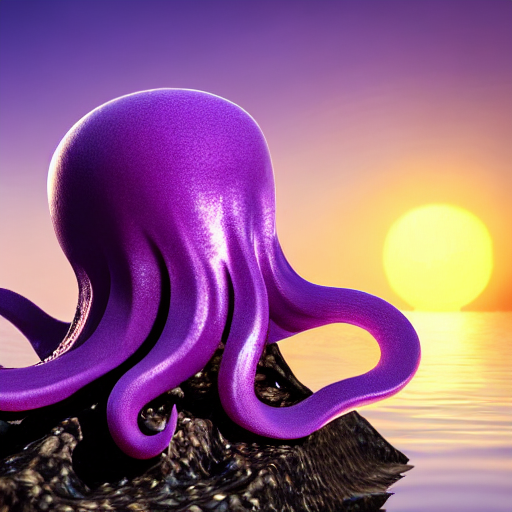} &
        \includegraphics[width=0.25\textwidth]{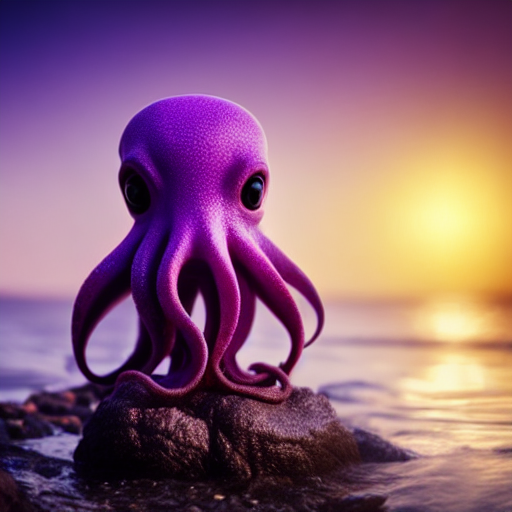} \\
        \includegraphics[width=0.25\textwidth]{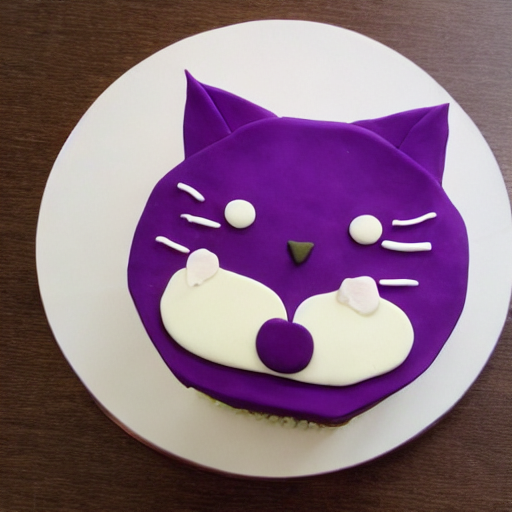} &
        \includegraphics[width=0.25\textwidth]{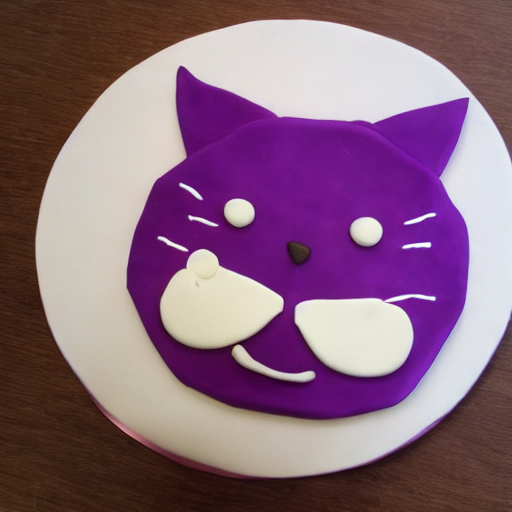} &
        \includegraphics[width=0.25\textwidth]{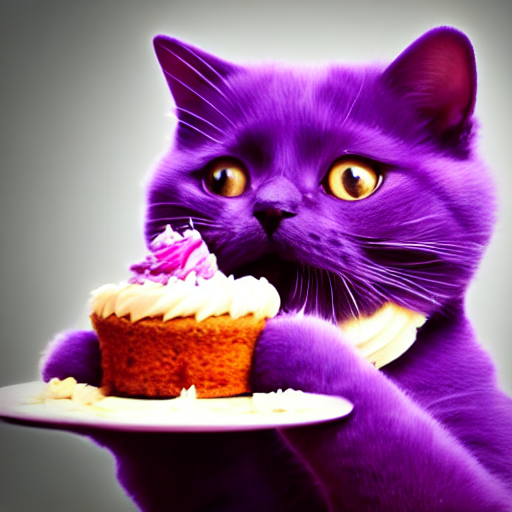} \\
        \includegraphics[width=0.25\textwidth]{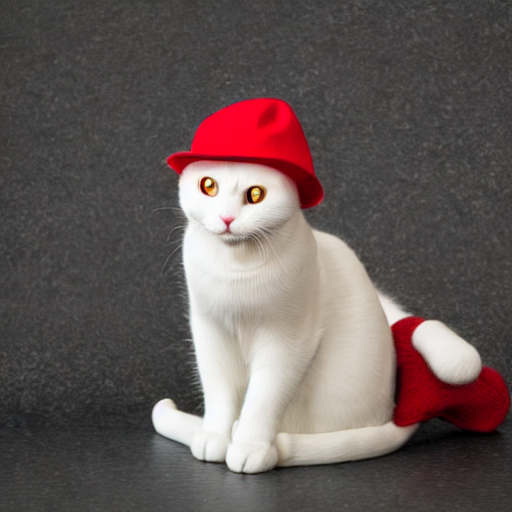} &
        \includegraphics[width=0.25\textwidth]{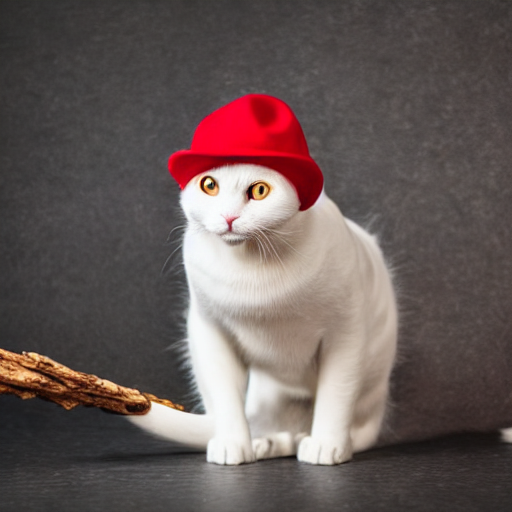} &
        \includegraphics[width=0.25\textwidth]{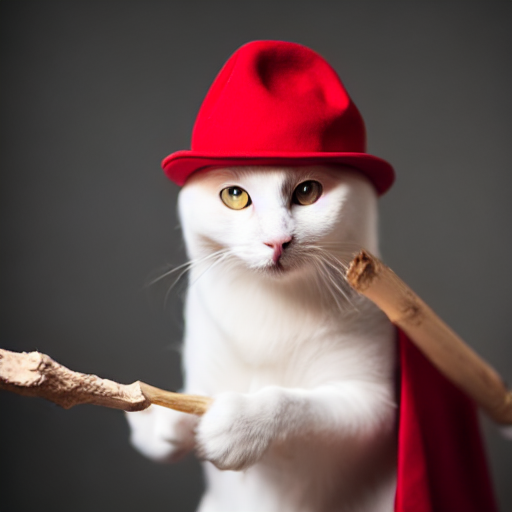}\\
    \end{tabular}
    \caption{Qualitative results on SD 1.5. Prompts: 1) a square green owl made of fimo 2) A smooth purple octopus sitting on a rock in the middle of the sea, waves crashing, golden hour, sun reflections, high quality 3d render 3) Purple cat eating cake 4) A white cat wearing a red hat holding sticks}
    \label{fig:sd15_qualitative_comparisons}
\end{figure}

\begin{figure}[htbp]
    \centering
    \includegraphics[width=\textwidth]{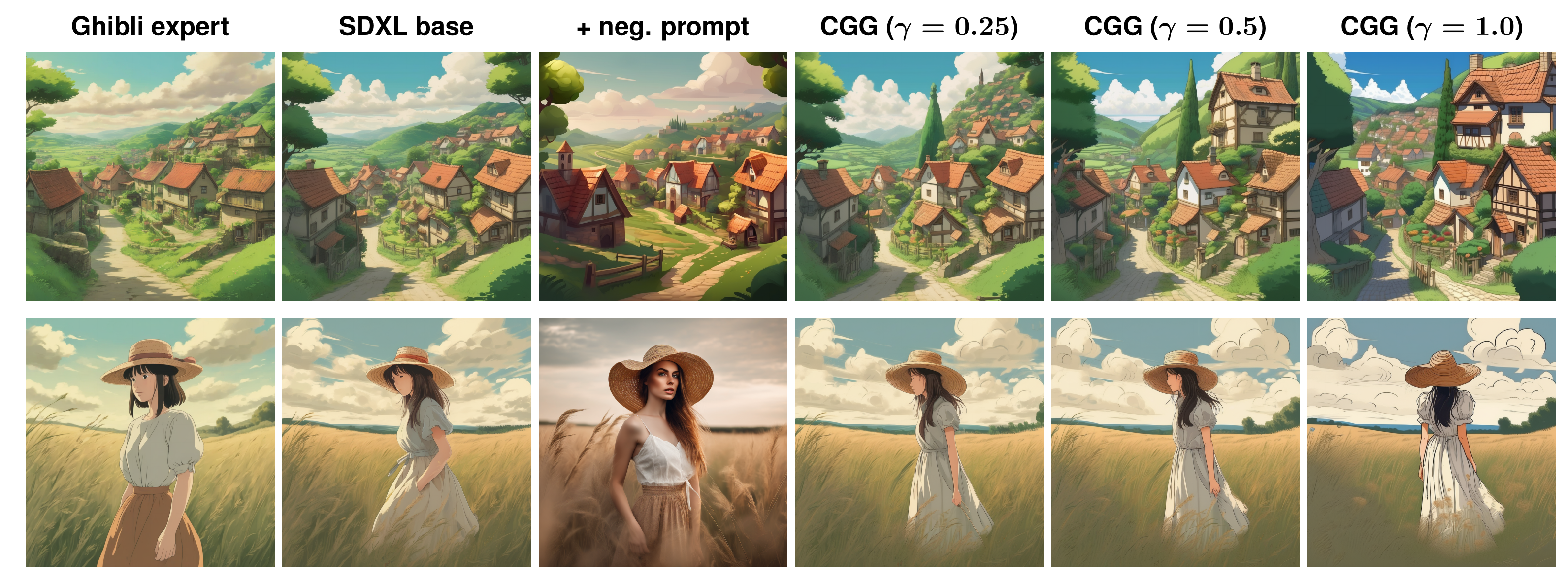}
    \caption{Full qualitative comparison of style avoidance across varying pair scales $\gamma \in \{0.25, 0.5, 1.0\}$ on two style-invocation prompts (top: countryside village; bottom: young woman in a windy meadow). Columns move from the Ghibli expert model, through the base SDXL model and text negative prompt, to CGG with increasing anti-expert scale $\gamma$.}
    \label{fig:style_avoidance_ghibli_full}
\end{figure}

\begin{figure}[htbp]
    \centering
    \begingroup
    \setlength{\tabcolsep}{2pt}
    \renewcommand{\arraystretch}{0.9}
    \scriptsize
    \begin{tabular}{@{}ccc@{}}
        \shortstack{\textbf{(a) Base SDXL}\\$s_\mathrm{base}$} &
        \shortstack{\textbf{(b) Text negative prompt}\\negative prompt names the style} &
        \shortstack{\textbf{(c) CGG anti-expert}\\$s_\mathrm{base}+\gamma(s_\mathrm{base}-s_\mathrm{Ghibli})$, $\gamma=1$} \\
        \includegraphics[width=0.2\textwidth]{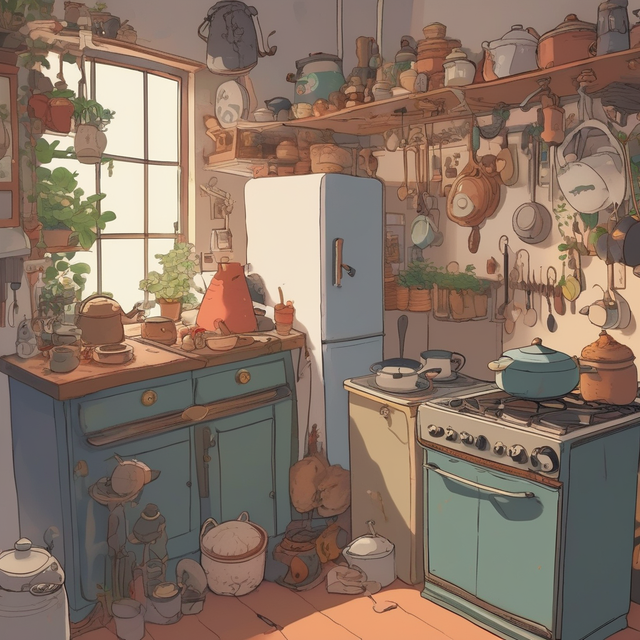} &
        \includegraphics[width=0.2\textwidth]{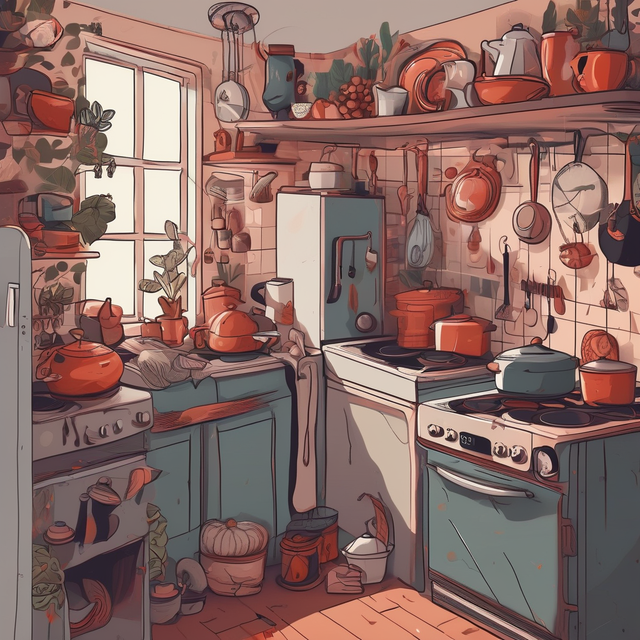} &
        \includegraphics[width=0.2\textwidth]{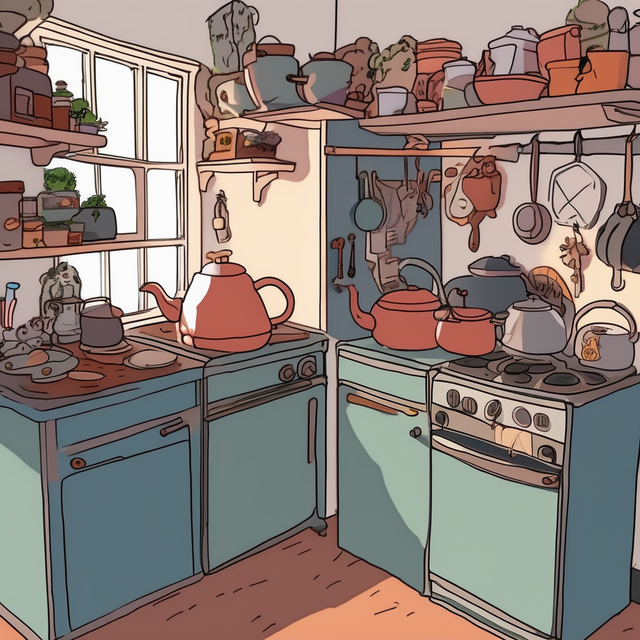} \\
        \includegraphics[width=0.2\textwidth]{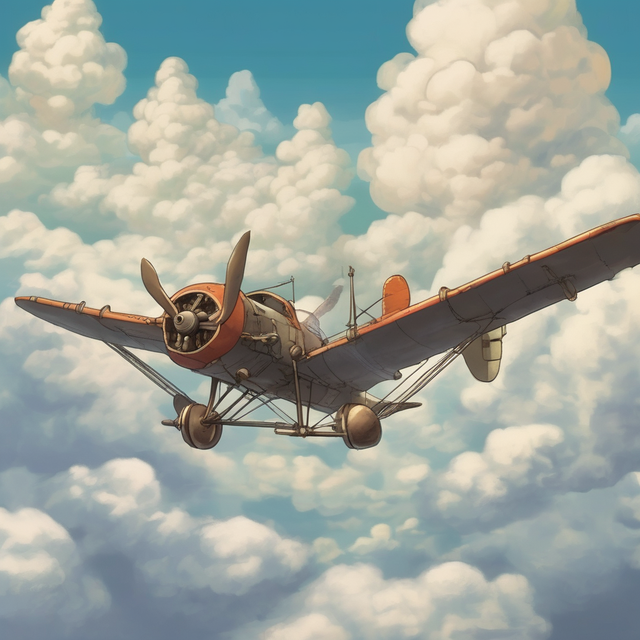} &
        \includegraphics[width=0.2\textwidth]{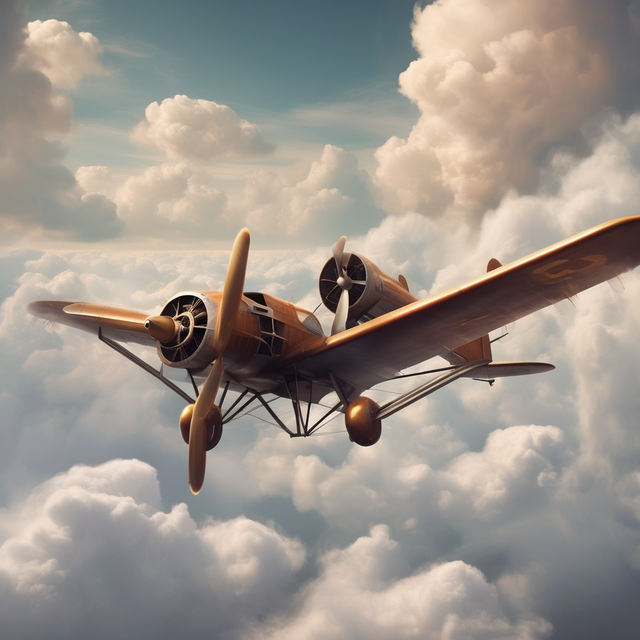} &
        \includegraphics[width=0.2\textwidth]{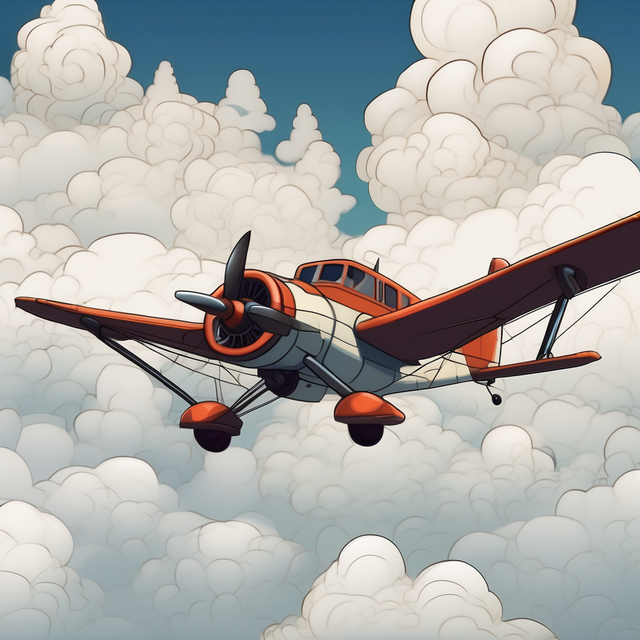} \\
        \includegraphics[width=0.2\textwidth]{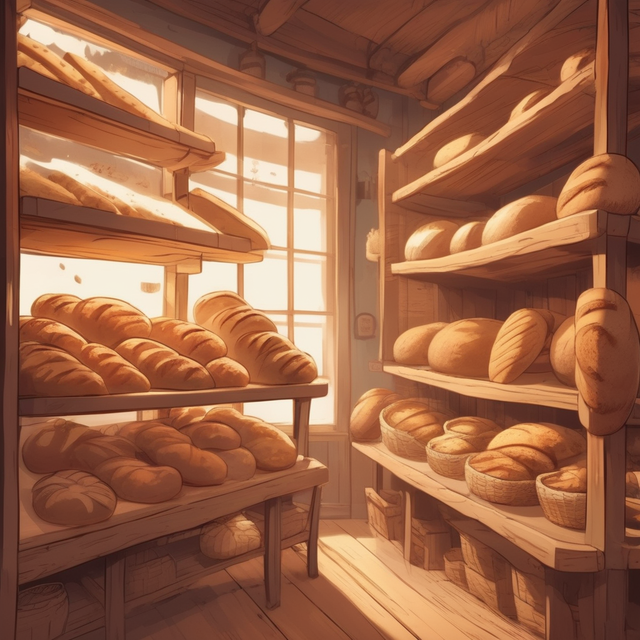} &
        \includegraphics[width=0.2\textwidth]{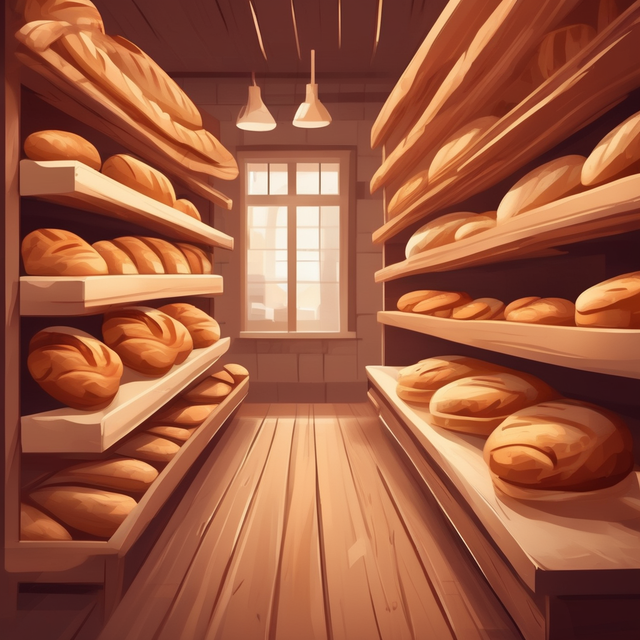} &
        \includegraphics[width=0.2\textwidth]{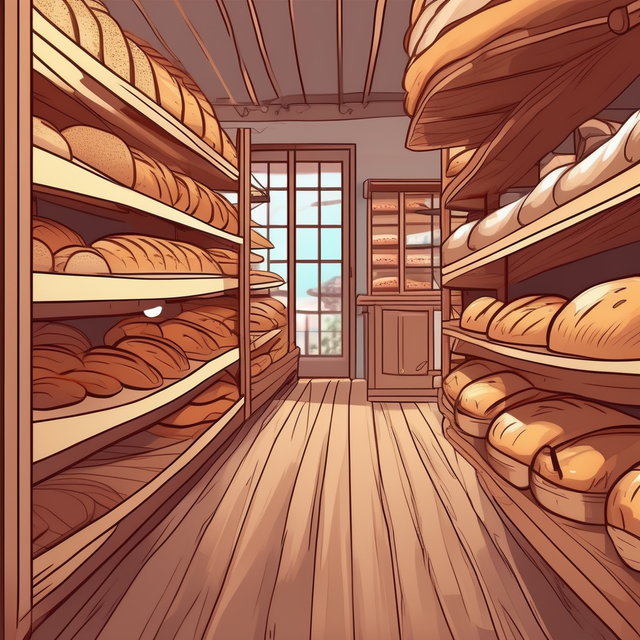} \\
        \includegraphics[width=0.2\textwidth]{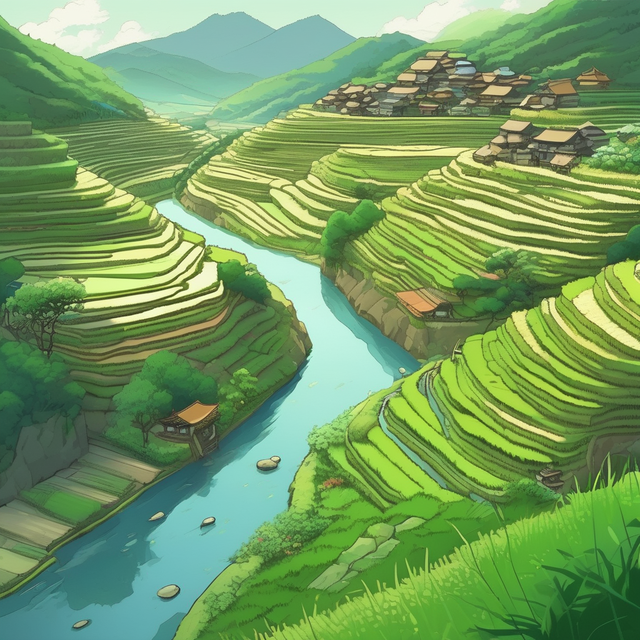} &
        \includegraphics[width=0.2\textwidth]{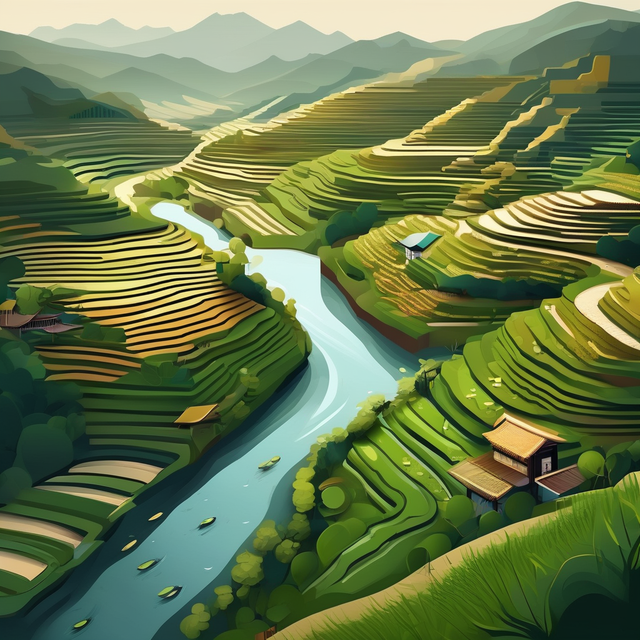} &
        \includegraphics[width=0.2\textwidth]{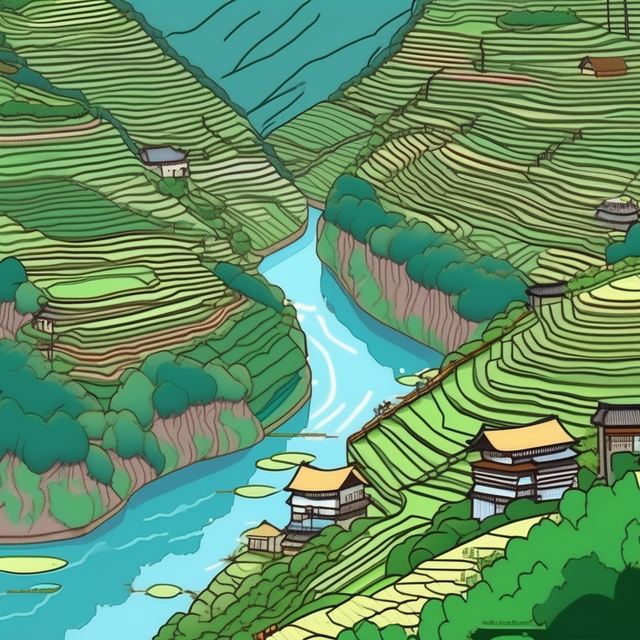} \\
    \end{tabular}
    \endgroup
    \caption{Avoiding a protected artistic style, on prompts that explicitly request it. All three columns share the prompt and the seed and differ only in where the style suppression is applied: (a) none, (b) a text negative prompt (``Studio Ghibli, ghibli, anime film still, hand-drawn animation, cel shading'') given to the same model, (c) a model-level anti-expert built by fusing a public Studio Ghibli SDXL LoRA into a standalone UNet and using it as the unfavored side of a CGG pair whose favored side and anchor are both pre-trained SDXL. Prompts, top to bottom: 1) a cozy cluttered kitchen with a kettle boiling on the stove, in the style of Studio Ghibli 2) an old propeller airplane flying above towering clouds, in the style of Studio Ghibli 3) a bakery interior with fresh bread on wooden shelves, Studio Ghibli animation style 4) a river valley with terraced rice fields in summer, in the style of Studio Ghibli. These four scenes are disjoint from the ones shown in the main text; aggregate scores over all twelve style-invocation prompts are in Table~\ref{tab:style_avoidance}.}
    \label{fig:style_avoidance_three_settings}
\end{figure}

\paragraph{Where the suppression is applied}
Figure~\ref{fig:style_avoidance_three_settings} isolates the three settings of Section~\ref{sec:experiments} on prompts that explicitly request a protected style, holding the prompt and the seed fixed across columns so that every visible difference is attributable to the suppression mechanism alone.
Column (a) is the base model, which honors the style request.
Column (b) suppresses the style inside the text-conditioning channel. The sampler resolves the conflict by moving the sample away from the picture the base model would have produced: the kitchen is pushed to a hot red palette, the sky above the propeller airplane drains to a muted haze, the bakery is recomposed around a different window, and the village disappears from the river valley.
Column (c) suppresses the style at the model level: the anti-expert subtracts the style checkpoint's contribution from the base score at each denoising step while the user's conditioning is passed through unchanged.
The layout and the objects the prompt named therefore survive: the kettles stay on the stove and the village stays on the terraces; what changes is the rendering, which leaves the studio's signature for a flatter, harder-edged look.
This is the qualitative counterpart of the matched-suppression comparison in Section~\ref{sec:experiments}: averaged over the twelve style-invocation prompts, the negative prompt reaches its suppression at a CLIP image--image similarity of $0.82$ to the base sample, whereas the anti-expert is at $0.92$ for the same amount of style removed and still $0.87$ at the $\gamma=1$ shown here, where it removes nearly twice as much.

\clearpage
\section{Extended Related Work}
\label{sec:extended_related_work}

This section provides an expanded review of related work, expanding on the comparisons presented in Section~\ref{sec:related-works}.

\paragraph{Favored and unfavored distributions for preference modeling}
Pairwise preference data contain both positive and negative information: they indicate what users favor and what they reject.
RLHF and DPO-style pipelines exploit this contrast to train reward models or directly optimize generative models~\citep{CP2017_NeurIPS_RLHF,RR2023_NeurIPS_DPO,WB2024_CVPR_DiffusionDPO}.
In language models, objectives such as SimPO, ORPO, and $f$-DPO further show that contrastive preference signals can improve alignment beyond supervised learning on favored examples alone~\citep{MY2024_NeurIPS_SimPO,HJ2025_ORPO,WC2023_fDPO}.
This perspective has inspired data-level fine-tuning methods that explicitly model the ``unfavored'' (anti-expert) side of preference data.
For instance, DExperts trains or selects expert and anti-expert language models from desirable and undesirable corpora, then combines them during decoding~\citep{LA2021_ACL_DExperts}.
In diffusion models, CHATS observes a mismatch between the fine-tuning objective of DiffusionDPO and its sampling procedure, and proposes a revised objective that jointly trains diffusion models corresponding to favored and unfavored distributions to better support inference-time sampling~\citep{MF2025_ICML_CHATS}.
Concurrently, Diffusion-NPO highlights the limitations of Diffusion-DPO in capturing unfavored preferences, modifying the preference-optimization objective to use negative preference data more directly~\citep{FW2025_ICLR_Diffusion_NPO}.
These methods use favored/unfavored structure at the data and loss level: NPO and CHATS modify the training objective and retrain models on preference-pair datasets, while DExperts relies on experts and anti-experts learned from attribute-specific corpora.
Among them, DExperts is closest in spirit to CGG: its decoding rule already adds to a base model a scaled contrast between a heterogeneous expert and anti-expert.
That rule, however, is defined on the next-token logits of an autoregressive factorization, where every decoding step exposes an exact conditional distribution over a discrete vocabulary, so the combined logits directly parameterize the next sampling step.
Diffusion sampling offers no such per-step likelihood: it follows a score field across a continuum of noise levels whose intermediate marginals differ from the data distribution, so a checkpoint contrast is meaningful only if it can be justified at every timestep.
CGG provides this justification: Eq.~\ref{eqn:implicit_reward_gradient} reads the score difference as the gradient of a timestep-indexed implicit reward, exact at the tilted optimum, and Eq.~\ref{eqn:contrastive_form} reads a designated pair as an implicit favored-versus-unfavored classifier.
A related line of work acts on the weights instead, removing an undesired concept such as a protected artistic style by permanently editing the generator, once per concept~\citep{GM2023_ICCV_ESD,GO2024_WACV_UCE}.
CGG instead lifts the same contrast to a model-level test-time composition rule.
Given existing diffusion checkpoints, CGG derives a contrastive preference direction directly from the denoising-score difference $s_{\text{expert}} - s_{\text{anti-expert}}$ at inference time, requiring no new loss design, no model retraining, and no access to the raw preference dataset that produced the checkpoints.
Because the contrast is applied during sampling rather than baked into the weights, the suppressed concept can also be changed, dialed in strength, or combined with others without retraining, which we examine for style avoidance in Section~\ref{sec:experiments}.

The contrast is therefore between \emph{data-level training}, where NPO, CHATS, and DPO variants fix the preference trade-off inside the weights of one checkpoint and must re-curate negative-preference data and retrain whenever it changes, and \emph{model-level reuse}, where off-the-shelf fine-tuned models that exhibit an undesirable attribute become anti-experts at inference time.

\paragraph{Test-time alignment for diffusion models}
As discussed in Section~\ref{sec:background}, the primary challenge of test-time alignment lies in the accurate estimation of the guidance gradient signal $\nabla_{\vx_t} r_{\text{test}}(\vc, \vx_t)$.
Existing test-time alignment methods address this through two main paradigms.
One line of work estimates the reward guidance signal during sampling by reconstructing a denoised estimate $\hat{\mathbf{x}}_0$ and evaluating its reward to adjust the guidance direction accordingly. This paradigm has been explored across various settings, including inverse problems and classifier guidance~\citep{CH2022_ICLR_DPS,YJ2023_ICCV_FreeDom,BA2023_CVPR_UniversalGuide}.
A second line of work adopts particle-based strategies, most commonly through Sequential Monte Carlo (SMC) sampling.
These methods maintain multiple particles during diffusion, resample them according to explicit reward evaluations, and thereby increase the likelihood of generating high-reward samples~\citep{WL2023_NeurIPS_ParticllSMC,SK2025_ICLR_DAS,RS2025_ICML_FKSteering}.
Later work reports that SMC-based approaches improve reward alignment over single-sample reward guidance~\citep{MS2025_ICML_FKCorrect,VJ2025_NeurIPS_DiffusionTreeSampling}.

CGG is complementary to these methods.
It constructs a preference guidance signal by composing diffusion distributions, while particle-based methods define how samples are propagated and resampled.
In Section~\ref{sec:experiments}, we combine CGG with FK Steering to study whether model-based contrastive guidance can be paired with sampling-based reward steering.

\paragraph{Compositional and test-time multi-preference alignment}
Building on the idea of compositional diffusion models~\citep{LN2022_ECCV_ComposableDiffusions,YD2024_ICML_PositionCompositionalGen}, recent works have extended compositional diffusion models from image generation to flexible test-time alignment.
For instance, RNE formulates test-time alignment through diffusion density estimation and energy-based modeling~\citep{HJ2025_arXiv_RNE}, while Product of Experts (PoE) integrates generative and discriminative experts by estimating dynamic transition kernels via Markov Chain Monte Carlo (MCMC) during sampling~\citep{ZY2025_arXiV_PoE}.
These approaches typically redesign the sampling path (transition kernel) at each diffusion step in order to incorporate the reward model.
CGG instead keeps the sampling path and changes only the score, by composing fine-tuned diffusion models.

\paragraph{Gradient-space guidance families}
The linear form of Eq.~\ref{eqn:contrastive_gradient_guidance}, an anchor score plus scaled score differences, is mathematically standard: it is the gradient-space expression of a product of experts, and variants of it recur throughout the guidance literature~\citep{HJ2022_CFG,KT2024_NeurIPS_autoguide,LJ2026_ICML_RLG}.
What separates gradient-space guidance methods is instead \emph{where the contrast comes from}, and Table~\ref{tab:guidance_taxonomy} organizes the diffusion guidance families along this axis.
Classifier-free guidance realizes the contrast at the \emph{conditioning} level: the pair is a single network under two conditions, so a negative concept must be routed through the prompt channel~\citep{HJ2022_CFG}; CFGRL extends this view by connecting CFG to policy improvement in reinforcement learning~\citep{FP2025_arXiv_CFGRL}.
The restriction matters in practice: when the user prompt and the negative concept name the same attribute, the two conditionings are in direct opposition, and Section~\ref{sec:multi-preference-results} measures the resulting over-correction on style-naming prompts, where text-level negation abandons the medium the user asked for while a model-level anti-expert leaves it intact (Table~\ref{tab:style_avoidance}).
A second family realizes the contrast within a \emph{training lineage}: Autoguidance contrasts a model against a degraded twin of itself~\citep{KT2024_NeurIPS_autoguide}, RLG contrasts an RL fine-tuned checkpoint against its own reference and interprets the guidance scale as the KL regularization strength~\citep{LJ2026_ICML_RLG}, and Diffusion Blend blends reward-specific fine-tuned models against their shared reference~\citep{CM2025_arXiv_DiffusionBlend}.
In this family the unfavored side is fixed by the training relation and coincides with the anchor, so the composition can only amplify attributes an aligned expert already encodes; Eq.~\ref{eqn:contrastive_gradient_guidance} recovers it with $N=1$ and $p_\mathrm{anc}=\kappa_1=p_\mathrm{ref}$.
CGG occupies the third family: the contrast is a \emph{free checkpoint} contrast, with the anchor, favored, and unfavored models chosen independently, the pair scales signed, and several pairs composed at once, e.g., an RLHF/DPO checkpoint against the reference model or an SFT-positive model against an SFT-negative model.
When preference trade-offs shift, traditional RLHF/DPO pipelines must retrain or re-optimize a scalarized model, and CFG can only re-word a fixed conditional/unconditional contrast.
CGG changes the trade-off by changing which model-level directions are composed and how strongly they are weighted, composing heterogeneous fine-tuned models with no retraining and no access to raw preference datasets, at the price of additional sampling-time computation that we quantify in Appendix Table~\ref{tab:inference_overhead}.

\clearpage
\section{Inference overhead}
\label{sec:inference_overhead}
Since CGG evaluates the favored and unfavored models for each active pair at each denoising step, its sampling latency scales with the number of composed pairs.
Table~\ref{tab:inference_overhead} reports wall-clock latency and peak VRAM under the shared SDXL sampling setup ($50$ denoising steps, $1024\times1024$ resolution, FP16, and CFG scale $7.5$) on a single NVIDIA H100 NVL, averaged over $19$ Pick-a-Pic v2 test prompts.
Swapping in a fine-tuned UNet (DPO) matches the base sampling cost ($3.21$ vs.\ $3.18$ sec/img).
CGG evaluates the favored and unfavored UNets of every pair at each denoising step, so its cost grows with the number of pairs: $12.53$ sec/img for two pairs and $24.93$ sec/img for the five-pair composition of Table~\ref{tab:five_reward_method_comparison_raw}.
FK Steering with four particles and a PickScore reward reaches $43.58$ sec/img, reflecting the cost of particle-based reward steering.
The two methods trade latency against memory in opposite directions.
Peak VRAM for CGG scales with the number of distinct experts held in memory, reaching $45.7$\,GB for the five-pair composition against $29.5$\,GB for FK Steering, which keeps a single UNet and batches its particles.
These numbers are for an implementation that holds each expert UNet as a full copy.
Storing experts and anti-experts as LoRA adapters~\citep{HS2022_ICLR_LoRA} on a shared frozen SDXL backbone would reduce memory; we have not measured this configuration.

\begin{table}[ht]
\centering
\small
\begin{tabular}{lcc}
\toprule
Method & sec/img ($\downarrow$) & Peak VRAM (GB, $\downarrow$) \\
\midrule
Base & 3.18 $\pm$ 0.02 & 11.5 \\
DPO & 3.21 $\pm$ 0.01 & 11.5 \\
CGG 2-pair & 12.53 $\pm$ 0.03 & 26.2 \\
CGG 5-pair & 24.93 $\pm$ 0.05 & 45.7 \\
FK Steering & 43.58 $\pm$ 0.40 & 29.5 \\
\bottomrule
\end{tabular}

\caption{Inference overhead on SDXL using $19$ Pick-a-Pic prompts, $50$ denoising steps, and CFG scale $7.5$. Lower values are better for both runtime and memory. FK Steering latency is measured with four particles and a single PickScore reward.}
\label{tab:inference_overhead}
\end{table}

\end{document}